\documentclass{article}

\usepackage{microtype}
\usepackage{graphicx}
\usepackage{booktabs}
\usepackage{caption}
\usepackage{subcaption}

\usepackage{multirow}
\usepackage{multicol}
\usepackage{natbib}

\usepackage[mathscr]{eucal}
\usepackage{amsmath}
\usepackage{amsfonts}
\usepackage{nicefrac}

\usepackage{balance}
\usepackage{xcolor}

\usepackage{url}
\usepackage{hyperref}
\usepackage{rotating}

\newcommand{\splitatcommas}[1]{%
  \begingroup
  \begingroup\lccode`~=`, \lowercase{\endgroup
    \edef~{\mathchar\the\mathcode`, \penalty0 \noexpand\hspace{0pt plus 1em}}%
  }\mathcode`,="8000 #1%
  \endgroup
}

\newcommand\blfootnote[1]{%
  \begingroup
  \renewcommand\thefootnote{}\footnote{#1}%
  \addtocounter{footnote}{-1}%
  \endgroup
}

\begin{document}

\title{Generalization and Regularization in DQN}
\author{Jesse Farebrother$^{*1}$, Marlos C. Machado$^{2}$, and Michael Bowling$^{1,3}$}

\date{%
\small{
    $^1$University of Alberta,
    $^2$Google Research,
    $^3$DeepMind Alberta
    }
}
\maketitle

\begin{abstract}
Deep reinforcement learning algorithms have shown an impressive ability to learn complex control policies in high-dimensional tasks. However, despite the ever-increasing performance on popular benchmarks, policies learned by deep reinforcement learning algorithms can struggle to generalize when evaluated in remarkably similar environments. In this paper we propose a protocol to evaluate generalization in reinforcement learning through different modes of Atari 2600 games. With that protocol we assess the generalization capabilities of DQN, one of the most traditional deep reinforcement learning algorithms, and we provide evidence suggesting that DQN overspecializes to the training environment. We then comprehensively evaluate the impact of dropout and $\ell_2$ regularization, as well as the impact of reusing learned representations to improve the generalization capabilities of DQN. Despite regularization being largely underutilized in deep reinforcement learning, we show that it can, in fact, help DQN learn more general features. These features can be reused and fine-tuned on similar tasks, considerably improving DQN's sample efficiency.
\end{abstract}

\blfootnote{$^*$Corresponding author. Contact: \href{mailto:jfarebro@cs.ualberta.ca}{jfarebro@cs.ualberta.ca}.}

\section{Introduction}

\noindent Recently, reinforcement learning (RL) algorithms have proven very successful on complex high-dimensional problems, in large part due to the use of deep neural networks for function approximation \citep[e.g.,][]{Mnih15,Silver16}. Despite the generality of the proposed solutions, applying these algorithms to slightly different environments often requires agents to learn the new task from scratch. The learned policies rarely generalize to other domains and the learned representations are seldom reusable. On the other hand, deep neural networks are lauded for their generalization capabilities \citep[e.g.,][]{Lecun98}, with some communities heavily relying on reusing learned representations in different problems. In light of the successes of supervised learning methods, the lack of generalization or reusable knowledge (i.e., policies, representation) acquired by current deep RL algorithms is somewhat surprising. 

In this paper we investigate whether the representations learned by deep RL methods can be generalized, or at the very least reused and refined on small variations to the task at hand.  We evaluate the generalization capabilities of DQN \citep{Mnih15}, one of the most representative algorithms in the family of value-based deep RL methods; and we further explore whether the experience gained by the supervised learning community to improve generalization and to avoid overfitting can be used in deep RL. We employ conventional supervised learning techniques such as regularization and fine-tuning (i.e., reusing and refining the representation)  to DQN and we show that a learned representation trained with regularization allows us to learn more general features that can be reused and fine-tuned.

We are interested in agents that generalize across tasks that have similar underlying dynamics but that have different observation spaces. In this context, we see generalization as the agent's ability to abstract aspects of the environment that do not matter. The main contributions of this work are:
\begin{enumerate}
\item We propose the use of the new modes and difficulties of Atari 2600 games as a platform for evaluating generalization in RL and we provide the first baseline results in this platform. These game modes allow agents to be trained in one environment and evaluated in a slightly different environment that still captures key concepts of the original environment (e.g., game sprites, dynamics).
\item Under this new notion of generalization in RL, we thoroughly evaluate the generalization capabilities of DQN and we provide evidence that it exhibits an overfitting trend.
\item Inspired by the current literature in regularizing deep neural networks to improve robustness and adaptability, we apply regularization techniques to DQN and show they vastly improve its sample efficiency when faced with new tasks.
We do so by analyzing the impact of regularization on the policy's ability to not only perform zero-shot generalization, but to also learn a more general representation amenable to fine-tuning on different problems.
\end{enumerate}

\section{Background}

We begin our exposition with an introduction of basic terms and concepts for supervised learning and reinforcement learning. We then discuss the related work, focusing on generalization in reinforcement learning.

\subsection{Regularization in Supervised Learning}
In the supervised learning problem we are given a dataset of examples represented by a matrix $X \in \mathbb{R}^{m \times n}$ with $m$ training examples of dimension $n$, and a vector $\mathbf{y} \in \mathbb{R}^{1 \times m}$ denoting the output target $y_i$ for each training example $X_i$. 
We want to learn a function which maps each training example $X_i$ to its predicted output label $\hat{y}_i$. The goal is to learn a robust model that accurately predicts $y_i$ from $X_i$ while generalizing to unseen training examples. In this paper we focus on using a neural network parameterized by the weights $\theta$ to learn the function $f$ such that $f(X_i;\, \theta) = \hat{y}_i$. We typically train these models by~minimizing
\begin{equation*}
\min_{\theta} \,\, \frac{\lambda}{2} \,\, {\left\| \theta \right\|}^{2}_{2} + \frac{1}{m} \sum_{i = 1}^{m}{L\big(y_i, f(X_i;\, \theta)\big)}\text{,}
\end{equation*}
where $L$ is a differentiable loss function which outputs a scalar determining the quality of the prediction (e.g., squared error loss).
The first term is a form of regularization, that is, $\ell_2$ regularization, which encourages generalization by imposing a penalty on large weight vectors. The hyperparameter $\lambda$ is the weighted importance of the regularization~term.

Another popular regularization technique is dropout \citep{Srivastava14}. When using dropout, during forward propagation each neural unit is set to zero according to a Bernoulli distribution with probability $p \in [0, 1]$, referred to as the dropout rate. Dropout discourages the network from relying on a small number of neurons to make a prediction, making memorization of the dataset harder.

Prior to training, the network parameters are usually initialized through a stochastic process such as Xavier initialization \citep{Glorot10}.
We can also initialize the network using pre-trained weights from a different task.
If we reuse one or more pre-trained layers we say the weights encoded by those layers will be fine-tuned during training~\citep[e.g.,][]{Razavian14, Long15}, a topic we explore in Section~\ref{sec:fine_tuning}.

\subsection{Reinforcement Learning}
In the reinforcement learning (RL) problem an agent interacts with an environment with the goal of maximizing cumulative long term reward. RL problems are often modeled as a Markov decision process (MDP), defined by a 5-tuple $\langle \mathscr{S}, \mathscr{A}, p, r, \gamma \rangle$. At a discrete time step $t$, the agent observes the current state $S_t \in \mathscr{S}$ and takes an action $A_t \in \mathscr{A}$ to transition to the next state $S_{t+1} \in \mathscr{S}$ according to the transition dynamics function $p(s' \,|\, s, a ) \doteq P(S_{t+1} = s' \,|\, S_t = s\,, A_t = a) $. The agent receives a reward signal $R_{t+1}$ according to the reward function $r : \mathscr{S} \times \mathscr{A} \to \mathbb{R}$. The agent's goal is to learn a policy $\pi : \mathscr{S} \times \mathscr{A} \rightarrow [0, 1]$, written as $\pi(a\,|\,s)$, which is defined as the conditional probability of taking action $a$ in state $s$. The learning agent refines its policy with the objective of maximizing the expected return, that is, the cumulative discounted reward incurred from time $t$, defined by $G_t \doteq \sum_{k=0}^\infty \gamma^k R_{t+k+1}$, where $\gamma \in [0, 1)$ is the discount~factor.

Q-learning \citep{Watkins92} is a traditional approach to learning an optimal policy from samples obtained from interactions with the environment. For a given policy $\pi$, we define the state-action value function as the expected return conditioned on a state and action ${q_{\pi}(s, a) \doteq \mathop{\mathbb{E}}_{\pi} \big[ G_t | S_0 = \, s ,  A_0 = \, a \big]}$. The agent iteratively updates the state-action value function based on samples from the environment using the update rule
\begin{equation*}
\begin{split}
Q(S_t, A_t) \gets & Q(S_t, A_t) + \alpha \big[ R_{t+1} + \gamma \max_{a' \in \mathcal{A}}{Q(S_{t+1}, a')} - Q(S_t, A_t) \big],
\end{split}
\end{equation*}
where $t$ denotes the current timestep and $\alpha$ the step size.
Generally, due to the exploding size of the state space in many real-world problems, it is intractable to learn a state-action pairing for the entire MDP. Instead we learn an approximation to the true function~$q_{\pi}$.

DQN approximates the state-action value function such that $ Q(s, a; \, \theta) \approx q_{\pi}(s, a)$, where $\theta$ denotes the weights of a neural network. The network takes as input some encoding of the current state $S_t$ and outputs $|\mathcal{A}|$ scalars corresponding to the state-action values for $S_t$.
DQN is trained to minimize
\begin{equation*}
\begin{split}
    L^{\text{\tiny\textsc{DQN}}} = &\mathop{\mathbb{E}}_{\tau \, \sim \, U(\cdot)} \big[ \big( R_{t+1} + \gamma \max_{a' \in \mathcal{A}} Q(S_{t+1}, a';\, \theta^{-}) - Q(S_t, A_t;\, \theta) \big{)}^2 \big],
\end{split}
\end{equation*}
where $\tau = ( S_t, A_t, R_{t+1}, S_{t+1} )$ are uniformly sampled from $U(\cdot)$, the experience replay buffer filled with experience collected by the agent. The weights $\theta^{-}$ of a duplicate network are updated less frequently for stability purposes.

\subsection{Related Work}
In reinforcement learning, regularization is rarely applied to value-based methods. The few existing studies often focus on single-task settings with linear function approximation \citep[e.g.,][]{Farahmand08,Kolter09}. Here we look at the reusability, in different tasks, of \emph{learned} representations. The closest work to ours is \citeauthor{Cobbe19}'s (\citeyear{Cobbe19}), which also looks at regularization techniques applied to deep RL. However, different from \citeauthor{Cobbe19}, here we also evaluate the impact of regularization when fine-tuning value functions. Moreover, in this paper we propose a different platform for evaluating generalization in RL, which we discuss below.

There are several recent papers that support our results with respect to the limited generalization capabilities of deep RL agents. Nevertheless, they often investigate generalization in light of different aspects of an environment such as noise \citep[e.g.,][]{Zhang18_2} and start state distribution \citep[e.g.,][]{Rajeswaran17,Zhang18_2,Zhang18}. There are also some proposals for evaluating generalization in RL through procedurally generated or parametrized environments \citep[e.g.,][]{Finn17,Juliani19,Justesen18,Whiteson11,Witty18}. These papers do not investigate generalization in deep RL the same way we do. Moreover,  as aforementioned, here we also propose using a different testbed, the modes and difficulties of Atari 2600 games. With respect to that, \citeauthor{Witty18}'s (\citeyear{Witty18}) work is directly related to ours, as they propose  parameterizing a single Atari 2600 game, Amidar, as a way to evaluate generalization in RL. The use of modes and difficulties is much more comprehensive and it is free of experimenters' bias. 

In summary, our work adds to the growing literature on generalization in reinforcement learning. To the best of our knowledge, our paper is the first to discuss overfitting in Atari 2600 games, to present results using the Atari 2600 modes as testbed, and to demonstrate the impact regularization can have in value function fine-tuning in reinforcement learning. 

\begin{figure*}[t]
    \centering
    \def\arraystretch{1.2}
    \setlength\tabcolsep{2pt}
    \begin{tabular}{cccc}
        \textsc{Freeway} & \textsc{Hero} & \textsc{Breakout} & \textsc{Space Invaders} \\
        \includegraphics[width=.24\textwidth]{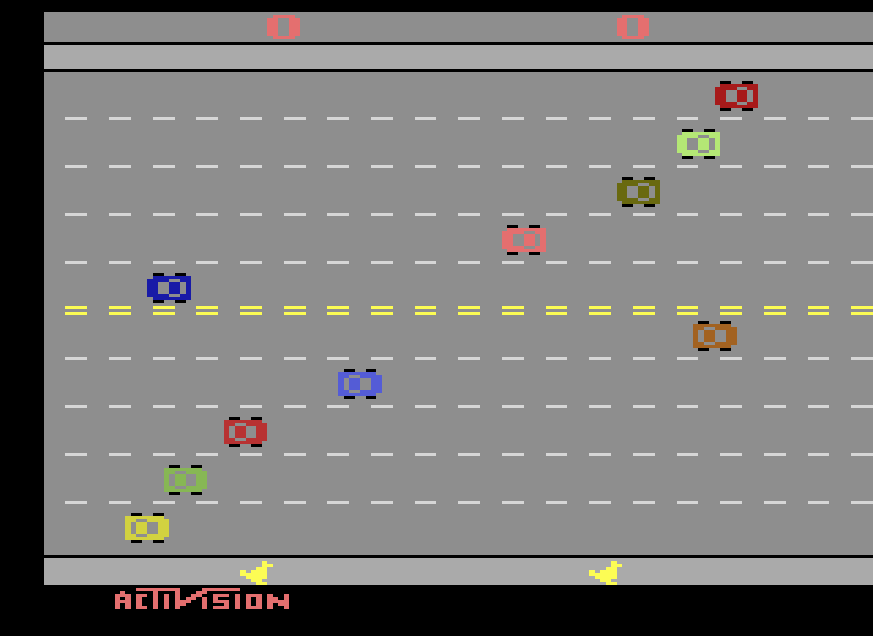}
        & \includegraphics[width=.24\textwidth]{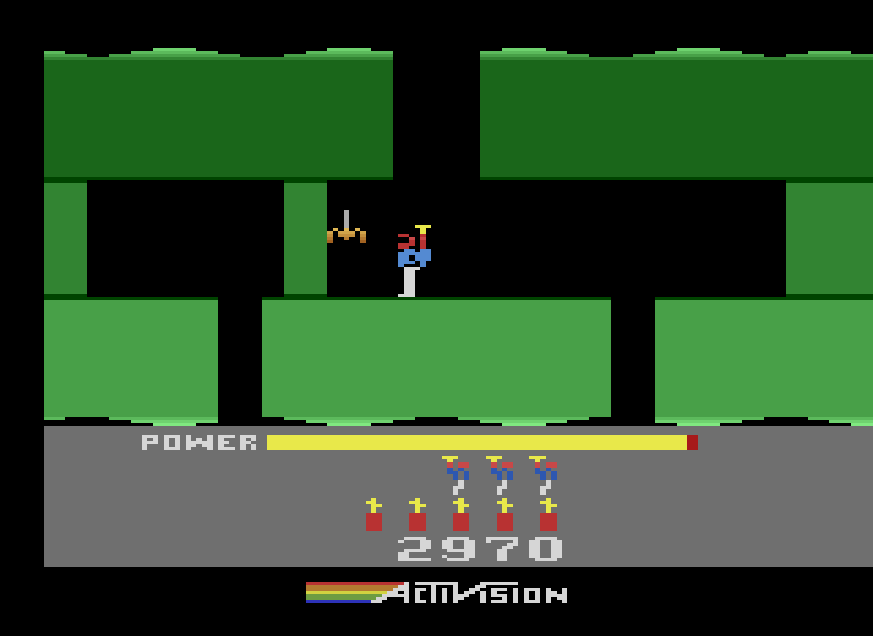}
        & \includegraphics[width=.24\textwidth]{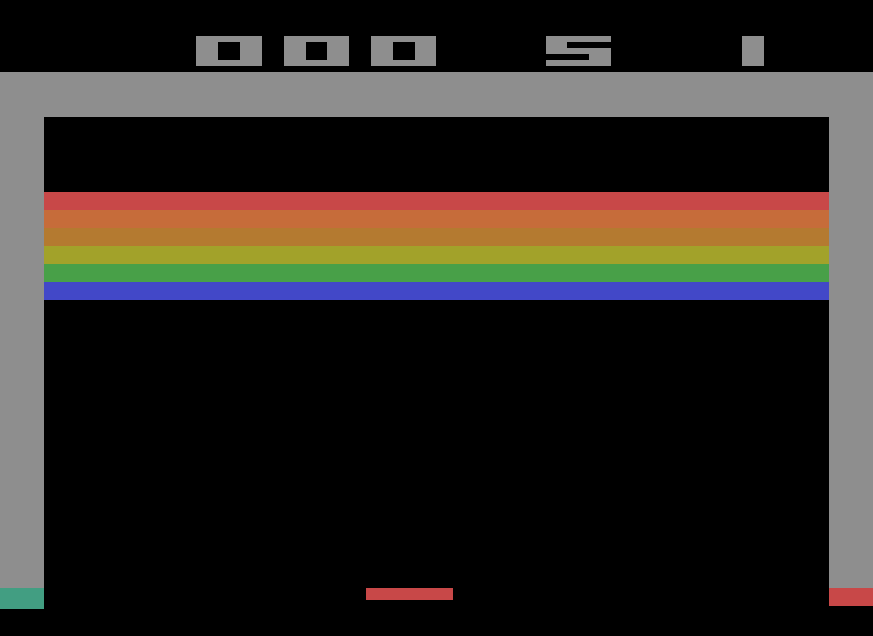}
        & \includegraphics[width=.24\textwidth]{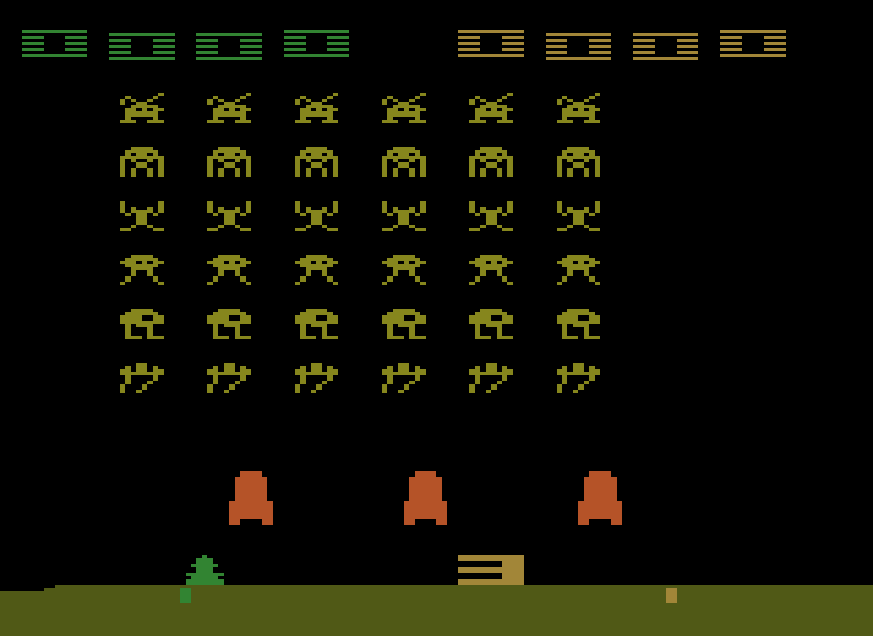} \\
        \includegraphics[width=.24\textwidth]{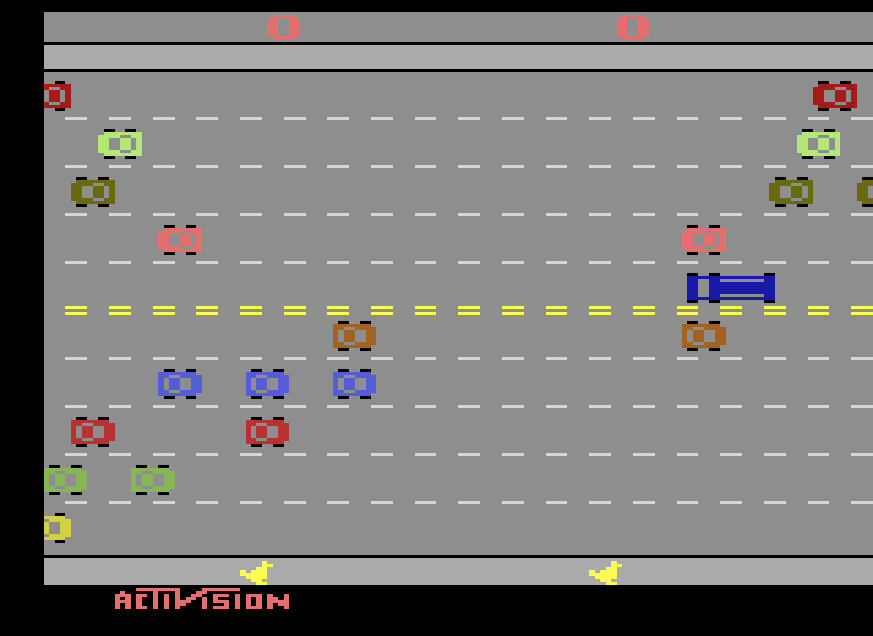}
        & \includegraphics[width=.24\textwidth]{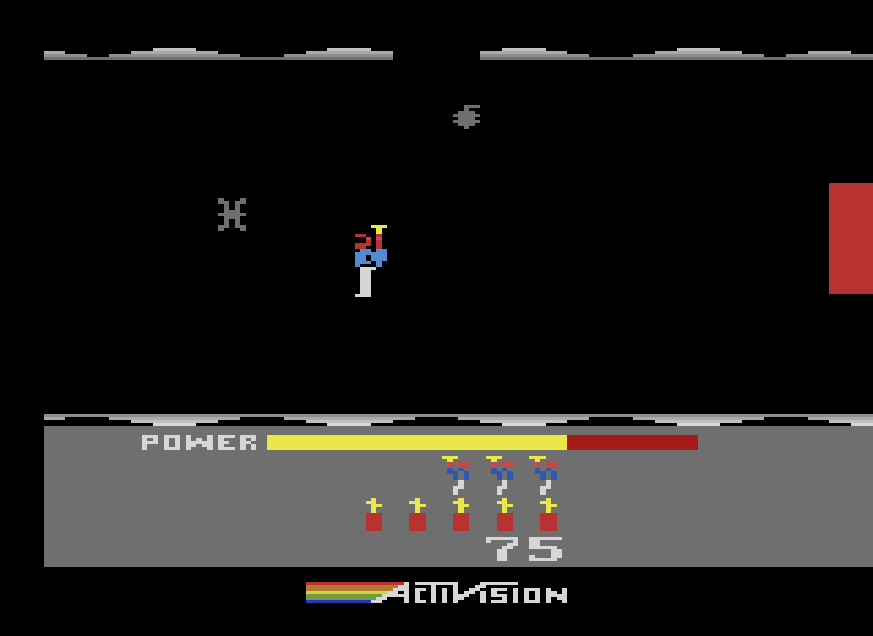}
        & \includegraphics[width=.24\textwidth]{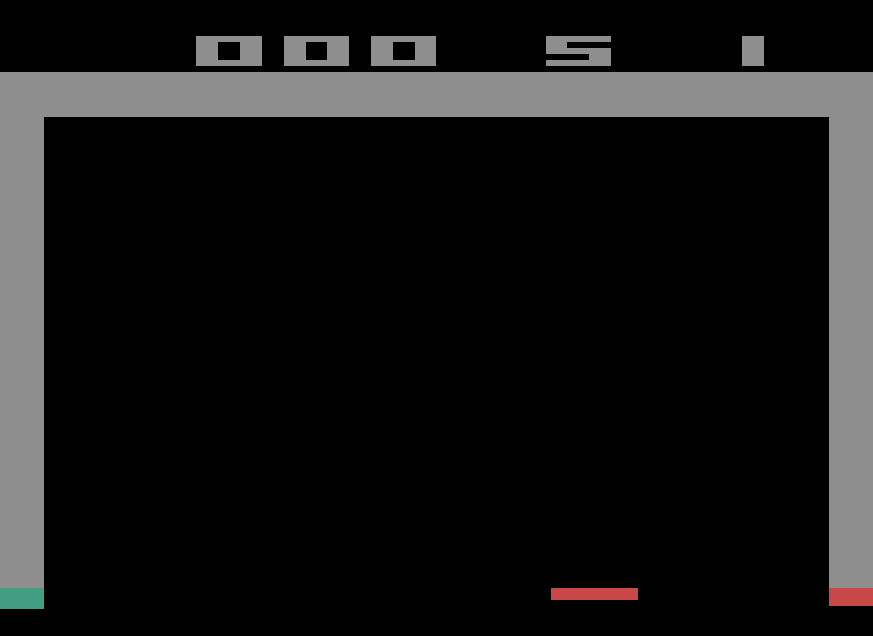}
        & \includegraphics[width=.24\textwidth]{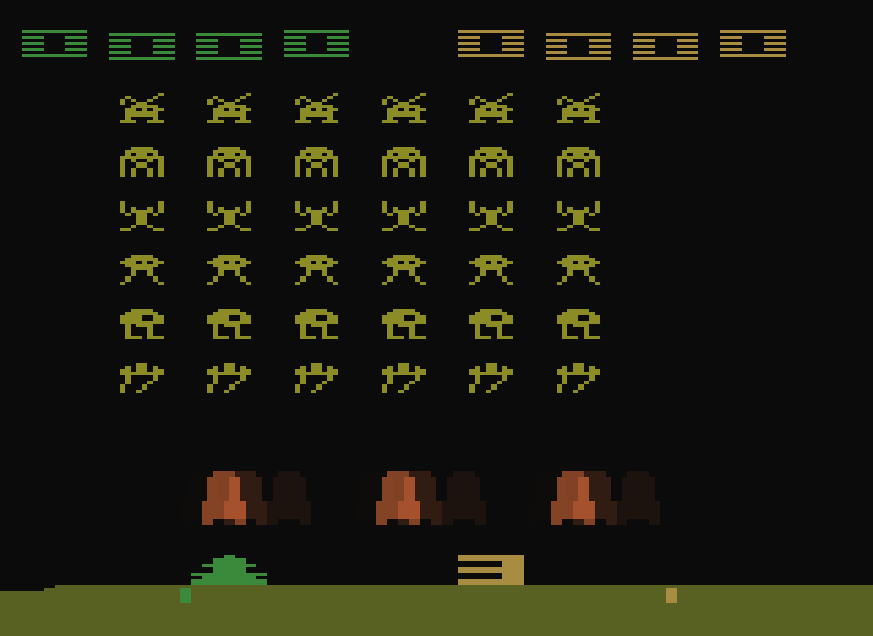}
    \end{tabular}
    \caption{Column show the variations between two flavours of each game.}
    \label{fig:alemodes}
\end{figure*}

\section{The ALE as a Platform for Evaluating Generalization in Reinforcement Learning}

The Arcade Learning Environment (ALE) is a platform used to evaluate agents across dozens of Atari 2600 games \citep{Bellemare13}. It is one of the standard evaluation platforms in the field and has led to several exciting algorithmic advances~\citep[e.g.,][]{Mnih15}.
The ALE poses the problem of general competency by having agents use the same learning algorithm to perform well in as many games as possible, without using any game specific knowledge.
Learning to play multiple games with the same agent, or learning to play a game by leveraging knowledge acquired in a different game is harder, with fewer successes being known~\citep{Rusu16, Kirkpatrick16, Parisotto16, Schwarz18,Espeholt18}.

Throughout this paper we evaluate the generalization capabilities of our agents using hold out test environments. We do so with different modes and difficulties of Atari 2600 games, features the ALE recently started to support \citep{Machado18a}. Game modes, which were originally native to the Atari 2600 console, generally give us modifications of each Atari 2600 game by modifying sprites, velocities, and the observability of objects. These modes offer an excellent framework for evaluating generalization in RL. They were designed several decades ago and remain free from experimenter's bias as they were not designed with the goal of being a testbed for AI agents, but with the goal of being varied\footnote{There are 48 Atari 2600 games with more than one flavour in the~ALE. These games have 414 different flavours \citep{Machado18a}. Notice that, on average, each game has less than 10 flavours though. This is another challenge since other settings often assume access to many more environment variations (e.g., via procedural content generation).} and entertaining to humans. Figure~\ref{fig:alemodes} depicts some of the different modes and difficulties available in the ALE. As \citet{Machado18a}, hereinafter we call each mode/difficult pair a \emph{flavour}.

Besides having the properties that made the ALE successful in the RL community, the different game flavours allow us to look at the problem of generalization in RL from a different perspective. Because of hardware limitations, the different flavours of an Atari 2600 game could not be too different from each other.\footnote{The Atari 2600 console has only 2KB of RAM.} Therefore, different flavours can be seen as small variations of the default game, with few latent variables being changed. In this context, we pose the problem of generalization in RL as the ability to identify invariances across tasks with high-dimensional observation spaces. Such an objective is based on the assumption that the underlying dynamics of the world does not vary much. Instead of requiring an agent to play multiple games that are visually very different or even non-analogous, the notion of generalization we propose requires agents to play games that are visually very similar and that can be played with policies that are conceptually similar, at least from a human perspective. In a sense, the notion of generalization we propose requires agents to be invariant to changes in the observation space. 

Introducing flavours to the ALE is not one of our contributions, this was done by \cite{Machado18a}. Nevertheless, here we provide a first concrete suggestion on how to use these flavours in reinforcement learning. Our paper also provides the first baseline results for different flavours of Atari 2600 games since \cite{Machado18a} incorporated them to the ALE but did not report any results on them. The baseline results for the traditional deep RL setting are available in Table~\ref{table:baselines} while the full baseline results for regularization are available in Table~\ref{table:baselinesreg}. Because these baseline results are quite broad, encompassing multiple games and flavours, and because we wanted to first discuss other experiments and analyses, Tables~\ref{table:baselines} and~\ref{table:baselinesreg} are at the end of the paper. They follow \citeauthor{Machado18a}'s (\citeyear{Machado18a}) suggestions on how to report Atari 2600 games results.

We believe our proposal is a more realistic and tractable way of defining generalization in decision-making problems.  Instead of focusing on the samples $(s, a, s', r)$, simply requiring them be drawn from the same distribution, we look at a more general notion of generalization where we consider multiple tasks, with the assumption that tasks are sampled from the same distribution, similar to the meta-RL setting. Nevertheless, we concretely constrain the distribution of tasks with the notion that only few latent variables describing the environment can vary. This also allows us to have a new perspective towards an agents' inability to succeed in slightly different tasks from those they are trained on. At the same time, this is more challenging than using, for example, different parametrizations of an environment, as often done when evaluating meta-RL algorithms. In fact, we could not obtain any positive results in these Atari 2600 games with traditional meta-RL algorithms \citep[e.g.,][]{Finn17,Nichol18a} and to the best of our knowledge, there are no reports of meta-RL algorithms succeeding in Atari 2600 games. Because of that, we do not further discuss these approaches.

In this paper we focus on a subset of Atari 2600 games with multiple flavours. Because we wanted to provide exhaustive results averaging over multiple trials, here we use $13$ flavours obtained from $4$ games: \textsc{Freeway}, \textsc{HERO}, \textsc{Breakout}, and \textsc{Space Invaders}.
In \textsc{Freeway}, the different modes vary the speed and number of vehicles, while different difficulties change how the player is penalized for running into a vehicle. In \textsc{HERO}, subsequent modes start the player off at increasingly harder levels of the game. The mode we use in \textsc{Breakout} makes the bricks partially observable. Modes of \textsc{Space Invaders} allow for oscillating shield barriers, increasing the width of the player sprite, and partially observable aliens. Figure~\ref{fig:alemodes} depicts some of these flavours and Figure~\ref{fig:description} further explains the  difference between the ALE flavours we used.\footnote{Videos of the different modes are available in the following link: \url{https://goo.gl/pCvPiD}.}

\begin{figure}[t]
\hspace{-0.4cm}
\fcolorbox{black}{lightgray}{
\begin{minipage}{\columnwidth}

\footnotesize

\textbf{\textsc{Freeway}}: a chicken must cross a road containing multiple lanes of moving traffic within a prespecified time limit. In all modes of \textsc{Freeway} the agent is rewarded for reaching the top of the screen and is subsequently teleported to the bottom of the screen.
If the chicken collides with a vehicle in difficulty 0 it gets bumped down one lane of traffic, alternatively, in difficulty 1 the chicken gets teleported to its starting position at the bottom of the screen.
Mode 1 changes some vehicle sprites to include buses, adds more vehicles to some lanes, and increases the velocity of all vehicles.
Mode 4 is almost identical to Mode 1; the only difference being vehicles can oscillate between two speeds.
Mode 0, with difficulty 0, is the default one.

\vspace{0.1cm}

\textbf{\textsc{Hero}}: you control a character who must navigate a maze in order to save a trapped miner within a cave system. 
The agent scores points for forward progression such as clearing an obstacle or killing an enemy.
Once the miner is rescued, the level is terminated and you continue to the next level in a different maze. Some levels have partially observable rooms, more enemies, and more difficult obstacles to traverse. Past the default mode (m0d0), each subsequent mode starts off at increasingly harder levels denoted by a level number increasing by multiples of $5$. The default mode starts you off at level $1$, mode 1 starts at level $5$, etc.

\vspace{0.1cm}

\textbf{\textsc{Breakout}}: you control a paddle which can move horizontally along the bottom of the screen. At the beginning of the game, or on a loss of life, the ball is set into motion and can bounce off the paddle and collide with bricks at the top of the screen. The objective of the game is to break all the bricks without having the ball fall below your paddles horizontal plane.
Subsequently, mode 12 of \textsc{Breakout} hides the bricks from the player until the ball collides with the bricks in which case the bricks flash for a brief moment before disappearing again.

\vspace{0.1cm}

\textbf{\textsc{Space Invaders}}: you control a spaceship which can move horizontally along the bottom of the screen. There is a grid of aliens above you and the objective of the game is to eliminate all the aliens. You are afforded some protection from the alien bullets with three barriers just above your spaceship. Difficulty 1 of \textsc{Space Invaders} widens your spaceships sprite making it harder to dodge enemy bullets. Mode 1 of \textsc{Space Invaders} causes the shields above you to oscillate horizontally. Mode 9 of \textsc{Space Invaders} is similar to Mode 12 of \textsc{Breakout} where the aliens are partially observable until struck with the player's bullet. Mode 0, with difficulty 0, is the default one.

\end{minipage}
}
\caption{Description of the game flavours used in the paper.}
\label{fig:description}
\end{figure}

\section{Generalization of the Policies Learned by DQN}

In order to test the generalization capabilities of DQN, we first evaluate whether a policy learned in one flavour can perform well in a different flavour. As aforementioned, different modes and difficulties of a single game look very similar.
If the representation encodes a robust policy we might expect it to be able to generalize to slight variations of the underlying reward signal, game dynamics, or observations. Evaluating the learned policy in a similar but different flavour can be seen as evaluating generalization in RL, similar to cross-validation in supervised~learning.

To evaluate DQN's ability to generalize across flavours, we evaluate the learned $\epsilon$-greedy policy on a new flavour after training for 50M frames in the default flavour, m0d0 (mode 0, difficulty 0). We measure the cumulative reward averaged over 100 episodes in the new flavour, adhering to the evaluation protocol suggested by \citet{Machado18a}. The results are summarized in Table~\ref{table:dpdp}.
Baseline results where the agent is trained from scratch for 50M frames in the target flavour used for evaluation are reported in the baseline column \textsc{Learn Scratch}.
Theoretically, this baseline can be seen as an upper bound on the performance DQN can achieve in that flavour, as it represents the agent's performance when evaluated in the same flavour it was trained on.
Full baseline results with the agent's performance after different number of frames can be found in Tables~\ref{table:baselines} and~\ref{table:baselinesreg}.

\begin{table}[t]
    \centering
    \caption{Direct policy evaluation. Each agent is initially trained in the default flavour for 50M frames then evaluated in each listed game flavour. Reported numbers are averaged over five runs. Std. dev. is reported between parentheses.}
    \footnotesize
    \setlength\tabcolsep{4pt}
    \resizebox*{0.75\columnwidth}{!}{
        \begin{tabular}{ll rl rl}

\multicolumn{2}{c}{\textsc{Game Variant}}
& \multicolumn{2}{c}{\textsc{Evaluation}}
& \multicolumn{2}{c}{\textsc{Learn Scratch}} \\ \midrule[0.4mm]

\multirow{4}{*}{\textsc{Freeway}}
& m1d0
& 0.2 & (0.2)
& \textbf{4.8} & \textbf{(9.3)} \\ \cmidrule(l){2-6}

& m1d1
& \textbf{0.1} & \textbf{(0.1)}
& 0.0 & (0.0) \\ \cmidrule(l){2-6}

& m4d0
& 15.8 & (1.0)
& \textbf{29.9} & \textbf{(0.7)} \\ \cmidrule[0.2mm]{1-6}

\multirow{3}{*}{\textsc{Hero}}
& m1d0
& 82.1 & (89.3)
& \textbf{1425.2} & \textbf{(1755.1)} \\ \cmidrule(l){2-6}

& m2d0
& 33.9 & (38.7)
& \textbf{326.1} & \textbf{(130.4)} \\ \cmidrule[0.2mm]{1-6}

\multirow{1}{*}{{\textsc{Breakout}}}

& m12d0
& 43.4 & (11.1)
& \textbf{67.6} & \textbf{(32.4)} \\ \cmidrule[0.2mm]{1-6}

\multirow{4}{*}{\textsc{Space Invaders}}

& m1d0
& 258.9 & (88.3)
& \textbf{753.6} & \textbf{(31.6)} \\ \cmidrule(l){2-6}

& m1d1
& 140.4 & (61.4)
& \textbf{698.5} & \textbf{(31.3)} \\ \cmidrule(l){2-6}

& m9d0
& 179.0 & (75.1)
& \textbf{518.0} & \textbf{(16.7)} \\ \cmidrule[0.2mm]{1-6}
\end{tabular}

    }
    \label{table:dpdp}
\end{table}

We can see in the results that the policies learned by DQN do not generalize well to different flavours, even when the flavours are remarkably similar. For example, in \textsc{Freeway}, a high-level policy applicable to all flavours is to go up while avoiding cars. This does not seem to be what DQN learns. For example, the default flavour m0d0 and m4d0 comprise of exactly the same sprites, the only difference is that in m4d0 some cars accelerate and decelerate over time. The close to optimal policy learned in m0d0 is only able to score 15.8 points when evaluated on m4d0, which is approximately half of what the policy learned from scratch in that flavour achieves (29.9 points). The learned policy when evaluated on flavours that differ more from m0d0 perform even worse (for example, when a new sprite is introduced, or when there are more cars in each lane).

As aforementioned, the different modes of \textsc{HERO} can be seen as giving the agent a curriculum or a natural progression. Interestingly, the agent trained in the default mode for 50M frames can progress to at least level 3 and sometimes level 4. Mode 1 starts the agent off at level 5 and performance in this mode suffers greatly during evaluation. There are very few game mechanics added to level 5, indicating that perhaps the agent is memorizing trajectories instead of learning a robust policy capable of solving each~level.

Results in some flavours suggest that the agent is overfitting to the flavour it is trained on. We tested this hypothesis by periodically evaluating the learned policy in each other flavour of that game. This process involved taking checkpoints of the network every $500{,}000$ frames and evaluating the $\epsilon$-greedy policy in the prescribed flavour for $100$ episodes, further averaged over five runs. 
The results obtained in \textsc{Freeway}, the most pronounced game in which we observe overfitting, are depicted in Figure~\ref{fig:policytransfer}. Learning curves for all flavours can be found in the Appendix.

\begin{figure}[t]
\centering
    \includegraphics[width=0.6\columnwidth]{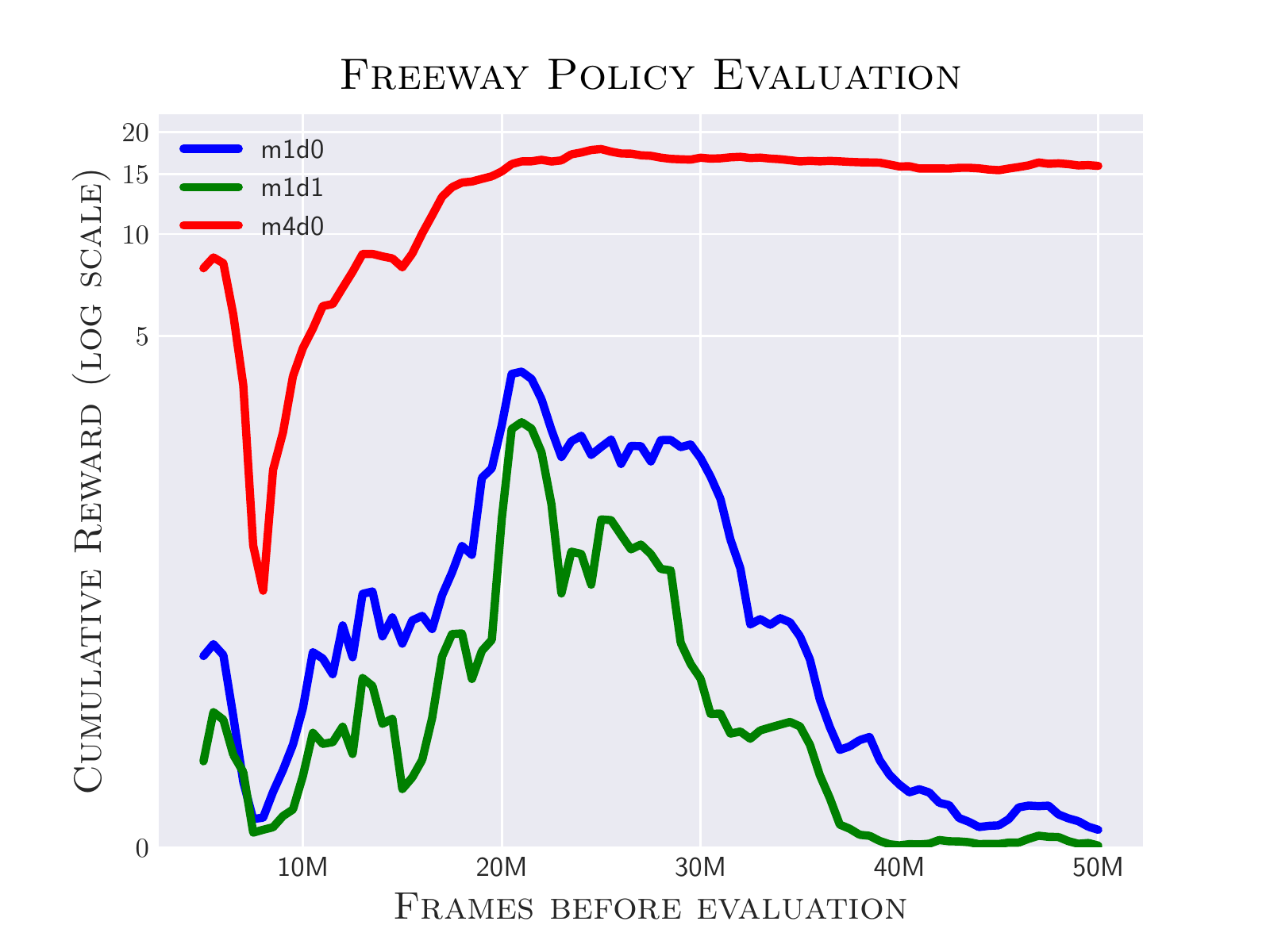}
    \caption{Performance of a trained agent in the default flavour of \textsc{Freeway} and evaluated every $500{,}000$ frames in each target flavour. Error bars were omitted for clarity and the learning curves were smoothed using a moving average over two data points. Results were averaged over five seeds.}
    \label{fig:policytransfer}
\end{figure}

In \textsc{Freeway}, while we see the policy's performance flattening out in m4d0, we do see the traditional bell-shaped curve associated to overfitting in the other modes. At first, improvements in the original policy do correspond to improvements in the performance of that policy in other flavours. With time, it seems that the agent starts to refine its policy for the specific flavour it is being trained on, overfitting to that flavour. With other game flavours being significantly more complex in their dynamics and gameplay, we do not observe this prominent bell-shaped curve.

In conclusion, when looking at Table~\ref{table:dpdp}, it seems that the policies learned by DQN struggle to generalize to even small variations encountered in game flavours.
The results in \textsc{Freeway} even exhibit a troubling notion of overfitting. Nevertheless, being able to generalize across small variations of the task the agent was trained on is a desirable property for truly autonomous agents.
Based on these results we evaluate whether deep RL can benefit from established methods from supervised learning promoting generalization.

\section{Regularization in DQN}

In order to evaluate the hypothesis that the observed lack of generalization is due to overfitting, we revisit some popular regularization methods from the supervised learning literature. We evaluate two forms of regularization: dropout and $\ell_2$ regularization.

First we want to understand the effect of regularization on deploying the learned policy in a different flavour. We do so by applying dropout to the first four layers of the network during training, that is, the three convolutional layers and the first fully connected layer. We also evaluate the use of $\ell_2$ regularization on all weights in the network during training.
A grid search was performed on \textsc{Freeway} to find reasonable hyperparameters for the convolutional and fully connected dropout rate
$p_\text{conv}, p_\text{fc} \in \splitatcommas{\{(0.05,0.1),\ (0.1,0.2),\ (0.15,0.3),\ (0.2,0.4),\ (0.25,0.5)\}}$
, and the $\ell_2$ regularization parameter $\lambda \in \splitatcommas{\{ 10^{-2}, 10^{-3}, 10^{-4}, 10^{-5}, 10^{-6} \}}$. 
Each parameter was swept individually as well as exhausting the cartesian product of both sets of parameters for a total of five runs per configuration. The in-depth ablation study, discussing the impact of different values for each parameter, and their interaction, can be found in the Appendix. We ended up combining dropout and $\ell_2$ regularization as this provided a good balance between training and evaluation performance. This confirms \citeauthor{Srivastava14}'s (\citeyear{Srivastava14}) result that these methods provide benefit in tandem. For all future experiments we use $\lambda = 10^{-4}$, and $p_\text{conv},p_\text{fc} = 0.05, 0.1$.

We follow the same evaluation scheme described when evaluating the non-regularized policy to different flavours. We evaluate the policy learned after 50M frames of the default mode of each game.
We contrast these results with the results presented in the previous section. This evaluation protocol allows us to directly evaluate the effect of regularization on the learned policy's ability to generalize. The results are presented in Table~\ref{table:dpdpreg}, on the next page, and the evaluation curves are available in the Appendix.

\begin{table}[t]
    \centering
    \caption{Policy evaluation using regularization. Each agent was initially trained in the default flavour for 50M frames with dropout and $\ell_2$ regularization then evaluated on each listed flavour. Reported numbers are averaged over five runs. Standard deviation is reported between parentheses.}
    \footnotesize
    \setlength\tabcolsep{4pt} 
    \resizebox{0.75\columnwidth}{!}{
        \begin{tabular}{ll rl rl}

\multicolumn{2}{p{2.5cm}}{\vspace{0mm}\centering\textsc{Game Variant}}
& \multicolumn{2}{p{1.7cm}}{\centering \textsc{Eval. with}\par\textsc{Regularization}}
& \multicolumn{2}{p{1.7cm}}{\centering\textsc{Eval. without}\par\textsc{Regularization}} \\ \midrule[0.4mm]

\multirow{4}{*}{\textsc{Freeway}}
& m1d0
& \textbf{5.8} & \textbf{(3.5)}
& 0.2 & (0.2) \\ \cmidrule(l){2-6}

& m1d1
& \textbf{4.4} & \textbf{(2.3)}
& 0.1 & (0.1) \\ \cmidrule(l){2-6}

& m4d0
& \textbf{20.6} & \textbf{(0.7)}
& 15.8 & (1.0) \\ \cmidrule[0.2mm]{1-6}

\multirow{3}{*}{
    \textsc{Hero}
}
& m1d0
& \textbf{116.8} & \textbf{(76.0)}
& 82.1 & (89.3) \\ \cmidrule(l){2-6}

& m2d0
& 30.0 & (36.7)
& \textbf{33.9} & \textbf{(38.7)} \\ \cmidrule[0.2mm]{1-6}

\multirow{1}{*}{
        \textsc{Breakout}
}

& m12d0
& 31.0 & (8.6)
& \textbf{43.4} & \textbf{(11.1)} \\ \cmidrule[0.2mm]{1-6}

\multirow{4}{*} {
    \textsc{Space Invaders}
}

& m1d0
& \textbf{456.0} & \textbf{(221.4)}
& 258.9 & (88.3) \\ \cmidrule(l){2-6}

& m1d1
& \textbf{146.0} & \textbf{(84.5)}
& 140.4 & (61.4) \\ \cmidrule(l){2-6}

& m9d0
& \textbf{290.0} & \textbf{(257.8)}
& 179.0 & (75.1) \\ \cmidrule[0.2mm]{1-6}
\end{tabular}
    }
    \label{table:dpdpreg}
\end{table}

When using regularization during training we sometimes observe a performance hit in the default flavour. Dropout generally requires increased training iterations to reach the same level of performance one would reach when not using dropout. However, maximal performance in one flavour is not our goal. We are interested in the setting where one may be willing to take lower performance on one task in order to obtain higher performance, or adaptability, on future tasks. Full baseline results using regularization can also be found in Table~\ref{table:baselinesreg}.

\begin{figure}[h!]
\centering
    \includegraphics[width=0.6\columnwidth]{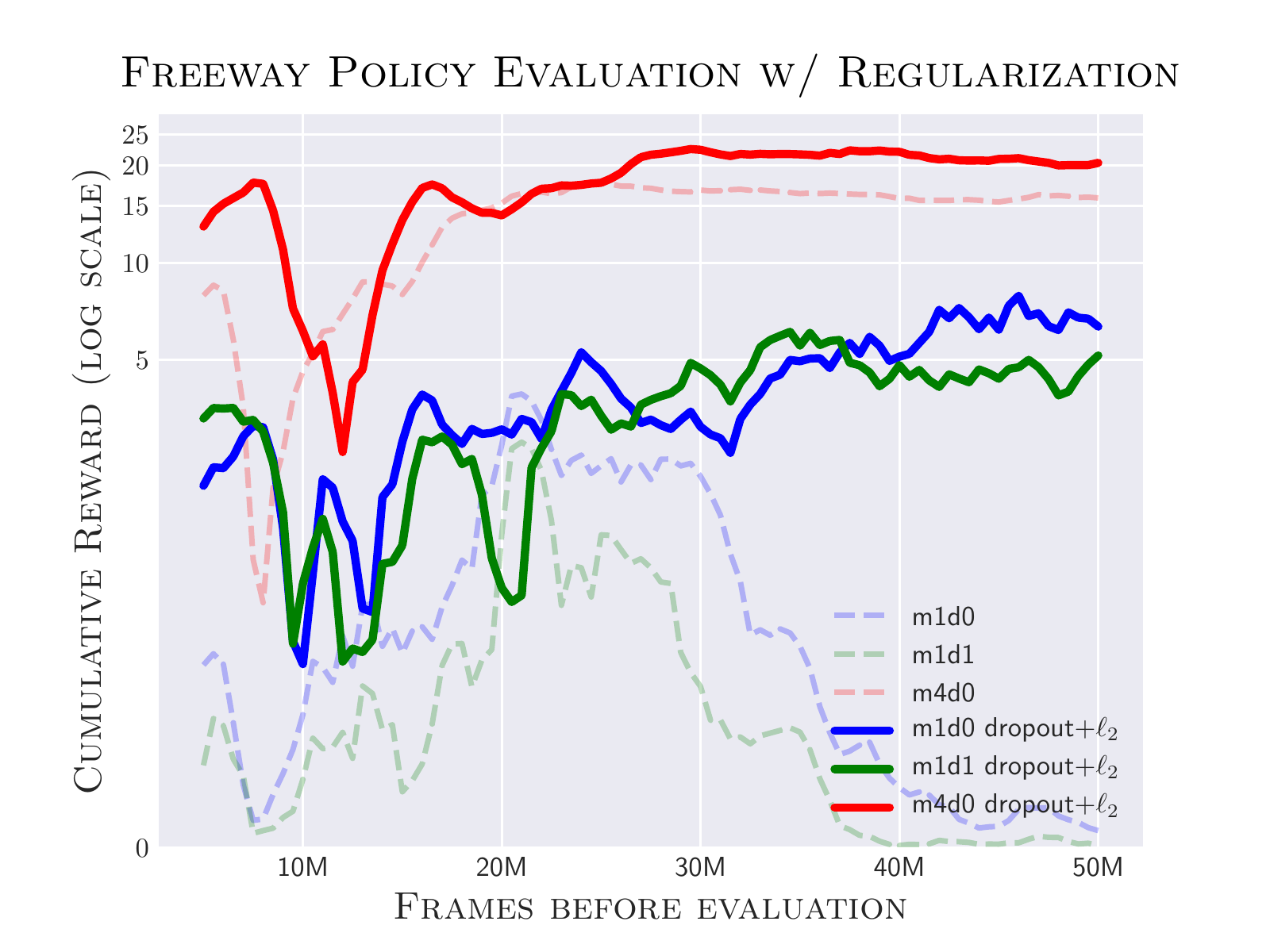}
    \caption{Performance of an agent evaluated every $500,000$ frames after it was trained in the default flavour of \textsc{Freeway} with dropout and $\ell_2$ regularization. Error bars were omitted for clarity and the learning curves were smoothed using a moving average ($n=2$). Results were averaged over five seeds. Dotted lines depict the data presented in Figure~\ref{fig:policytransfer}.}
    \label{fig:policytransferreg}
\end{figure}

In most flavours, when looking at Table~\ref{table:dpdpreg}, we see that evaluating the policy trained with regularization does not negatively impact performance when compared to the performance of the policy trained without regularization. In some flavours we even see an increase in performance. When using regularization the agent's performance in \textsc{Freeway} improves for all flavours and the agent even learns a policy capable of outperforming the baseline learned from scratch in two of the three flavours.
Moreover, in \textsc{Freeway} we now observe increasing performance during evaluation throughout most of the learning procedure as depicted in Figure~\ref{fig:policytransferreg}, on the next page.
These results seem to confirm the notion of overfitting observed in Figure~\ref{fig:policytransfer}.

Despite slight improvements from these techniques, regularization by itself does not seem sufficient to enable policies to generalize across flavours. Learning from scratch in these new flavours is still more beneficial than re-using a policy learned with regularization. As shown in the next section, the real benefit of regularization in deep RL seems to come from the ability to learn more general features. These features lead to a more adaptable representation which can be reused and subsequently fine-tuned on other flavours.

\clearpage

\section{Value function fine-tuning}\label{sec:fine_tuning}

We hypothesize that the benefit of regularizing deep RL algorithms may not come from improvements during evaluation, but instead in having a good parameter initialization that can be adapted to new tasks that are similar.
We evaluate this hypothesis using two common practices in machine learning. First, we use the weights trained with regularization as the initialization for the entire network. We subsequently fine-tune all weights in the network. This is similar to what classification methods do in computer vision problems \citep[e.g.,][]{Razavian14}. Secondly, we evaluate reusing and fine-tuning only early layers of the network. This has been shown to improve generalization in some settings \citep[e.g.,][]{Yosinski14}, and is sometimes used in natural language processing problems \citep[e.g.,][]{Mou16, Howard18}. 
\vspace{-0.15cm}
\subsection{Fine-Tuning the Entire Neural Network}
In this setting we take the weights of the network trained in the default flavour for 50M frames and use them to initialize the network commencing training in the new flavour for 50M frames. We perform this set of experiments twice (for the weights trained with and without regularization, as described in the previous section). Each run is averaged over five seeds. For comparison, we provide a baseline trained from scratch for 50M and 100M frames in each flavour.
Directly comparing the performance obtained after fine-tuning to the performance after 50M frames (\textsc{Scratch}) shows the benefit of re-using a representation learned in a different task instead of randomly initializing the network. Comparing the performance obtained after fine-tuning to the performance of 100M frames (\textsc{Scratch}) lets us take into consideration the sample efficiency of the whole learning process.
The results are presented on the next page, in Table~\ref{table:warmbaseline}.

Fine-tuning from a \emph{non-regularized representation} yields conflicting conclusions.
Although in \textsc{Freeway} we obtained positive fine-tuning results, we note that rewards are so sparse in mode 1 that this initialization is likely to be acting as a form of optimistic initialization, biasing the agent to go up. The agent observes rewards more often, therefore, it learns quicker about the new flavour. However, the agent is still unable to reach the maximum score in these~flavours.

The results of fine-tuning the \emph{regularized representation} are more exciting.
In \textsc{Freeway} we observe the highest scores on m1d0 and m1d1 throughout the whole paper. In \textsc{HERO} we vastly outperform fine-tuning from a non-regularized representation. In \textsc{Space Invaders} we obtain higher scores across the board when comparing to the same amount of experience. These results suggest that reusing a regularized representation in deep RL might allow us to learn more general features which can be more successfully fine-tuned.

\begin{sidewaystable}
    \centering
    \caption{Experiments fine-tuning the entire network with and without regularization (dropout + $\ell_2$). An agent is trained with dropout + $\ell_2$ regularization in the default flavour of each game for 50M frames, then DQN's parameters were used to initialize the fine-tuning procedure on each new flavour for 50M frames. The baseline agent is trained from scratch up to 100M frames. Standard deviation is reported between parentheses.}
    \resizebox{\textwidth}{!}{
        \begin{tabular}{ll rl rl rl rl rl rl}
\multicolumn{2}{c}{}
& \multicolumn{4}{c}{\textsc{Fine-tuning}}
& \multicolumn{4}{c}{\textsc{Regularized Fine-tuning}}
& \multicolumn{4}{c}{\textsc{Scratch}} \\

\cmidrule(lr){3-6}
\cmidrule(lr){7-10}
\cmidrule(lr){11-14} \\

\multicolumn{2}{c}{\textsc{Game Variant}}
& \multicolumn{2}{c}{10M} & \multicolumn{2}{c}{50M}
& \multicolumn{2}{c}{10M} & \multicolumn{2}{c}{50M}
& \multicolumn{2}{c}{50M} &  \multicolumn{2}{c}{100M} \\ \midrule[0.4mm]

\multirow{3}{*}{\textsc{Freeway}}
& m1d0
& 2.9 & (3.7) 
& 22.5 & (7.5) 

& 20.2 & (1.9) 
& \textbf{25.4} & \textbf{(0.2)} 

& 4.8 & (9.3)
& 7.5 & (11.5) \\  \cmidrule(l){2-14}

& m1d1
& 0.1 & (0.2) 
& 17.4 & (11.4) 

& 18.5 & (2.8) 
& \textbf{25.4} & \textbf{(0.4)} 

& 0.0 & (0.0)
& 2.5 & (7.3) \\  \cmidrule(l){2-14}

& m4d0
& 20.8 & (1.1) 
& 31.4 & (0.5) 

& 22.6 & (0.7) 
& 32.2 & (0.5) 

& 29.9 & (0.7)
& \textbf{32.8} & \textbf{(0.2)} \\ \cmidrule[0.2mm]{1-14}

\multirow{2}{*}{\textsc{Hero}}
& m1d0
& 220.7 & (98.2) 
& 496.7 & (362.8) 

& 322.5 & (39.3) 
& 4104.6 & (2192.8) 

& 1425.2 & (1755.1)
& \textbf{5026.8} & \textbf{(2174.6)} \\ \cmidrule(l){2-14}

& m2d0
& 74.4 & (31.7) 
& 92.5 & (26.2) 

& 84.8 & (56.1) 
& 211.0 & (100.6) 

& 326.1 & (130.4)
& \textbf{323.5} & \textbf{(76.4)} \\ \cmidrule[0.2mm]{1-14}

\multirow{1}{*}{\textsc{Breakout}}
& m12d0
& 11.5 & (10.7) 
& 69.1 & (14.9) 

& 48.2 & (4.1) 
& \textbf{96.1} & \textbf{(11.2)} 

& 67.6 & (32.4)
& 55.2 & (37.2) \\ \cmidrule[0.2mm]{1-14}

\multirow{3}{*}{\textsc{Space Invaders}}

& m1d0
& 617.8 & (55.9) 
& 926.1 & (56.6) 

& 701.8 & (28.5) 
& \textbf{1033.5} & \textbf{(89.7)} 

& 753.6 & (31.6)
& 979.7 & (39.8) \\ \cmidrule(l){2-14}

& m1d1
& 482.6 & (63.4) 
& 799.4 & (52.5) 

& 656.7 & (25.5) 
& \textbf{920.0} & \textbf{(83.5)} 

& 698.5 & (31.3)
& 906.9 & (56.5) \\ \cmidrule(l){2-14}

& m9d0
& 354.8 & (59.4) 
& 574.1 & (37.0) 

& 519.0 & (31.1) 
& \textbf{583.0} & \textbf{(17.5)} 

& 518.0 & (16.7)
& 567.7 & (40.1) \\ \cmidrule[0.2mm]{1-14}
\end{tabular}

    }
    \label{table:warmbaseline}
\end{sidewaystable}

\begin{sidewaystable}
    \centering
    \caption{Experiments fine-tuning early layers of the network trained with regularization. An agent is trained with dropout + $\ell_2$ regularization in the default flavour of each game for 50M frames, then DQN's parameters were used to initialize the corresponding layers to be further fine-tuned on each new flavour. Remaining layers were randomly initialized. We also compare against fine-tuning the entire network from Table~\ref{table:warmbaseline}. Standard deviation is reported between parentheses.}
    \resizebox{\textwidth}{!}{
        \begin{tabular}{ll rl rl rl rl rl rl}
\multicolumn{2}{c}{}
& \multicolumn{4}{p{2.2in}}{\centering\textsc{Regularized Fine-Tuning}\par \textsc{3Conv}}
& \multicolumn{4}{p{2.2in}}{\centering \textsc{Regularized Fine-Tuning}\par \textsc{3Conv + 1FC}}
& \multicolumn{4}{p{2.2in}}{\centering\textsc{Regularized Fine-Tuning}\par \textsc{Entire Network}} \\

\cmidrule(lr){3-6}
\cmidrule(lr){7-10}
\cmidrule(lr){11-14} \\

\multicolumn{2}{c}{\textsc{Game Variant}}
& \multicolumn{2}{p{0.8in}}{\centering 10M}
& \multicolumn{2}{p{0.95in}}{\centering 50M}

& \multicolumn{2}{p{0.8in}}{\centering 10M}
& \multicolumn{2}{p{0.95in}}{\centering 50M}

& \multicolumn{2}{p{0.8in}}{\centering 10M}
& \multicolumn{2}{p{0.95in}}{\centering 50M} \\ \midrule[0.4mm]

\multirow{3}{*}{\textsc{Freeway}}
& m1d0
& 0.0 & (0.0) 
& 0.7 & (1.4) 

& 0.1 & (0.1) 
& 4.9 & (9.9) 

& 20.2 & (1.9)
& \textbf{25.4} & \textbf{(0.2)} \\  \cmidrule(l){2-14}

& m1d1
& 0.0 & (0.0) 
& 0.0 & (0.0) 

& 0.1 & (0.1) 
& 10.0 & (12.3) 

& 18.5 & (2.8)
& \textbf{25.4} & \textbf{(0.4)} \\  \cmidrule(l){2-14}

& m4d0
& 7.3 & (3.5) 
& 30.4 & (0.6) 

& 4.9 & (4.8) 
& 30.7 & (1.7) 

& 22.6 & (0.7)
& \textbf{32.2} & \textbf{(0.5)} \\ \cmidrule[0.2mm]{1-14}

\multirow{2}{*}{\textsc{Hero}}
& m1d0
& 405.1 & (82.0) 
& 1949.1 & (2076.4) 

& 350.3 & (52.1) 
& 3085.3 & (2055.6) 

& 322.5 & (39.3)
& \textbf{4104.6} & \textbf{(2192.8)} \\ \cmidrule(l){2-14}

& m2d0
& 232.1 & (30.1) 
& \textbf{455.2} & \textbf{(170.4)} 

& 150.4 & (38.5) 
& 307.6 & (64.8) 

& 84.8 & (56.1)
& 211.0 & (100.6) \\ \cmidrule[0.2mm]{1-14}

\multirow{1}{*}{\textsc{Breakout}}
& m12d0
& 4.3 & (1.7) 
& 63.7 & (26.6) 

& 5.4 & (0.8) 
& 89.1 & (16.7) 

& 48.2 & (4.1)
& \textbf{96.1} & \textbf{(11.2)} \\ \cmidrule[0.2mm]{1-14}

\multirow{3}{*}{\textsc{Space Invaders}}

& m1d0
& 669.3 & (29.1) 
& 998.1 & (78.8) 

& 681.3 & (17.2) 
& 989.6 & (39.4) 

& 701.8 & (28.5)
& \textbf{1033.5} & \textbf{(89.7)} \\ \cmidrule(l){2-14}

& m1d1
& 609.8 & (16.6) 
& 836.3 & (55.9) 

& 638.7 & (19.1) 
& 883.4 & (38.1) 

& 656.7 & (25.5)
& \textbf{920.0} & \textbf{(83.5)} \\ \cmidrule(l){2-14}

& m9d0
& 436.1 & (18.9) 
& 581.0 & (12.2) 

& 439.9 & (40.3) 
& \textbf{586.7} & \textbf{(39.7)} 

& 519.0 & (31.1)
& 583.0 & (17.5) \\ \cmidrule[0.2mm]{1-14}
\end{tabular}

    }
    \label{table:layer_comparison}
\end{sidewaystable}

Initializing the network with a regularized representation also seems to be better than initializing the network randomly, that is, when learning from scratch. These results are impressive when we consider the potential regularization has in reducing the sample complexity of deep RL algorithms. Initializing the network with a regularized representation seems even better than learning from scratch when we take the total number of frames seen between two flavours into consideration. When we look at the rows \textsc{Regularized Fine-tuning} and \textsc{Scratch} in Table~\ref{table:warmbaseline} we are comparing two algorithms that observed 100M frames. However, to generate the results in the column \textsc{Scratch} for two flavours we used 200M frames while we only used used 150M frames to generate the results in the column  \textsc{Regularized Fine-tuning} (50M frames are used to learn in the default flavour and then 50M frames are used in each flavour you actually care about). Obviously, this distinction becomes larger as more tasks are taken into~consideration.

\subsection{Fine-Tuning Early Layers to Learn Co-Adaptations}
We also investigated which layers may encode general features able to be fine-tuned. We were inspired by other studies showing that neural networks can re-learn co-adaptations when their final layers are randomly initialized, sometimes improving generalization \citep{Yosinski14}. We conjectured DQN may benefit from re-learning the co-adaptations between early layers comprising general features and the randomly initialized layers which ultimately assign state-action values. We hypothesized that it might be beneficial to re-learn the final layers from scratch since state-action values are ultimately conditioned on the flavour at hand. Therefore, we also evaluated whether fine-tuning only the convolutional layers, or the convolutional layers and the first fully connected layer, was more effective than fine-tuning the whole network. This does not seem to be the case. The performance when we fine-tune the whole network is consistently better than when we re-learn co-adaptations, as shown in Table~\ref{table:layer_comparison}.

\section{Discussion and conclusion}

Many studies have tried to explain generalization of deep neural networks in supervised learning settings \citep[e.g.,][]{Zhang18, Dinh17}. Analyzing generalization and overfitting in deep RL has its own issues on top of the challenges posed in the supervised learning case.
Actually, generalization in RL can be seen in different ways.
We can talk about generalization in RL in terms of conditioned sub-goals within an environment \citep[e.g.,][]{Andrychowicz17, Sutton95}, learning multiple tasks at once \citep[e.g.,][]{Teh17, Parisotto16}, or sequential task learning as in a continual learning setting \citep[e.g.,][]{Schwarz18, Kirkpatrick16}.
In this paper we evaluated generalization in terms of small variations of high-dimensional control tasks. This provides a candid evaluation method to study how well features and policies learned by deep neural networks in RL problems can generalize. The approach of studying generalization with respect to the representation learning problem intersects nicely with the aforementioned problems in RL where generalization is key.

The results presented in this paper suggest that DQN generalizes poorly, even when tasks have very similar underlying dynamics. Given this lack of generality, we investigated whether dropout and $\ell_2$ regularization can improve generalization in deep reinforcement learning. Other forms of regularization that have been explored in the past are sticky-actions, random initial states, entropy regularization~\citep[e.g.,][]{Zhang18}, and procedural generation of environments \citep[e.g.,][]{Justesen18}. More related to our work, regularization in the form of weight constraints has been applied in the continual learning setting in order to reduce the catastrophic forgetting exhibited by fine-tuning on many sequential tasks \citep{Kirkpatrick16, Schwarz18}.
Similar weight constraint methods were explored in multitask learning \citep{Teh17}.

Evaluation practices in RL often focus on training and evaluating agents on exactly the same task. Consequently, regularization has traditionally been underutilized in deep RL. With a renewed emphasis on generalization in RL, regularization applied to the representation learning problem can be a feasible method for improving generalization on closely related tasks.
Our results suggest that dropout and $\ell_2$ regularization seem to be able to learn more general purpose features which can be adapted to similar problems. Although other communities relying on deep neural networks have shown similar successes, this is of particular importance for the deep RL community which struggles with sample efficiency \citep{Henderson18}. This work is also related to recent meta-learning procedures like MAML \citep{Finn17} which aim to find a parameter initialization that can be quickly adapted to new tasks.
As previously mentioned, techniques such as MAML \citep{Finn17} and REPTILE \citep{Nichol18b} did not succeed in the setting we used.

\begin{table*}[t]
    \centering
    \caption{DQN baseline results for each tested game flavour. We report the average over five runs (std. deviations are reported between parentheses). Results were obtained with the default value of sticky actions \citep{Machado18a}.}
    \footnotesize
    \resizebox*{\columnwidth}{!}{
    \begin{tabular}{ll  rl rl  rl rl}

\multicolumn{2}{c}{\textsc{Game Variant}}
& \multicolumn{2}{c}{\textsc{10M}}
& \multicolumn{2}{c}{\textsc{50M}}
& \multicolumn{2}{c}{\textsc{100M}}
& \multicolumn{2}{c}{\textsc{Best Action}} \\  \midrule[0.4mm]

\multirow{5}{*}{\rotatebox{90}{\textsc{Freeway}}}

& m0d0
& $3.0$ & ($1.0$)
& $31.4$ & ($0.2$)
& $32.1$ & ($0.1$)
& $23.0$ & ($1.4$) \\ \cmidrule(l){2-10}

& m1d0
& $0.0$ & ($0.1$)
& $4.8$ & ($9.3$)
& $7.5$ & ($11.5$)
& $5.0$ & ($1.5$) \\  \cmidrule(l){2-10}
                                                     
& m1d1
& $0.0$ & ($0.0$)
& $0.0$ & ($0.0$)
& $2.5$ & ($7.3$)
& $4.2$ & ($1.3$) \\  \cmidrule(l){2-10}

& m4d0
& $4.4$ & ($1.4$)
& $29.9$ & ($0.7$)
& $32.8$ & ($0.2$)
& $7.5$ & ($2.8$) \\ \cmidrule[0.2mm]{1-10}

\multirow{4}{*}{\rotatebox{90}{\textsc{Hero}}}

& m0d0
& $3187.8$ & ($78.3$)
& $9034.4$ & ($1610.9$)
& $13961.0$ & ($181.9$)
& $150.0$ & ($0.0$)  \\ \cmidrule(l){2-10}

& m1d0
& $326.9$ & ($40.3$)
& $1425.2$ & ($1755.1$)
& $5026.8$ & ($2174.6$)
& $75.8$ & ($7.5$)  \\ \cmidrule(l){2-10}

& m2d0
& $116.3$ & ($11.0$)
& $326.1$ & ($130.4$)
& $323.5$ & ($76.4$)
& $12.0$ & ($27.5$) \\ \cmidrule[0.2mm]{1-10}
    
\multirow{4}{*}{\rotatebox{90}{\textsc{Breakout }}}

& \multirow{2}{*}{m0d0}
& \multirow{2}{*}{$17.5$} & \multirow{2}{*}{($2.0$)}
& \multirow{2}{*}{$72.5$} & \multirow{2}{*}{($7.7$)}
& \multirow{2}{*}{$73.4$} & \multirow{2}{*}{($13.5$)}
& \multirow{2}{*}{$2.3$} & \multirow{2}{*}{($1.3$)} \\ & \\ \cmidrule(l){2-10}

& \multirow{2}{*}{m12d0}
& \multirow{2}{*}{$17.7$} & \multirow{2}{*}{($1.3$)}
& \multirow{2}{*}{$67.6$} & \multirow{2}{*}{($32.4$)}
& \multirow{2}{*}{$55.2$} & \multirow{2}{*}{($37.2$)}
& \multirow{2}{*}{$1.8$} & \multirow{2}{*}{($1.1$)} \\ & \\ \cmidrule[0.2mm]{1-10}

\multirow{9}{*}{\rotatebox{90}{\textsc{Space Invaders}}}

& \multirow{2}{*}{m0d0}
& \multirow{2}{*}{$250.3$} & \multirow{2}{*}{($16.2$)}
& \multirow{2}{*}{$698.8$} & \multirow{2}{*}{($32.2$)}
& \multirow{2}{*}{$927.1$} & \multirow{2}{*}{($85.3$)}
& \multirow{2}{*}{$243.6$} & \multirow{2}{*}{($95.9$)} \cr \\ \cmidrule(l){2-10}

& \multirow{2}{*}{m1d0}
& \multirow{2}{*}{$203.6$} & \multirow{2}{*}{($24.3$)}
& \multirow{2}{*}{$753.6$} & \multirow{2}{*}{($31.6$)}
& \multirow{2}{*}{$979.7$} & \multirow{2}{*}{($39.8$)}
& \multirow{2}{*}{$192.6$} & \multirow{2}{*}{($65.7$)} \cr \\ \cmidrule(l){2-10}

& \multirow{2}{*}{m1d1}
& \multirow{2}{*}{$193.6$} & \multirow{2}{*}{($11.0$)}
& \multirow{2}{*}{$698.5$} & \multirow{2}{*}{($31.3$)}
& \multirow{2}{*}{$906.9$} & \multirow{2}{*}{($56.5$)}
& \multirow{2}{*}{$180.9$} & \multirow{2}{*}{($101.9$)} \cr \\ \cmidrule(l){2-10}

& \multirow{2}{*}{m9d0}
& \multirow{2}{*}{$173.0$} & \multirow{2}{*}{($17.8$)}
& \multirow{2}{*}{$518.0$} & \multirow{2}{*}{($16.7$)}
& \multirow{2}{*}{$567.7$} & \multirow{2}{*}{($40.1$)}
& \multirow{2}{*}{$174.6$} & \multirow{2}{*}{($65.9$)} \cr \\ \cmidrule[0.2mm]{1-10}
\end{tabular}
}
    \label{table:baselines}
\end{table*}

\begin{table*}[h!]
    \centering
    \caption{Baseline results in the default flavour with dropout and $\ell_2$ regularization. We report the average over five runs (std. deviations are reported between parentheses). We used the default value of sticky actions~\citep{Machado18a}.}
    \footnotesize
    \resizebox{\textwidth}{!}{
        \begin{tabular}{ll  rl rl  rl rl}

\multicolumn{2}{c}{\textsc{Game Variant}}
& \multicolumn{2}{c}{\textsc{10M}}
& \multicolumn{2}{c}{\textsc{50M}}
& \multicolumn{2}{c}{\textsc{100M}}
& \multicolumn{2}{c}{\textsc{Best Action}} \\ \midrule[0.4mm]

\multirow{1}{*}{\textsc{Freeway}}

& m0d0
& $4.6$ & ($5.0$)
& $25.9$ & ($0.6$)
& $29.0$ & ($0.8$)
& $23.0$ & ($1.4$) \\ \cmidrule[0.2mm]{1-10}

\multirow{1}{*}{\textsc{Hero}}

& m0d0
& $2466.5$ & ($630.8$)
& $6505.9$ & ($1843.0$)
& $12446.9$ & ($397.4$)
& $150.0$ & ($0.0$)  \\ \cmidrule[0.2mm]{1-10}

\multirow{1}{*}{\textsc{Breakout}}

& m0d0
& $6.1$ & ($2.7$)
& $34.1$ & ($1.8$)
& $66.4$ & ($3.6$)
& $2.3$ & ($1.3$) \\ \cmidrule[0.2mm]{1-10}

\multirow{1}{*}{\textsc{Space Invaders}}

& m0d0
& $214.6$ & ($13.8$)
& $623.1$ & ($16.3$)
& $617.4$ & ($29.6$)
& $243.6$ & ($95.9$) \\ \cmidrule[0.2mm]{1-10}

\end{tabular}

    }
    \label{table:baselinesreg}
\end{table*}

Some of the results here can also be seen under the light of curriculum learning. The regularization techniques we have evaluated here seem to be effective in leveraging situations where an easier task is presented first, sometimes leading to unseen performance levels (e.g., \textsc{Freeway}).

Finally, it is obvious that we want algorithms that can generalize across tasks. Ultimately we want agents that can keep learning as they interact with the world in a continual learning fashion. We believe the flavours of Atari 2600 games can be a stepping stone towards this goal. Our results suggested that regularizing and fine-tuning representations in deep RL might be a viable approach towards improving sample efficiency and generalization on multiple tasks. It is particularly interesting that fine-tuning a regularized network was the most successful approach because this might also be applicable in the continual learning settings where the environment changes without the agent being told so, and re-initializing layers of a network is obviously not an option.

\section*{Acknowledgments}

The authors would like to thank Matthew E. Taylor, Tom van de Wiele, and Marc G. Bellemare for useful discussions, as well as Vlad Mnih for feedback on a preliminary draft of the manuscript. This work was supported by funding from NSERC and Alberta Innovates Technology Futures through the Alberta Machine Intelligence Institute (Amii). Computing resources were provided by Compute Canada through CalculQu\'ebec. Marlos C. Machado performed part of this work while at the University of Alberta.

\bibliographystyle{ACM-Reference-Format}  
\bibliography{refs}  


\begin{thebibliography}{36}


\ifx \showCODEN    \undefined \def \showCODEN     #1{\unskip}     \fi
\ifx \showDOI      \undefined \def \showDOI       #1{#1}\fi
\ifx \showISBNx    \undefined \def \showISBNx     #1{\unskip}     \fi
\ifx \showISBNxiii \undefined \def \showISBNxiii  #1{\unskip}     \fi
\ifx \showISSN     \undefined \def \showISSN      #1{\unskip}     \fi
\ifx \showLCCN     \undefined \def \showLCCN      #1{\unskip}     \fi
\ifx \shownote     \undefined \def \shownote      #1{#1}          \fi
\ifx \showarticletitle \undefined \def \showarticletitle #1{#1}   \fi
\ifx \showURL      \undefined \def \showURL       {\relax}        \fi
\providecommand\bibfield[2]{#2}
\providecommand\bibinfo[2]{#2}
\providecommand\natexlab[1]{#1}
\providecommand\showeprint[2][]{arXiv:#2}

\bibitem[\protect\citeauthoryear{Andrychowicz, Crow, Ray, Schneider, Fong,
  Welinder, McGrew, Tobin, Abbeel, and Zaremba}{Andrychowicz
  et~al\mbox{.}}{2017}]%
        {Andrychowicz17}
\bibfield{author}{\bibinfo{person}{Marcin Andrychowicz},
  \bibinfo{person}{Dwight Crow}, \bibinfo{person}{Alex Ray},
  \bibinfo{person}{Jonas Schneider}, \bibinfo{person}{Rachel Fong},
  \bibinfo{person}{Peter Welinder}, \bibinfo{person}{Bob McGrew},
  \bibinfo{person}{Josh Tobin}, \bibinfo{person}{Pieter Abbeel}, {and}
  \bibinfo{person}{Wojciech Zaremba}.} \bibinfo{year}{2017}\natexlab{}.
\newblock \showarticletitle{{Hindsight Experience Replay}}. In
  \bibinfo{booktitle}{\emph{Advances in Neural Information Processing Systems
  (NeurIPS)}}. \bibinfo{pages}{5048--5058}.
\newblock


\bibitem[\protect\citeauthoryear{Bellemare, Naddaf, Veness, and
  Bowling}{Bellemare et~al\mbox{.}}{2013}]%
        {Bellemare13}
\bibfield{author}{\bibinfo{person}{Marc~G. Bellemare}, \bibinfo{person}{Yavar
  Naddaf}, \bibinfo{person}{Joel Veness}, {and} \bibinfo{person}{Michael
  Bowling}.} \bibinfo{year}{2013}\natexlab{}.
\newblock \showarticletitle{{The Arcade Learning Environment: An Evaluation
  Platform for General Agents}}.
\newblock \bibinfo{journal}{\emph{Journal of Artificial Intelligence Research}}
   \bibinfo{volume}{47} (\bibinfo{year}{2013}), \bibinfo{pages}{253--279}.
\newblock


\bibitem[\protect\citeauthoryear{Cobbe, Klimov, Hesse, Kim, and Schulman}{Cobbe
  et~al\mbox{.}}{2019}]%
        {Cobbe19}
\bibfield{author}{\bibinfo{person}{Karl Cobbe}, \bibinfo{person}{Oleg Klimov},
  \bibinfo{person}{Christopher Hesse}, \bibinfo{person}{Taehoon Kim}, {and}
  \bibinfo{person}{John Schulman}.} \bibinfo{year}{2019}\natexlab{}.
\newblock \showarticletitle{{Quantifying Generalization in Reinforcement
  Learning}}. In \bibinfo{booktitle}{\emph{Proceedings of the International
  Conference on Machine Learning (ICML)}}. \bibinfo{pages}{1282--1289}.
\newblock


\bibitem[\protect\citeauthoryear{Dinh, Pascanu, Bengio, and Bengio}{Dinh
  et~al\mbox{.}}{2017}]%
        {Dinh17}
\bibfield{author}{\bibinfo{person}{Laurent Dinh}, \bibinfo{person}{Razvan
  Pascanu}, \bibinfo{person}{Samy Bengio}, {and} \bibinfo{person}{Yoshua
  Bengio}.} \bibinfo{year}{2017}\natexlab{}.
\newblock \showarticletitle{{Sharp Minima Can Generalize For Deep Nets}}. In
  \bibinfo{booktitle}{\emph{Proceedings of the International Conference on
  Machine Learning (ICML)}}. \bibinfo{pages}{1019--1028}.
\newblock


\bibitem[\protect\citeauthoryear{Espeholt, Soyer, Munos, Simonyan, Mnih, Ward,
  Doron, Firoiu, Harley, Dunning, Legg, and Kavukcuoglu}{Espeholt
  et~al\mbox{.}}{2018}]%
        {Espeholt18}
\bibfield{author}{\bibinfo{person}{Lasse Espeholt}, \bibinfo{person}{Hubert
  Soyer}, \bibinfo{person}{R{\'{e}}mi Munos}, \bibinfo{person}{Karen Simonyan},
  \bibinfo{person}{Volodymyr Mnih}, \bibinfo{person}{Tom Ward},
  \bibinfo{person}{Yotam Doron}, \bibinfo{person}{Vlad Firoiu},
  \bibinfo{person}{Tim Harley}, \bibinfo{person}{Iain Dunning},
  \bibinfo{person}{Shane Legg}, {and} \bibinfo{person}{Koray Kavukcuoglu}.}
  \bibinfo{year}{2018}\natexlab{}.
\newblock \showarticletitle{{{IMPALA:} Scalable Distributed Deep-RL with
  Importance Weighted Actor-Learner Architectures}}. In
  \bibinfo{booktitle}{\emph{Proceedings of the International Conference on
  Machine Learning (ICML)}}. \bibinfo{pages}{1406--1415}.
\newblock


\bibitem[\protect\citeauthoryear{Farahmand, Ghavamzadeh, Szepesv{\'{a}}ri, and
  Mannor}{Farahmand et~al\mbox{.}}{2008}]%
        {Farahmand08}
\bibfield{author}{\bibinfo{person}{Amir~Massoud Farahmand},
  \bibinfo{person}{Mohammad Ghavamzadeh}, \bibinfo{person}{Csaba
  Szepesv{\'{a}}ri}, {and} \bibinfo{person}{Shie Mannor}.}
  \bibinfo{year}{2008}\natexlab{}.
\newblock \showarticletitle{{Regularized Policy Iteration}}. In
  \bibinfo{booktitle}{\emph{Advances in Neural Information Processing Systems
  (NeurIPS)}}. \bibinfo{pages}{441--448}.
\newblock


\bibitem[\protect\citeauthoryear{Finn, Abbeel, and Levine}{Finn
  et~al\mbox{.}}{2017}]%
        {Finn17}
\bibfield{author}{\bibinfo{person}{Chelsea Finn}, \bibinfo{person}{Pieter
  Abbeel}, {and} \bibinfo{person}{Sergey Levine}.}
  \bibinfo{year}{2017}\natexlab{}.
\newblock \showarticletitle{{Model-Agnostic Meta-Learning for Fast Adaptation
  of Deep Networks}}. In \bibinfo{booktitle}{\emph{Proceedings of the
  International Conference on Machine Learning (ICML)}}.
  \bibinfo{pages}{1126--1135}.
\newblock


\bibitem[\protect\citeauthoryear{Glorot and Bengio}{Glorot and Bengio}{2010}]%
        {Glorot10}
\bibfield{author}{\bibinfo{person}{Xavier Glorot} {and} \bibinfo{person}{Yoshua
  Bengio}.} \bibinfo{year}{2010}\natexlab{}.
\newblock \showarticletitle{Understanding the Difficulty of Training Deep
  Feedforward Neural Networks}. In \bibinfo{booktitle}{\emph{Proceedings of the
  International Conference on Artificial Intelligence and Statistics
  (AISTATS)}}. \bibinfo{pages}{249--256}.
\newblock


\bibitem[\protect\citeauthoryear{Henderson, Islam, Bachman, Pineau, Precup, and
  Meger}{Henderson et~al\mbox{.}}{2018}]%
        {Henderson18}
\bibfield{author}{\bibinfo{person}{Peter Henderson}, \bibinfo{person}{Riashat
  Islam}, \bibinfo{person}{Philip Bachman}, \bibinfo{person}{Joelle Pineau},
  \bibinfo{person}{Doina Precup}, {and} \bibinfo{person}{David Meger}.}
  \bibinfo{year}{2018}\natexlab{}.
\newblock \showarticletitle{Deep Reinforcement Learning That Matters}. In
  \bibinfo{booktitle}{\emph{Proceedings of the {AAAI} Conference on Artificial
  Intelligence (AAAI)}}. \bibinfo{pages}{3207--3214}.
\newblock


\bibitem[\protect\citeauthoryear{Howard and Ruder}{Howard and Ruder}{2018}]%
        {Howard18}
\bibfield{author}{\bibinfo{person}{Jeremy Howard} {and}
  \bibinfo{person}{Sebastian Ruder}.} \bibinfo{year}{2018}\natexlab{}.
\newblock \showarticletitle{Fine-tuned Language Models for Text
  Classification}.
\newblock \bibinfo{journal}{\emph{CoRR}}  \bibinfo{volume}{abs/1801.06146}
  (\bibinfo{year}{2018}).
\newblock


\bibitem[\protect\citeauthoryear{Juliani, Khalifa, Berges, Harper, Teng, Henry,
  Crespi, Togelius, and Lange}{Juliani et~al\mbox{.}}{2019}]%
        {Juliani19}
\bibfield{author}{\bibinfo{person}{Arthur Juliani}, \bibinfo{person}{Ahmed
  Khalifa}, \bibinfo{person}{Vincent{-}Pierre Berges},
  \bibinfo{person}{Jonathan Harper}, \bibinfo{person}{Ervin Teng},
  \bibinfo{person}{Hunter Henry}, \bibinfo{person}{Adam Crespi},
  \bibinfo{person}{Julian Togelius}, {and} \bibinfo{person}{Danny Lange}.}
  \bibinfo{year}{2019}\natexlab{}.
\newblock \showarticletitle{{Obstacle Tower: {A} Generalization Challenge in
  Vision, Control, and Planning}}. In \bibinfo{booktitle}{\emph{Proceedings of
  the International Joint Conference on Artificial Intelligence (IJCAI)}}.
  \bibinfo{pages}{2684--2691}.
\newblock


\bibitem[\protect\citeauthoryear{Justesen, Torrado, Bontrager, Khalifa,
  Togelius, and Risi}{Justesen et~al\mbox{.}}{2018}]%
        {Justesen18}
\bibfield{author}{\bibinfo{person}{Niels Justesen},
  \bibinfo{person}{Ruben~Rodriguez Torrado}, \bibinfo{person}{Philip
  Bontrager}, \bibinfo{person}{Ahmed Khalifa}, \bibinfo{person}{Julian
  Togelius}, {and} \bibinfo{person}{Sebastian Risi}.}
  \bibinfo{year}{2018}\natexlab{}.
\newblock \showarticletitle{Procedural Level Generation Improves Generality of
  Deep Reinforcement Learning}.
\newblock \bibinfo{journal}{\emph{CoRR}}  \bibinfo{volume}{abs/1806.10729}
  (\bibinfo{year}{2018}).
\newblock


\bibitem[\protect\citeauthoryear{Kirkpatrick, Pascanu, Rabinowitz, Veness,
  Desjardins, Rusu, Milan, Quan, Ramalho, Grabska{-}Barwinska, Hassabis,
  Clopath, Kumaran, and Hadsell}{Kirkpatrick et~al\mbox{.}}{2016}]%
        {Kirkpatrick16}
\bibfield{author}{\bibinfo{person}{James Kirkpatrick}, \bibinfo{person}{Razvan
  Pascanu}, \bibinfo{person}{Neil~C. Rabinowitz}, \bibinfo{person}{Joel
  Veness}, \bibinfo{person}{Guillaume Desjardins}, \bibinfo{person}{Andrei~A.
  Rusu}, \bibinfo{person}{Kieran Milan}, \bibinfo{person}{John Quan},
  \bibinfo{person}{Tiago Ramalho}, \bibinfo{person}{Agnieszka
  Grabska{-}Barwinska}, \bibinfo{person}{Demis Hassabis},
  \bibinfo{person}{Claudia Clopath}, \bibinfo{person}{Dharshan Kumaran}, {and}
  \bibinfo{person}{Raia Hadsell}.} \bibinfo{year}{2016}\natexlab{}.
\newblock \showarticletitle{{Overcoming Catastrophic Forgetting in Neural
  Networks}}.
\newblock \bibinfo{journal}{\emph{CoRR}}  \bibinfo{volume}{abs/1612.00796}
  (\bibinfo{year}{2016}).
\newblock


\bibitem[\protect\citeauthoryear{Kolter and Ng}{Kolter and Ng}{2009}]%
        {Kolter09}
\bibfield{author}{\bibinfo{person}{J.~Zico Kolter} {and}
  \bibinfo{person}{Andrew~Y. Ng}.} \bibinfo{year}{2009}\natexlab{}.
\newblock \showarticletitle{{Regularization and Feature Selection in
  Least-Squares Temporal Difference Learning}}. In
  \bibinfo{booktitle}{\emph{Proceedings of the International Conference on
  Machine Learning (ICML)}}. \bibinfo{pages}{521--528}.
\newblock


\bibitem[\protect\citeauthoryear{Lecun, Bottou, Bengio, and Haffner}{Lecun
  et~al\mbox{.}}{1998}]%
        {Lecun98}
\bibfield{author}{\bibinfo{person}{Yann Lecun}, \bibinfo{person}{L\'eon
  Bottou}, \bibinfo{person}{Yoshua Bengio}, {and} \bibinfo{person}{Patrick
  Haffner}.} \bibinfo{year}{1998}\natexlab{}.
\newblock \showarticletitle{{Gradient-based Learning Applied to Document
  Recognition}}.
\newblock \bibinfo{journal}{\emph{IEEE}} \bibinfo{volume}{86},
  \bibinfo{number}{11} (\bibinfo{year}{1998}), \bibinfo{pages}{2278--2324}.
\newblock
\showISSN{0018-9219}


\bibitem[\protect\citeauthoryear{Long, Shelhamer, and Darrell}{Long
  et~al\mbox{.}}{2015}]%
        {Long15}
\bibfield{author}{\bibinfo{person}{Jonathan Long}, \bibinfo{person}{Evan
  Shelhamer}, {and} \bibinfo{person}{Trevor Darrell}.}
  \bibinfo{year}{2015}\natexlab{}.
\newblock \showarticletitle{{Fully Convolutional Networks for Semantic
  Segmentation}}. In \bibinfo{booktitle}{\emph{Proceedings of the {IEEE}
  Conference on Computer Vision and Pattern Recognition (CVPR)}}.
  \bibinfo{pages}{3431--3440}.
\newblock


\bibitem[\protect\citeauthoryear{Machado, Bellemare, Talvitie, Veness,
  Hausknecht, and Bowling}{Machado et~al\mbox{.}}{2018}]%
        {Machado18a}
\bibfield{author}{\bibinfo{person}{Marlos~C. Machado}, \bibinfo{person}{Marc~G.
  Bellemare}, \bibinfo{person}{Erik Talvitie}, \bibinfo{person}{Joel Veness},
  \bibinfo{person}{Matthew~J. Hausknecht}, {and} \bibinfo{person}{Michael
  Bowling}.} \bibinfo{year}{2018}\natexlab{}.
\newblock \showarticletitle{{Revisiting the Arcade Learning Environment:
  Evaluation Protocols and Open Problems for General Agents}}.
\newblock \bibinfo{journal}{\emph{Journal of Artificial Intelligence Research}}
   \bibinfo{volume}{61} (\bibinfo{year}{2018}), \bibinfo{pages}{523--562}.
\newblock


\bibitem[\protect\citeauthoryear{Mnih, Kavukcuoglu, Silver, Rusu, Veness,
  Bellemare, Graves, Riedmiller, Fidjeland, Ostrovski, Petersen, Beattie,
  Sadik, Antonoglou, King, Kumaran, Wierstra, Legg, and Hassabis}{Mnih
  et~al\mbox{.}}{2015}]%
        {Mnih15}
\bibfield{author}{\bibinfo{person}{Volodymyr Mnih}, \bibinfo{person}{Koray
  Kavukcuoglu}, \bibinfo{person}{David Silver}, \bibinfo{person}{Andrei~A.
  Rusu}, \bibinfo{person}{Joel Veness}, \bibinfo{person}{Marc~G. Bellemare},
  \bibinfo{person}{Alex Graves}, \bibinfo{person}{Martin~A. Riedmiller},
  \bibinfo{person}{Andreas Fidjeland}, \bibinfo{person}{Georg Ostrovski},
  \bibinfo{person}{Stig Petersen}, \bibinfo{person}{Charles Beattie},
  \bibinfo{person}{Amir Sadik}, \bibinfo{person}{Ioannis Antonoglou},
  \bibinfo{person}{Helen King}, \bibinfo{person}{Dharshan Kumaran},
  \bibinfo{person}{Daan Wierstra}, \bibinfo{person}{Shane Legg}, {and}
  \bibinfo{person}{Demis Hassabis}.} \bibinfo{year}{2015}\natexlab{}.
\newblock \showarticletitle{{Human-Level Control through Deep Reinforcement
  Learning}}.
\newblock \bibinfo{journal}{\emph{Nature}} \bibinfo{volume}{518},
  \bibinfo{number}{7540} (\bibinfo{year}{2015}), \bibinfo{pages}{529--533}.
\newblock


\bibitem[\protect\citeauthoryear{Mou, Meng, Yan, Li, Xu, Zhang, and Jin}{Mou
  et~al\mbox{.}}{2016}]%
        {Mou16}
\bibfield{author}{\bibinfo{person}{Lili Mou}, \bibinfo{person}{Zhao Meng},
  \bibinfo{person}{Rui Yan}, \bibinfo{person}{Ge Li}, \bibinfo{person}{Yan Xu},
  \bibinfo{person}{Lu Zhang}, {and} \bibinfo{person}{Zhi Jin}.}
  \bibinfo{year}{2016}\natexlab{}.
\newblock \showarticletitle{{How Transferable are Neural Networks in {NLP}
  Applications?}}. In \bibinfo{booktitle}{\emph{Proceedings of the Conference
  on Empirical Methods in Natural Language Processing (EMNLP)}}.
  \bibinfo{pages}{479--489}.
\newblock


\bibitem[\protect\citeauthoryear{Nichol, Achiam, and Schulman}{Nichol
  et~al\mbox{.}}{2018a}]%
        {Nichol18a}
\bibfield{author}{\bibinfo{person}{Alex Nichol}, \bibinfo{person}{Joshua
  Achiam}, {and} \bibinfo{person}{John Schulman}.}
  \bibinfo{year}{2018}\natexlab{a}.
\newblock \showarticletitle{On First-Order Meta-Learning Algorithms}.
\newblock \bibinfo{journal}{\emph{CoRR}}  \bibinfo{volume}{abs/1803.02999}
  (\bibinfo{year}{2018}).
\newblock


\bibitem[\protect\citeauthoryear{Nichol, Pfau, Hesse, Klimov, and
  Schulman}{Nichol et~al\mbox{.}}{2018b}]%
        {Nichol18b}
\bibfield{author}{\bibinfo{person}{Alex Nichol}, \bibinfo{person}{Vicki Pfau},
  \bibinfo{person}{Christopher Hesse}, \bibinfo{person}{Oleg Klimov}, {and}
  \bibinfo{person}{John Schulman}.} \bibinfo{year}{2018}\natexlab{b}.
\newblock \showarticletitle{Gotta Learn Fast: {A} New Benchmark for
  Generalization in {RL}}.
\newblock \bibinfo{journal}{\emph{CoRR}}  \bibinfo{volume}{abs/1804.03720}
  (\bibinfo{year}{2018}).
\newblock


\bibitem[\protect\citeauthoryear{Parisotto, Ba, and Salakhutdinov}{Parisotto
  et~al\mbox{.}}{2016}]%
        {Parisotto16}
\bibfield{author}{\bibinfo{person}{Emilio Parisotto},
  \bibinfo{person}{Lei~Jimmy Ba}, {and} \bibinfo{person}{Ruslan
  Salakhutdinov}.} \bibinfo{year}{2016}\natexlab{}.
\newblock \showarticletitle{{Actor-Mimic: Deep Multitask and Transfer
  Reinforcement Learning}}. In \bibinfo{booktitle}{\emph{Proceedings of the
  International Conference on Learning Representations (ICLR)}}.
\newblock


\bibitem[\protect\citeauthoryear{Rajeswaran, Lowrey, Todorov, and
  Kakade}{Rajeswaran et~al\mbox{.}}{2017}]%
        {Rajeswaran17}
\bibfield{author}{\bibinfo{person}{Aravind Rajeswaran},
  \bibinfo{person}{Kendall Lowrey}, \bibinfo{person}{Emanuel Todorov}, {and}
  \bibinfo{person}{Sham~M. Kakade}.} \bibinfo{year}{2017}\natexlab{}.
\newblock \showarticletitle{Towards Generalization and Simplicity in Continuous
  Control}. In \bibinfo{booktitle}{\emph{Advances in Neural Information
  Processing Systems (NeurIPS)}}. \bibinfo{pages}{6550--6561}.
\newblock


\bibitem[\protect\citeauthoryear{Razavian, Azizpour, Sullivan, and
  Carlsson}{Razavian et~al\mbox{.}}{2014}]%
        {Razavian14}
\bibfield{author}{\bibinfo{person}{Ali~Sharif Razavian},
  \bibinfo{person}{Hossein Azizpour}, \bibinfo{person}{Josephine Sullivan},
  {and} \bibinfo{person}{Stefan Carlsson}.} \bibinfo{year}{2014}\natexlab{}.
\newblock \showarticletitle{{CNN} Features Off-the-Shelf: An Astounding
  Baseline for Recognition}. In \bibinfo{booktitle}{\emph{Workshops of the
  {IEEE} Conference on Computer Vision and Pattern Recognition (CVPR)}}.
  \bibinfo{pages}{512--519}.
\newblock


\bibitem[\protect\citeauthoryear{Rusu, Rabinowitz, Desjardins, Soyer,
  Kirkpatrick, Kavukcuoglu, Pascanu, and Hadsell}{Rusu et~al\mbox{.}}{2016}]%
        {Rusu16}
\bibfield{author}{\bibinfo{person}{Andrei~A. Rusu}, \bibinfo{person}{Neil~C.
  Rabinowitz}, \bibinfo{person}{Guillaume Desjardins}, \bibinfo{person}{Hubert
  Soyer}, \bibinfo{person}{James Kirkpatrick}, \bibinfo{person}{Koray
  Kavukcuoglu}, \bibinfo{person}{Razvan Pascanu}, {and} \bibinfo{person}{Raia
  Hadsell}.} \bibinfo{year}{2016}\natexlab{}.
\newblock \showarticletitle{Progressive Neural Networks}.
\newblock \bibinfo{journal}{\emph{CoRR}}  \bibinfo{volume}{abs/1606.04671}
  (\bibinfo{year}{2016}).
\newblock


\bibitem[\protect\citeauthoryear{Schwarz, Czarnecki, Luketina,
  Grabska{-}Barwinska, Teh, Pascanu, and Hadsell}{Schwarz
  et~al\mbox{.}}{2018}]%
        {Schwarz18}
\bibfield{author}{\bibinfo{person}{Jonathan Schwarz}, \bibinfo{person}{Wojciech
  Czarnecki}, \bibinfo{person}{Jelena Luketina}, \bibinfo{person}{Agnieszka
  Grabska{-}Barwinska}, \bibinfo{person}{Yee~Whye Teh}, \bibinfo{person}{Razvan
  Pascanu}, {and} \bibinfo{person}{Raia Hadsell}.}
  \bibinfo{year}{2018}\natexlab{}.
\newblock \showarticletitle{Progress {\&} Compress: {A} Scalable Framework for
  Continual Learning}. In \bibinfo{booktitle}{\emph{Proceedings of the
  International Conference on Machine Learning (ICML)}}.
  \bibinfo{pages}{4535--4544}.
\newblock


\bibitem[\protect\citeauthoryear{Silver, Huang, Maddison, Guez, Sifre, van~den
  Driessche, Schrittwieser, Antonoglou, Panneershelvam, Lanctot, Dieleman,
  Grewe, Nham, Kalchbrenner, Sutskever, Lillicrap, Leach, Kavukcuoglu, Graepel,
  and Hassabis}{Silver et~al\mbox{.}}{2016}]%
        {Silver16}
\bibfield{author}{\bibinfo{person}{David Silver}, \bibinfo{person}{Aja Huang},
  \bibinfo{person}{Chris~J. Maddison}, \bibinfo{person}{Arthur Guez},
  \bibinfo{person}{Laurent Sifre}, \bibinfo{person}{George van~den Driessche},
  \bibinfo{person}{Julian Schrittwieser}, \bibinfo{person}{Ioannis Antonoglou},
  \bibinfo{person}{Vedavyas Panneershelvam}, \bibinfo{person}{Marc Lanctot},
  \bibinfo{person}{Sander Dieleman}, \bibinfo{person}{Dominik Grewe},
  \bibinfo{person}{John Nham}, \bibinfo{person}{Nal Kalchbrenner},
  \bibinfo{person}{Ilya Sutskever}, \bibinfo{person}{Timothy~P. Lillicrap},
  \bibinfo{person}{Madeleine Leach}, \bibinfo{person}{Koray Kavukcuoglu},
  \bibinfo{person}{Thore Graepel}, {and} \bibinfo{person}{Demis Hassabis}.}
  \bibinfo{year}{2016}\natexlab{}.
\newblock \showarticletitle{Mastering the Game of Go with Deep Neural Networks
  and Tree Search}.
\newblock \bibinfo{journal}{\emph{Nature}} \bibinfo{volume}{529},
  \bibinfo{number}{7587} (\bibinfo{year}{2016}), \bibinfo{pages}{484--489}.
\newblock


\bibitem[\protect\citeauthoryear{Srivastava, Hinton, Krizhevsky, Sutskever, and
  Salakhutdinov}{Srivastava et~al\mbox{.}}{2014}]%
        {Srivastava14}
\bibfield{author}{\bibinfo{person}{Nitish Srivastava},
  \bibinfo{person}{Geoffrey~E. Hinton}, \bibinfo{person}{Alex Krizhevsky},
  \bibinfo{person}{Ilya Sutskever}, {and} \bibinfo{person}{Ruslan
  Salakhutdinov}.} \bibinfo{year}{2014}\natexlab{}.
\newblock \showarticletitle{Dropout: a Simple Way to Prevent Neural Networks
  from Overfitting}.
\newblock \bibinfo{journal}{\emph{Journal of Machine Learning Research}}
  \bibinfo{volume}{15}, \bibinfo{number}{1} (\bibinfo{year}{2014}),
  \bibinfo{pages}{1929--1958}.
\newblock


\bibitem[\protect\citeauthoryear{Sutton}{Sutton}{1995}]%
        {Sutton95}
\bibfield{author}{\bibinfo{person}{Richard~S. Sutton}.}
  \bibinfo{year}{1995}\natexlab{}.
\newblock \showarticletitle{Generalization in Reinforcement Learning:
  Successful Examples Using Sparse Coarse Coding}. In
  \bibinfo{booktitle}{\emph{Advances in Neural Information Processing Systems
  (NeurIPS)}}. \bibinfo{pages}{1038--1044}.
\newblock


\bibitem[\protect\citeauthoryear{Teh, Bapst, Czarnecki, Quan, Kirkpatrick,
  Hadsell, Heess, and Pascanu}{Teh et~al\mbox{.}}{2017}]%
        {Teh17}
\bibfield{author}{\bibinfo{person}{Yee~Whye Teh}, \bibinfo{person}{Victor
  Bapst}, \bibinfo{person}{Wojciech~M. Czarnecki}, \bibinfo{person}{John Quan},
  \bibinfo{person}{James Kirkpatrick}, \bibinfo{person}{Raia Hadsell},
  \bibinfo{person}{Nicolas Heess}, {and} \bibinfo{person}{Razvan Pascanu}.}
  \bibinfo{year}{2017}\natexlab{}.
\newblock \showarticletitle{{Distral: Robust Multitask Reinforcement
  Learning}}. In \bibinfo{booktitle}{\emph{Advances in Neural Information
  Processing Systems (NeurIPS)}}. \bibinfo{pages}{4496--4506}.
\newblock


\bibitem[\protect\citeauthoryear{Watkins and Dayan}{Watkins and Dayan}{1992}]%
        {Watkins92}
\bibfield{author}{\bibinfo{person}{Christopher Watkins} {and}
  \bibinfo{person}{Peter Dayan}.} \bibinfo{year}{1992}\natexlab{}.
\newblock \showarticletitle{{Technical Note: \cal Q-Learning}}.
\newblock \bibinfo{journal}{\emph{Machine Learning}} \bibinfo{volume}{8},
  \bibinfo{number}{3-4} (\bibinfo{year}{1992}).
\newblock


\bibitem[\protect\citeauthoryear{Whiteson, Tanner, Taylor, and Stone}{Whiteson
  et~al\mbox{.}}{2011}]%
        {Whiteson11}
\bibfield{author}{\bibinfo{person}{Shimon Whiteson}, \bibinfo{person}{Brian
  Tanner}, \bibinfo{person}{Matthew~E. Taylor}, {and} \bibinfo{person}{Peter
  Stone}.} \bibinfo{year}{2011}\natexlab{}.
\newblock \showarticletitle{Protecting Against Evaluation Overfitting in
  Empirical Reinforcement Learning}. In \bibinfo{booktitle}{\emph{{IEEE}
  Symposium on Adaptive Dynamic Programming And Reinforcement Learning
  (ADPRL)}}. \bibinfo{pages}{120--127}.
\newblock


\bibitem[\protect\citeauthoryear{Witty, Lee, Tosch, Atrey, Littman, and
  Jensen}{Witty et~al\mbox{.}}{2018}]%
        {Witty18}
\bibfield{author}{\bibinfo{person}{Sam Witty}, \bibinfo{person}{Jun~Ki Lee},
  \bibinfo{person}{Emma Tosch}, \bibinfo{person}{Akanksha Atrey},
  \bibinfo{person}{Michael~L. Littman}, {and} \bibinfo{person}{David~D.
  Jensen}.} \bibinfo{year}{2018}\natexlab{}.
\newblock \showarticletitle{{Measuring and Characterizing Generalization in
  Deep Reinforcement Learning}}.
\newblock \bibinfo{journal}{\emph{CoRR}}  \bibinfo{volume}{abs/1812.02868}
  (\bibinfo{year}{2018}).
\newblock


\bibitem[\protect\citeauthoryear{Yosinski, Clune, Bengio, and Lipson}{Yosinski
  et~al\mbox{.}}{2014}]%
        {Yosinski14}
\bibfield{author}{\bibinfo{person}{Jason Yosinski}, \bibinfo{person}{Jeff
  Clune}, \bibinfo{person}{Yoshua Bengio}, {and} \bibinfo{person}{Hod Lipson}.}
  \bibinfo{year}{2014}\natexlab{}.
\newblock \showarticletitle{How Transferable are Features in Deep Neural
  Networks?}. In \bibinfo{booktitle}{\emph{Advances in Neural Information
  Processing Systems (NeurIPS)}}. \bibinfo{pages}{3320--3328}.
\newblock


\bibitem[\protect\citeauthoryear{Zhang, Ballas, and Pineau}{Zhang
  et~al\mbox{.}}{2018a}]%
        {Zhang18_2}
\bibfield{author}{\bibinfo{person}{Amy Zhang}, \bibinfo{person}{Nicolas
  Ballas}, {and} \bibinfo{person}{Joelle Pineau}.}
  \bibinfo{year}{2018}\natexlab{a}.
\newblock \showarticletitle{{A Dissection of Overfitting and Generalization in
  Continuous Reinforcement Learning}}.
\newblock \bibinfo{journal}{\emph{CoRR}}  \bibinfo{volume}{abs/1806.07937}
  (\bibinfo{year}{2018}).
\newblock


\bibitem[\protect\citeauthoryear{Zhang, Vinyals, Munos, and Bengio}{Zhang
  et~al\mbox{.}}{2018b}]%
        {Zhang18}
\bibfield{author}{\bibinfo{person}{Chiyuan Zhang}, \bibinfo{person}{Oriol
  Vinyals}, \bibinfo{person}{R{\'{e}}mi Munos}, {and} \bibinfo{person}{Samy
  Bengio}.} \bibinfo{year}{2018}\natexlab{b}.
\newblock \showarticletitle{A Study on Overfitting in Deep Reinforcement
  Learning}.
\newblock \bibinfo{journal}{\emph{CoRR}}  \bibinfo{volume}{abs/1804.06893}
  (\bibinfo{year}{2018}).
\newblock


\end{thebibliography}

\clearpage

\section*{Appendix}
\subsection*{Game Modes}
\label{appendix:modes}

\let\thefootnote\relax\footnotetext{Videos of the different modes are available in the following link: \url{https://goo.gl/pCvPiD}.}
We provide a brief description of each game flavour used in the paper. 

\subsubsection*{\textsc{Freeway}}

\begin{figure*}[h]
    \centering
    \def\arraystretch{1.2}
    \setlength\tabcolsep{2pt}
    \begin{tabular}{ccc}
        \textsc{Freeway} m0d0 & \textsc{Freeway} m1d0 & \textsc{Freeway} m4d0 \\
        \includegraphics[width=.31\textwidth]{figures/alemodes/freeway_m0.png}
        & \includegraphics[width=.31\textwidth]{figures/alemodes/freeway_m1.png}
        & \includegraphics[width=.31\textwidth]{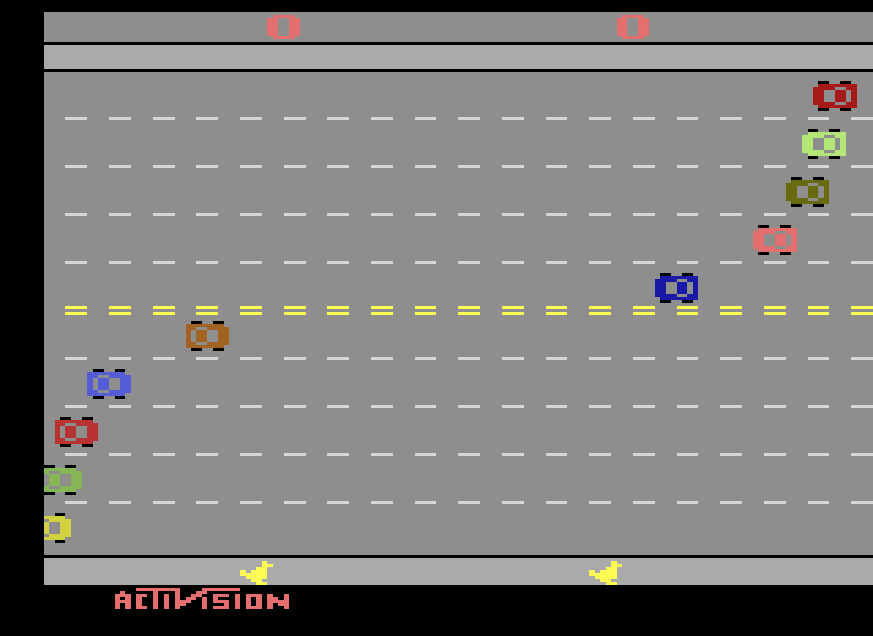}
    \end{tabular}
    \label{fig:freeway_modes}
\end{figure*}

In \textsc{Freeway} a chicken must cross a road containing multiple lanes of moving traffic within a prespecified time limit. In all modes of \textsc{Freeway} the agent is rewarded for reaching the top of the screen and is subsequently teleported to the bottom of the screen.
If the chicken collides with a vehicle in difficulty 0 it gets bumped down one lane of traffic, alternatively, in difficulty 1 the chicken gets teleported to its starting position at the bottom of the screen.
Mode 1 changes some vehicle sprites to include buses, adds more vehicles to some lanes, and increases the velocity of all vehicles.
Mode 4 is almost identical to Mode 1; the only difference being vehicles can oscillate between two speeds.

\subsubsection*{\textsc{Hero}}

\begin{figure*}[h]
    \centering
    \def\arraystretch{1.2}
    \setlength\tabcolsep{2pt}
    \begin{tabular}{ccc}
        \textsc{Hero} m0d0 & \textsc{Hero} m1d0 & \textsc{Hero} m2d0 \\
        \includegraphics[width=.31\textwidth]{figures/alemodes/hero_m0.png}
        & \includegraphics[width=.31\textwidth]{figures/alemodes/hero_m1.png}
        & \includegraphics[width=.31\textwidth]{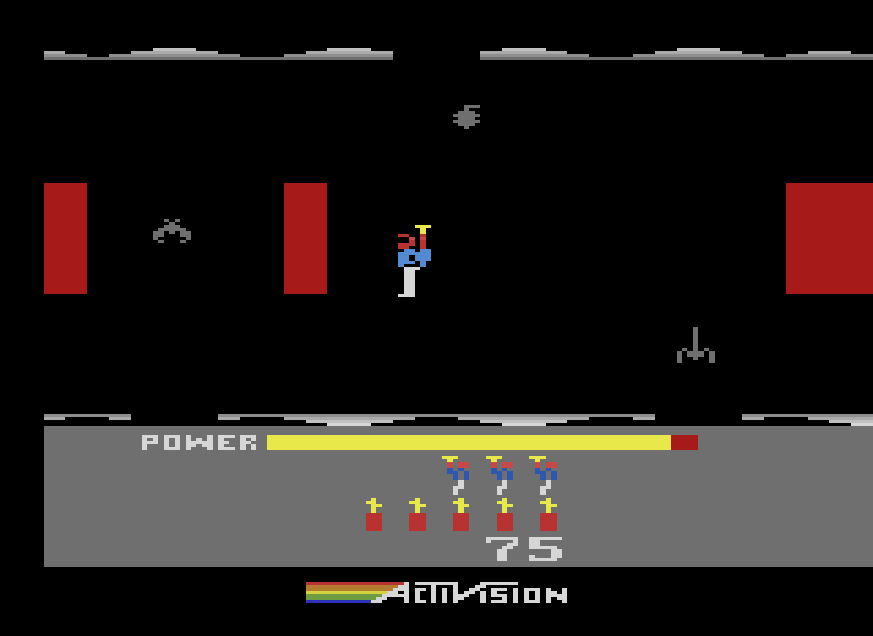}
    \end{tabular}
    \label{fig:hero_modes}
\end{figure*}

In \textsc{Hero} you control a character who must navigate a maze in order to save a trapped miner within a cave system. 
The agent scores points for any forward progression such as clearing an obstacle or killing an enemy.
Once the miner is rescued, the level is terminated and you continue to the next level with a different maze. Some levels have partially observable rooms, more enemies, and more difficult obstacles to traverse. Past the default mode, each subsequent mode starts off at increasingly harder levels denoted by a level number increasing by multiples of $5$. The default mode starts you off at level $1$, mode 1 starts at level $5$, and so on.

\subsubsection*{\textsc{Breakout}}

\begin{figure*}[h]
    \centering
    \def\arraystretch{1.2}
    \setlength\tabcolsep{2pt}
    \begin{tabular}{cc}
        \textsc{Breakout} m0d0 & \textsc{Breakout} m12d0 \\
        \includegraphics[width=.31\textwidth]{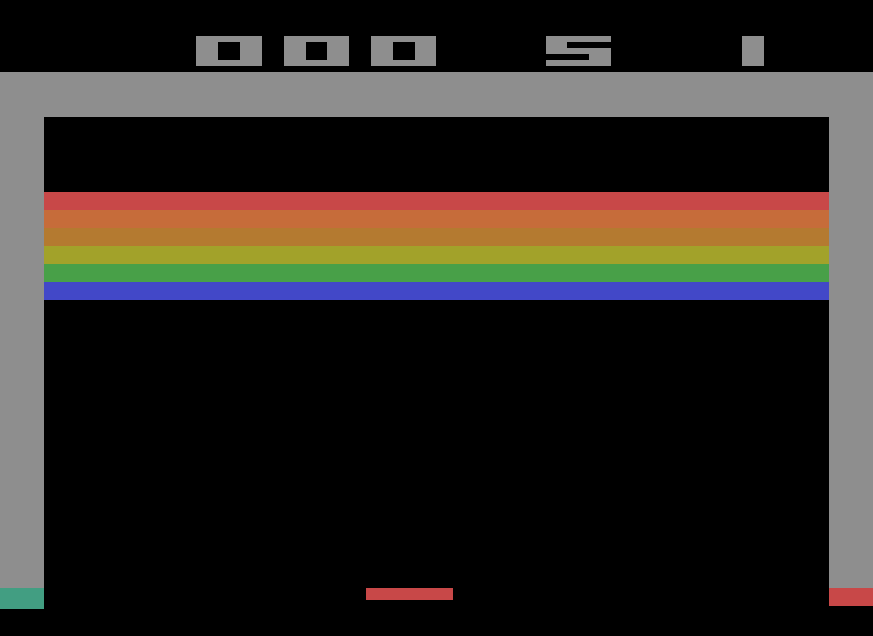}
        & \includegraphics[width=.31\textwidth]{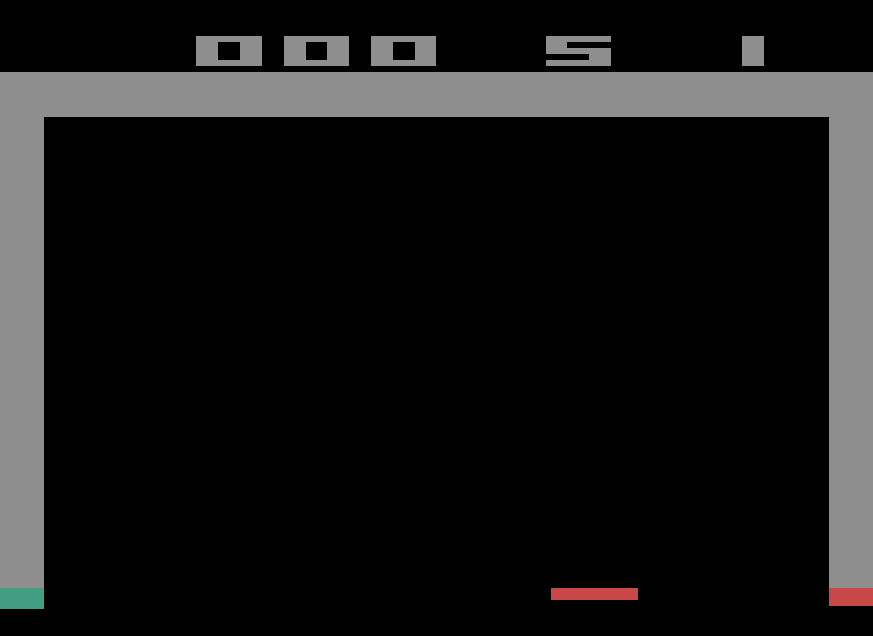}
    \end{tabular}
    \label{fig:breakout_modes}
\end{figure*}

In \textsc{Breakout} you control a paddle which can move horizontally along the bottom of the screen. At the beginning of the game, or on a loss of life the ball is set into motion and can bounce off the paddle and collide with bricks at the top of the screen. The objective of the game is to break all the bricks without having the ball fall below your paddles horizontal plane.
Subsequently, mode 12 of \textsc{Breakout} hides the bricks from the player until the ball collides with the bricks in which case the bricks flash for a brief moment before disappearing again.

\subsubsection*{\textsc{Space Invaders}}

\begin{figure*}[h]
    \centering
    \def\arraystretch{1.2}
    \setlength\tabcolsep{2pt}
    \begin{tabular}{ccc}
        \textsc{Space Invaders} m0d0 & \textsc{Space Invaders} m1d1 & \textsc{Space Invaders} m9d0\\
        \includegraphics[width=.31\textwidth]{figures/alemodes/space_invaders_m0.png}
        & \includegraphics[width=.31\textwidth]{figures/alemodes/space_invaders_m1.png}
        & \includegraphics[width=.31\textwidth]{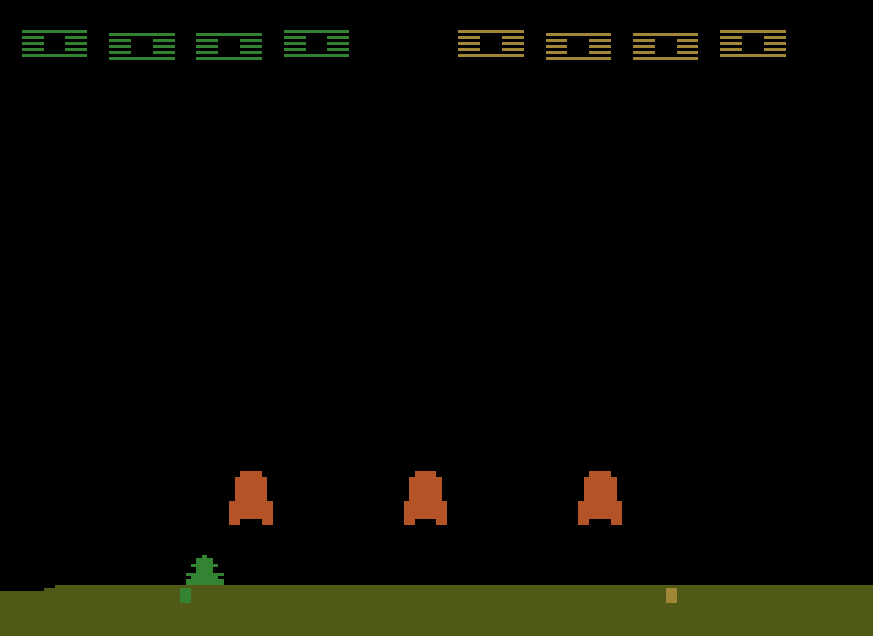}
    \end{tabular}
    \label{fig:breakout_modes}
\end{figure*}

When playing \textsc{Space Invaders} you control a spaceship which can move horizontally along the bottom of the screen. There is a grid of aliens above you and the objective of the game is to eliminate all the aliens. You are afforded some protection from the alien bullets with three barriers just above your spaceship. Difficulty 1 of \textsc{Space Invaders} widens your spaceships sprite making it harder to dodge enemy bullets. Mode 1 of \textsc{Space Invaders} causes the shields above you to oscillate horizontally. Mode 9 of \textsc{Space Invaders} is similar to Mode 12 of \textsc{Breakout} where the aliens are partially observable until struck with the player's bullet.

\subsection*{Experimental Details}

\subsubsection*{Architecture and hyperparameters}

\begin{figure}[ht!]
    \centering
    \includegraphics[scale=0.3]{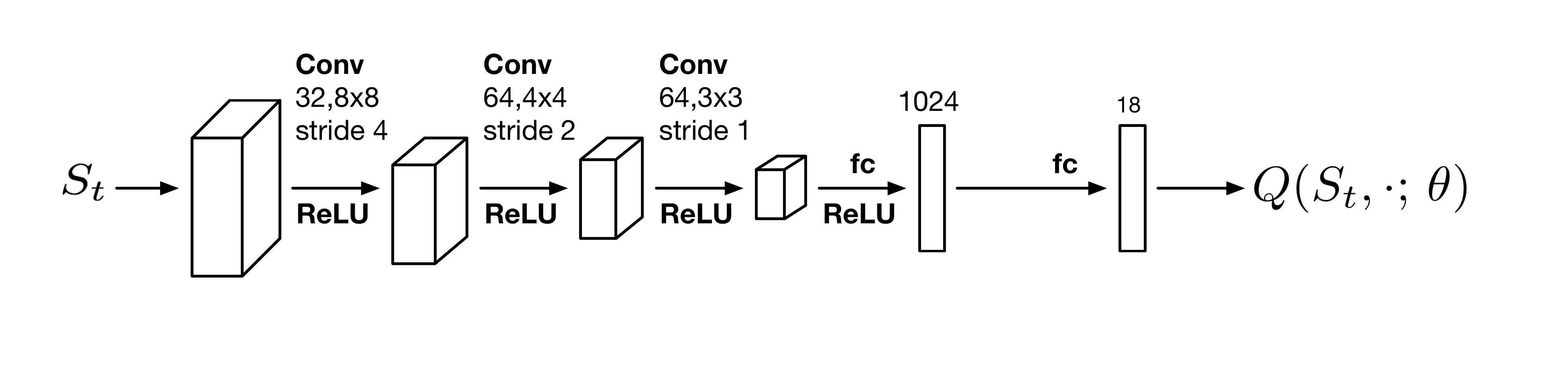}
    \vspace{-1cm}
    \caption{Network architecture used by DQN to predict state-action values.}
    \label{fig:architechture}
\end{figure}

All experiments performed in this paper utilized the neural network architecture proposed by \citet{Mnih15}. That is, a convolutional neural network with three convolutional layers and two fully connected layers. A visualization of this network can be found in Figure~\ref{fig:architechture}. Unless otherwise specified, hyperparametes are kept consistent with the ALE baselines discussed by \citet{Machado18a}. A summary of the parameters, which were consistent across all experiments, can be found in in Table~\ref{table:hyperparams}.

\vspace{-0.5cm}

\begin{figure}[ht!]
    \noindent
    \centering
    \captionof{table}{Hyperparameters for baseline results.}
    \begin{minipage}{0.49\linewidth}
        \centering
        \begin{tabular}{l l}
            Learning rate $\alpha$ & $0.00025$ \\
            Minibatch size & $32$ \\
            Learning frequency & $4$ \\
            Frame skip & $5$ \\
            Sticky action prob. & $0.25$
        \end{tabular}
    \end{minipage}
    \hfill
    \begin{minipage}{0.49\linewidth}
        \centering
        \begin{tabular}{l l}
            Replay buffer size & $1,000,000$ \\
            $\epsilon$ decay period & 1M frames \\
            $\epsilon$ initial & $1.0$ \\
            $\epsilon$ final & $0.01$ \\
            Discount factor $\gamma$ & 0.99 \\
        \end{tabular}
    \end{minipage}
    \label{table:hyperparams}
\end{figure}

\subsubsection*{Evaluation}

We adhere to the evaluation methodologies set out by \citet{Machado18a}. This includes the use of all 18 primitive actions in the ALE, not utilizing loss of life as episode termination, and the use of sticky actions to inject stochasticity. Each result outlined in this paper averages the agents performance over $100$ episodes further averaged over five runs. We do not take the maximum over runs nor the maximum over the learning curve.

When comparing results in this paper and with other evaluation methodologies it is worth noting the following terminology and time scales. We use a frame skip of $5$ frames, i.e., following every action executed by the agent the simulator advances $5$ frames into the future. The agent will take $\nicefrac{\text{\# frames}}{5}$ actions within the environment over the duration of each experiment.
One step of stochastic gradient descent to update the network parameters is performed every $4$ actions. The training routine will perform $\nicefrac{\text{\# frames}}{5 \cdot 4}$ gradient updates over the duration of each experiment. Therefore, when we discuss experiments with a duration of 50M frames this is in actuality 50M simulator frames, 10M agent steps, and 2.5M gradient updates.

\let\thefootnote\relax\footnotetext{Code available at \url{https://github.com/jessefarebro/dqn-ale}.}

\subsection*{Regularization Ablation Study}

To gain better insight into the overfitting results presented in the paper, we performed an ablation study on the two main hyperparameters used to study generalization, $\ell_2$ regularization and dropout \citep{Srivastava14}. To perform this ablation study we trained an agent in the default flavour of \textsc{Freeway} (i.e., m0d0) for 50M frames and evaluated it in two different flavours, \textsc{Freeway} m1d0, and \textsc{Freeway} m4d0. In the evaluation phase we took checkpoints every $500, 000$ frames during training and subsequently recorded the mean performance over $100$ episodes. All results presented in this section are averaged over 5 seeds.

We tested the effects of $\ell_2$ regularization, dropout, and the combination of these two methods. We varied the weighted importance $\lambda$ of our $\ell_2$ term in the DQN loss function as well as studied the dropout rate for the three convolutional layers $p_{\text{conv}}$, and the first fully connected layer $p_{\text{fc}}$. We used the loss function

\begin{equation*}
\begin{split}
    L^{\text{\tiny\textsc{DQN}}} = &\mathop{\mathbb{E}}_{\tau \, \sim \, U(\cdot)} \big[ \big( R_{t+1} + \gamma \max_{a' \in \mathcal{A}} Q(S_{t+1}, a';\, \theta^{-}) - Q(S_t, A_t;\, \theta) \big{)}^2 \big] + \lambda \,\, {\left\| \theta \right\|}^{2}_{2}\,\,\text{,}
\end{split}
\end{equation*}
where $\tau = ( S_t, A_t, R_{t+1}, S_{t+1} )$ are uniformly sampled from $U(\cdot)$, the experience replay buffer filled with experience collected by the agent. We considered the values $\lambda \in \splitatcommas{\{ 10^{-2}, 10^{-3}, 10^{-4}, 10^{-5}, 10^{-6} \}}$ for $\ell_2$ regularization, as well as the values $p_\text{conv}, p_\text{fc} \in \splitatcommas{\{(0.05,0.1),(0.1,0.2),(0.15,0.3),(0.2,0.4),(0.25,0.5)\}}$ for dropout. We conclude by analyzing the cartesian product of these two sets to study the effects of combining the two methods.

\subsubsection*{$\ell_2$ regularization}
\label{sec:ablation_l2}

\begin{figure}[t]
    \centering
    \begin{subfigure}{.31\textwidth}
        \centering
        \captionsetup{width=.8\linewidth}
        \includegraphics[width=\linewidth]{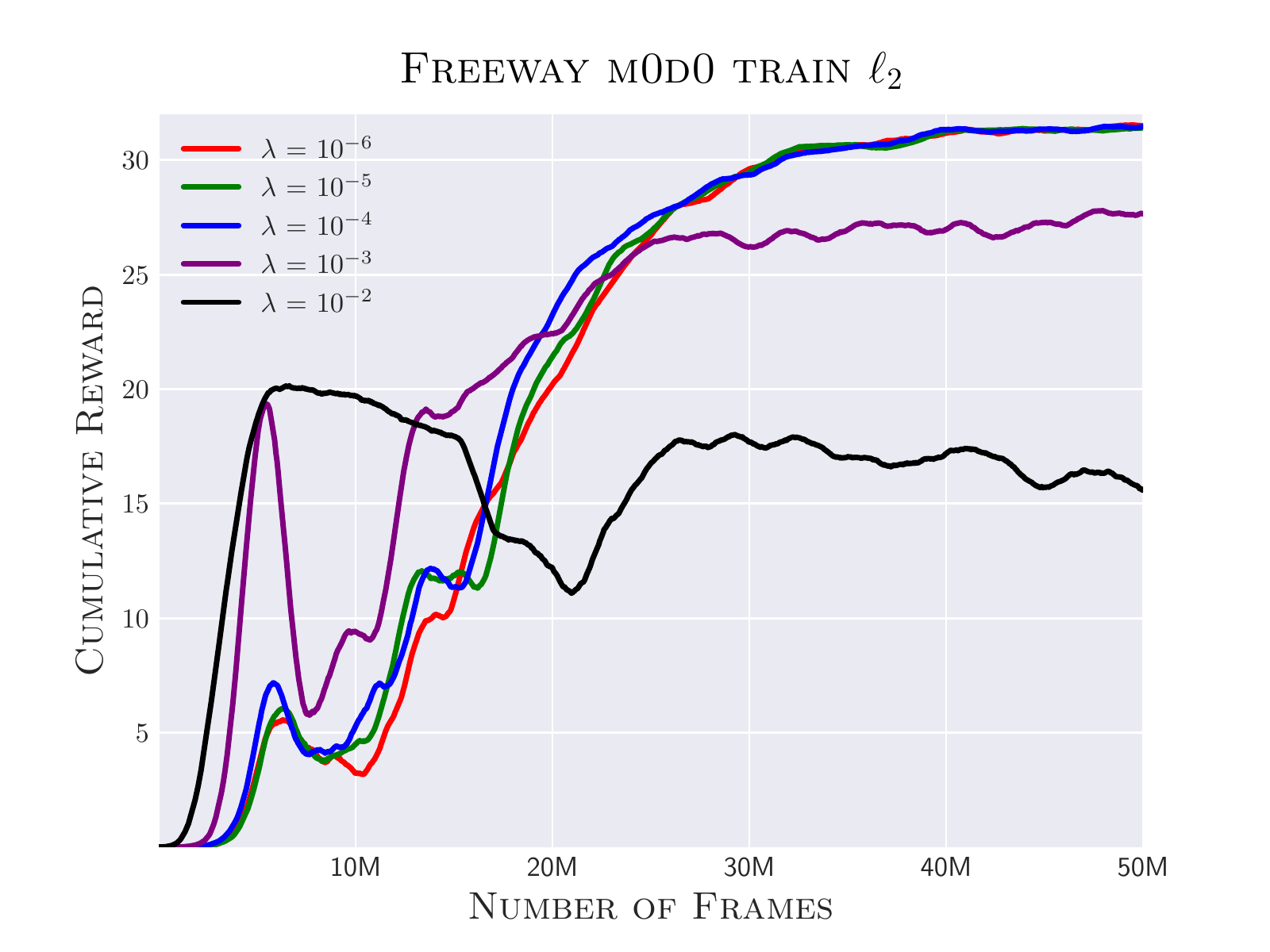}
        \caption{Performance during training in the default mode of \textsc{Freeway} with various values for $\lambda$.}
    \end{subfigure}
    \begin{subfigure}{.31\textwidth}
        \centering
        \captionsetup{width=.8\linewidth}
        \includegraphics[width=\linewidth]{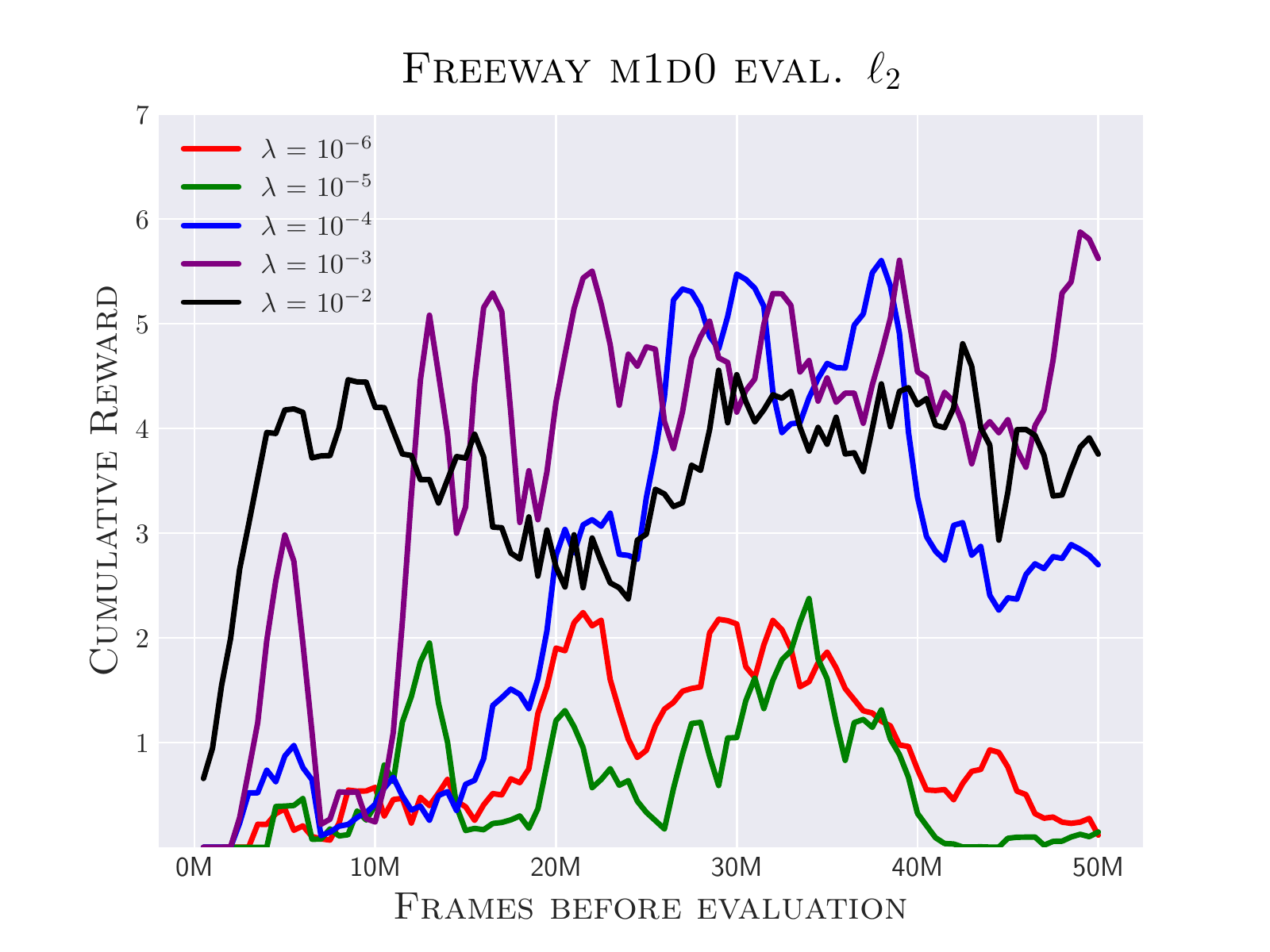}
        \caption{Performance in \textsc{Freeway} m1d0 from an agent trained with various values of $\lambda$ in \textsc{Freeway} m0d0.}
    \end{subfigure}
    \begin{subfigure}{.31\textwidth}
        \centering
        \captionsetup{width=.8\linewidth}
        \includegraphics[width=\linewidth]{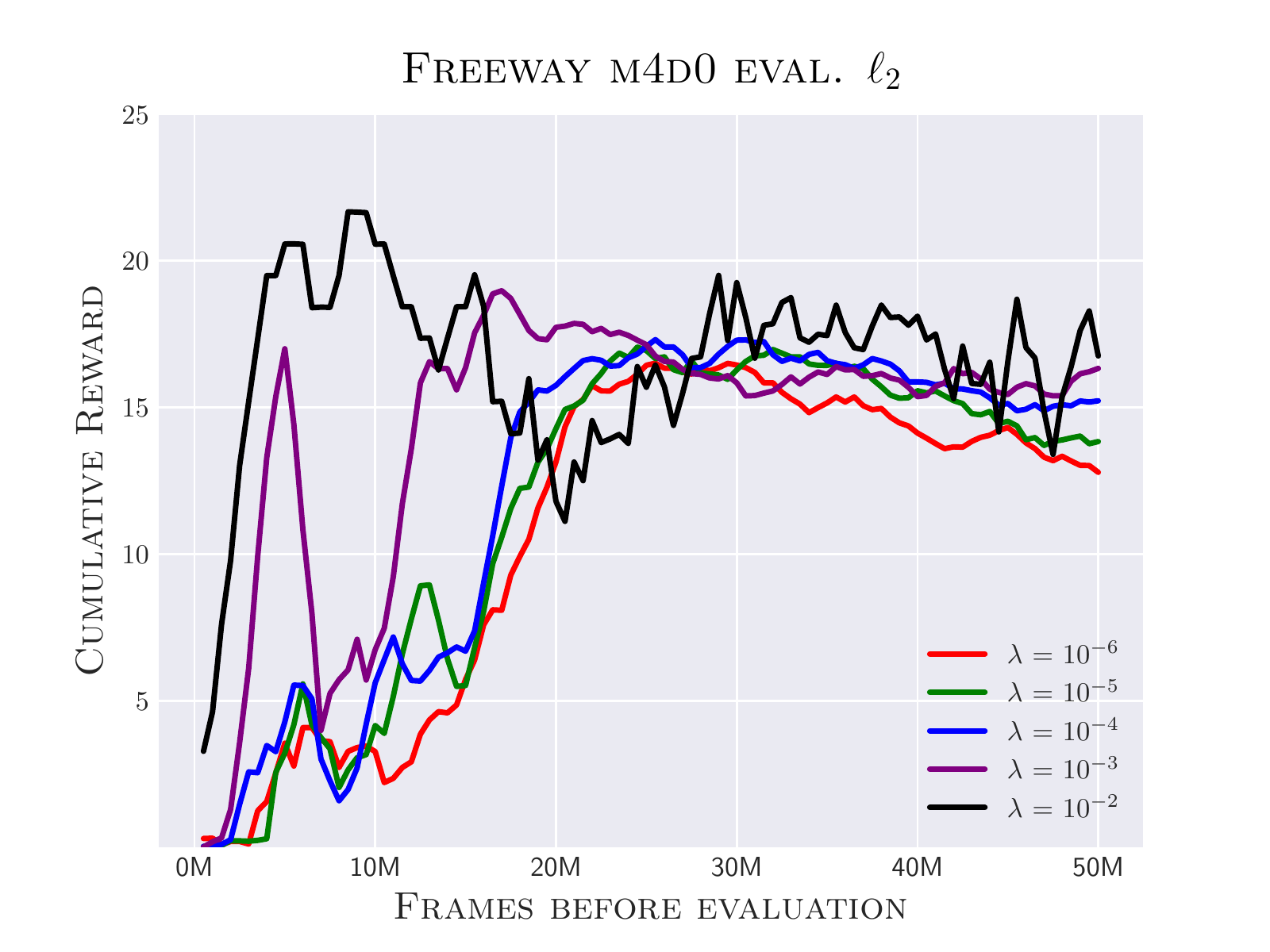}
        \caption{Performance in \textsc{Freeway} m4d0 from an agent trained with various values of $\lambda$ in \textsc{Freeway} m0d0.}
    \end{subfigure}
    \caption{Training and evaluation performance for DQN in \textsc{Freeway} using different values of $\lambda$.}
    \label{fig:ablation_l2}
\end{figure}

We begin by analyzing the training performance for DQN in \textsc{Freeway} m0d0 for different values of $\lambda$. We also provide evaluation curves for m1d0, and m4d0 of \textsc{Freeway}. Both sets of experiments are presented in Figure~\ref{fig:ablation_l2}.

Large values of $\lambda$ seem to hurt training performance and smaller values are weak enough that the agent begins to overfit to m0d0. It is worth noting the performance during evaluation in m4d0 is similar to an agent trained without $\ell_2$ regularization. The benefits of $\ell_2$ do not seem to be apparent in m4d0 but provide improvement in m1d0.

\subsubsection*{Dropout}
\label{sec:ablation_dropout}

\begin{figure}[t]
    \centering
    \begin{subfigure}{.31\textwidth}
        \centering
        \captionsetup{width=.8\linewidth}
        \includegraphics[width=\linewidth]{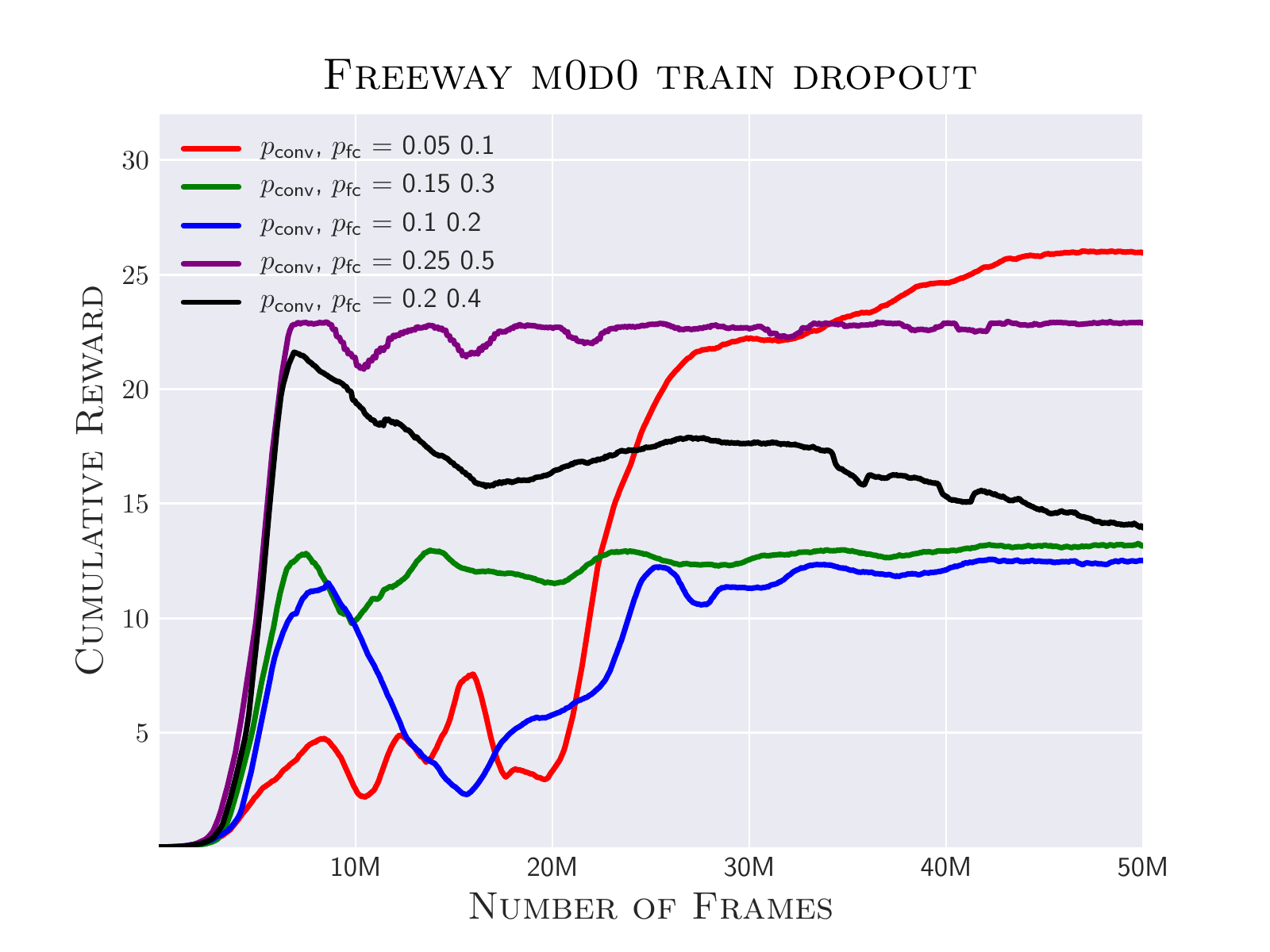}
        \caption{Performance during training in the default mode of \textsc{Freeway} with various values for $p_\text{conv}, p_\text{fc}$.}
    \end{subfigure}
    \begin{subfigure}{.31\textwidth}
        \centering
        \captionsetup{width=.8\linewidth}
        \includegraphics[width=\linewidth]{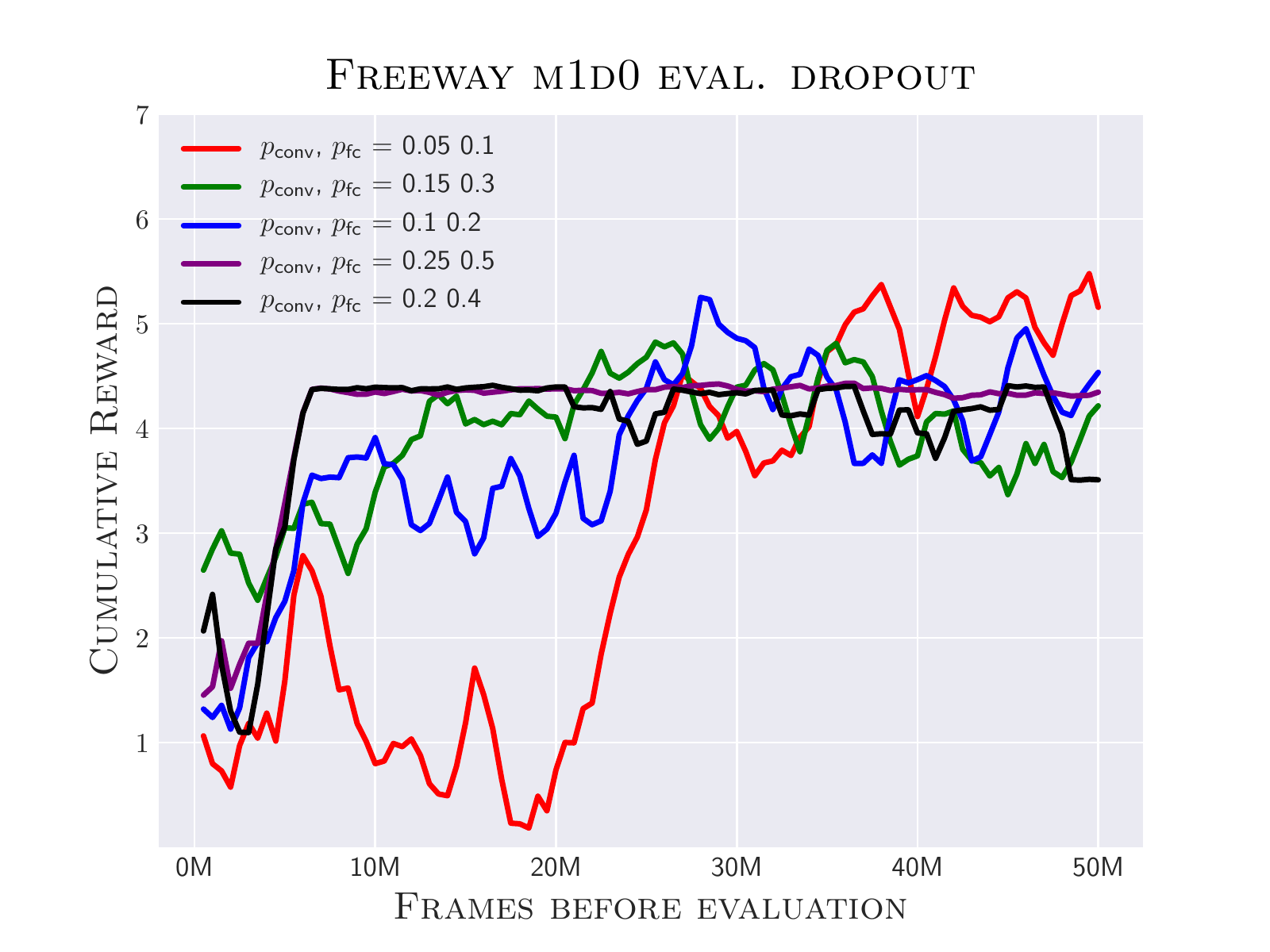}
        \caption{Performance in \textsc{Freeway} m1d0 from an agent trained with various values for $p_\text{conv}, p_\text{fc}$ in m0d0.}
    \end{subfigure}
    \begin{subfigure}{.31\textwidth}
        \centering
        \captionsetup{width=.8\linewidth}
        \includegraphics[width=\linewidth]{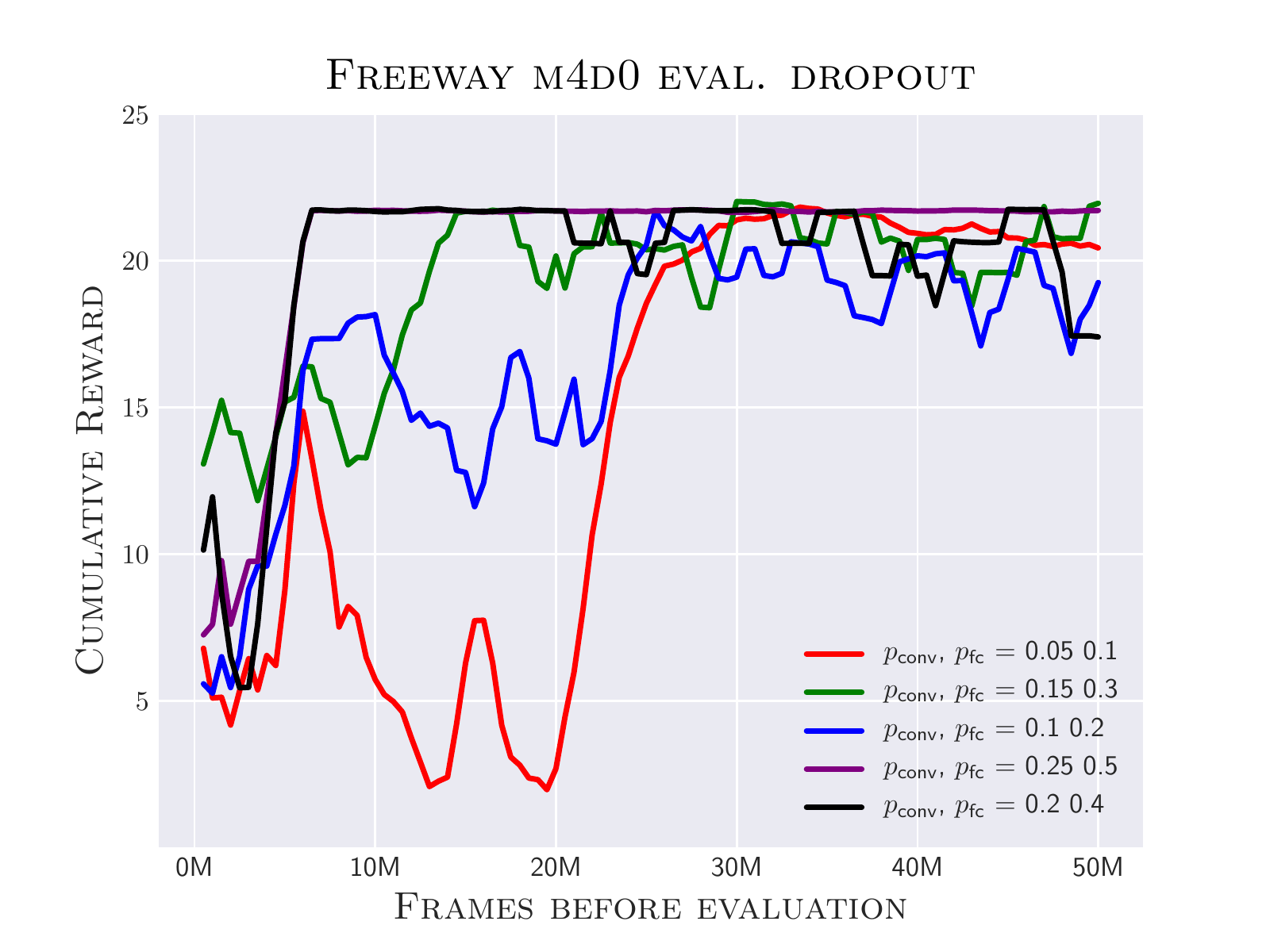}
        \caption{Performance in \textsc{Freeway} m4d0 from an agent trained with various values for $p_\text{conv}, p_\text{fc}$ in m0d0.}
    \end{subfigure}
    \caption{Training and evaluation performance for DQN in \textsc{Freeway} using different values $p_\text{conv}, p_\text{fc}$, the dropout rate for the convolutional layers and the first fully connected layer respectively.}
    \label{fig:ablation_dropout}
\end{figure}

We provide results in Figure~\ref{fig:ablation_dropout} depicting the training performance of the \textsc{Freeway} m0d0 agent with varying values of $p_\text{conv}, p_\text{fc}$. As with $\ell_2$ regularization, we further evaluate each agent checkpoint for $100$ episodes in the target flavour~during~training. 

Dropout seems to have a much bigger impact on the training performance when contrasting the results presented for $\ell_2$ regularization in Figure~\ref{fig:ablation_l2}. Curiously, larger values for the dropout rate can cause the agents' performance to flatline in both training and evaluation. The network may learn to bias a specific action, or sequence of actions independent of the state. However, reasonable dropout rates seem to improve the agents ability to generalize in both m1d0 and m4d0.

\subsubsection*{Combining $\ell_2$ regularization and dropout}

\begin{figure}[h!]
    \centering
    \subcaptionbox*{}{\includegraphics[width = .32\textwidth]{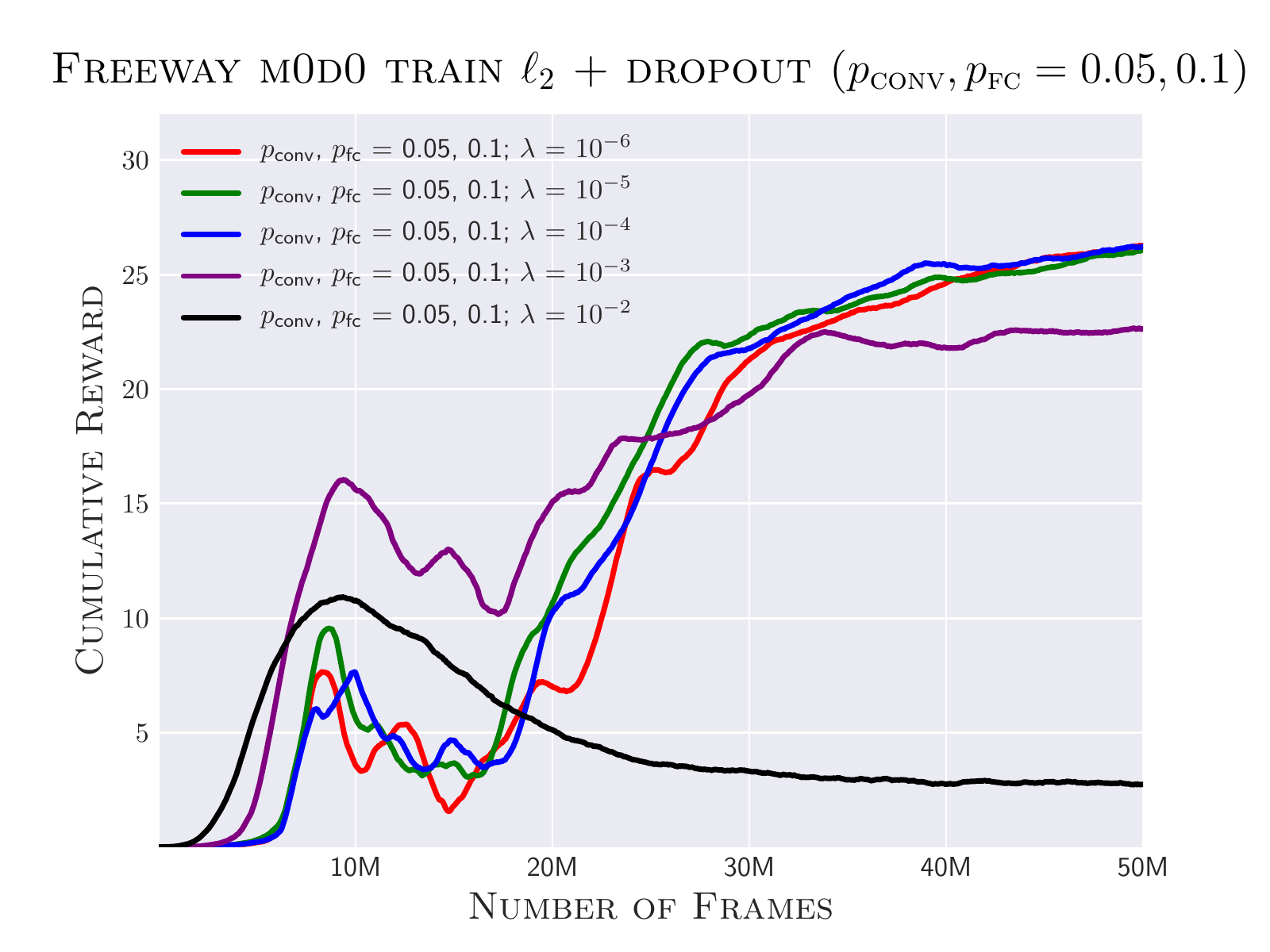}}
    \subcaptionbox*{}{\includegraphics[width = .32\textwidth]{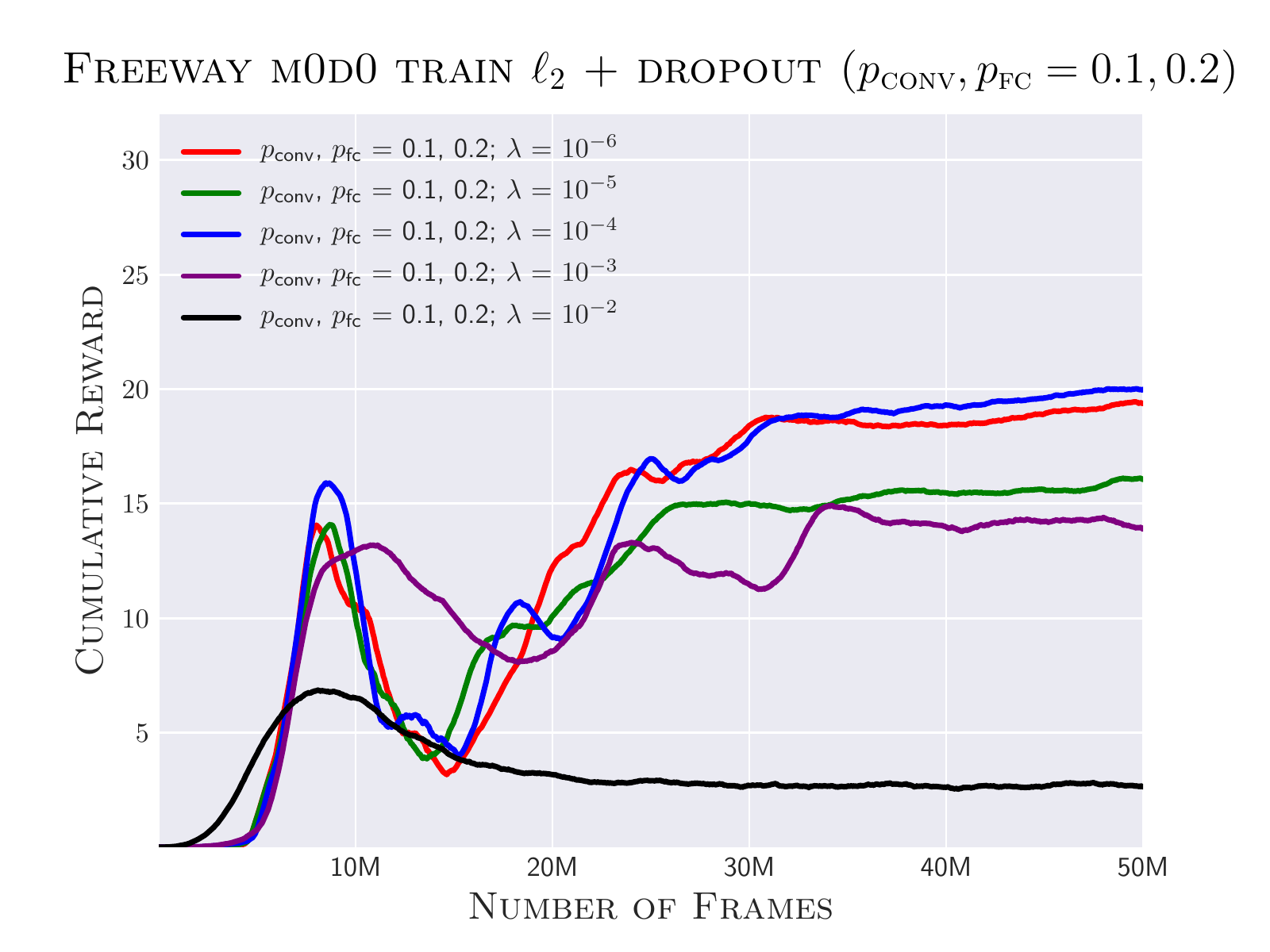}}
    \subcaptionbox*{}{\includegraphics[width = .32\textwidth]{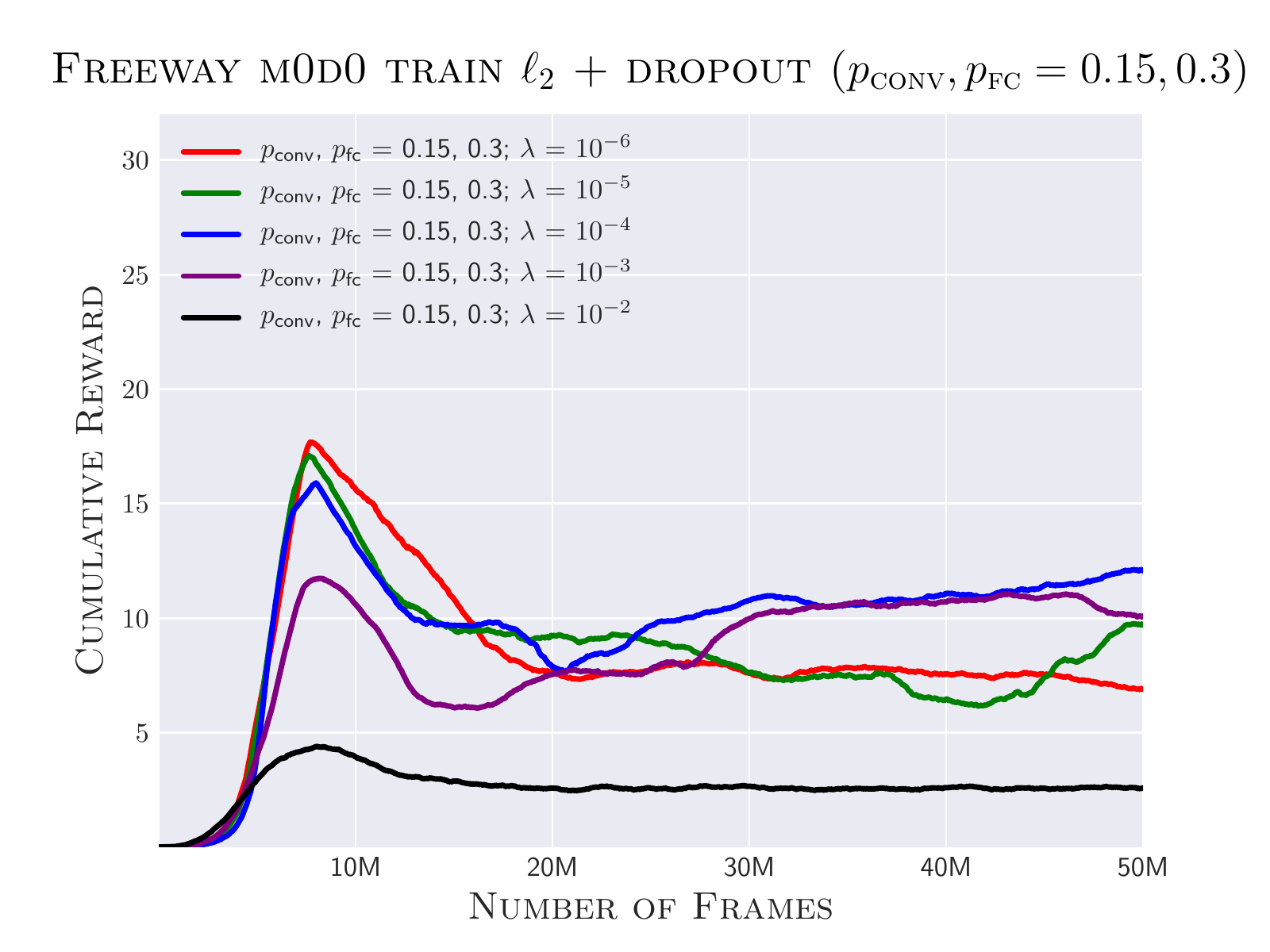}}
    \\
    \subcaptionbox*{}{\includegraphics[width = .32\textwidth]{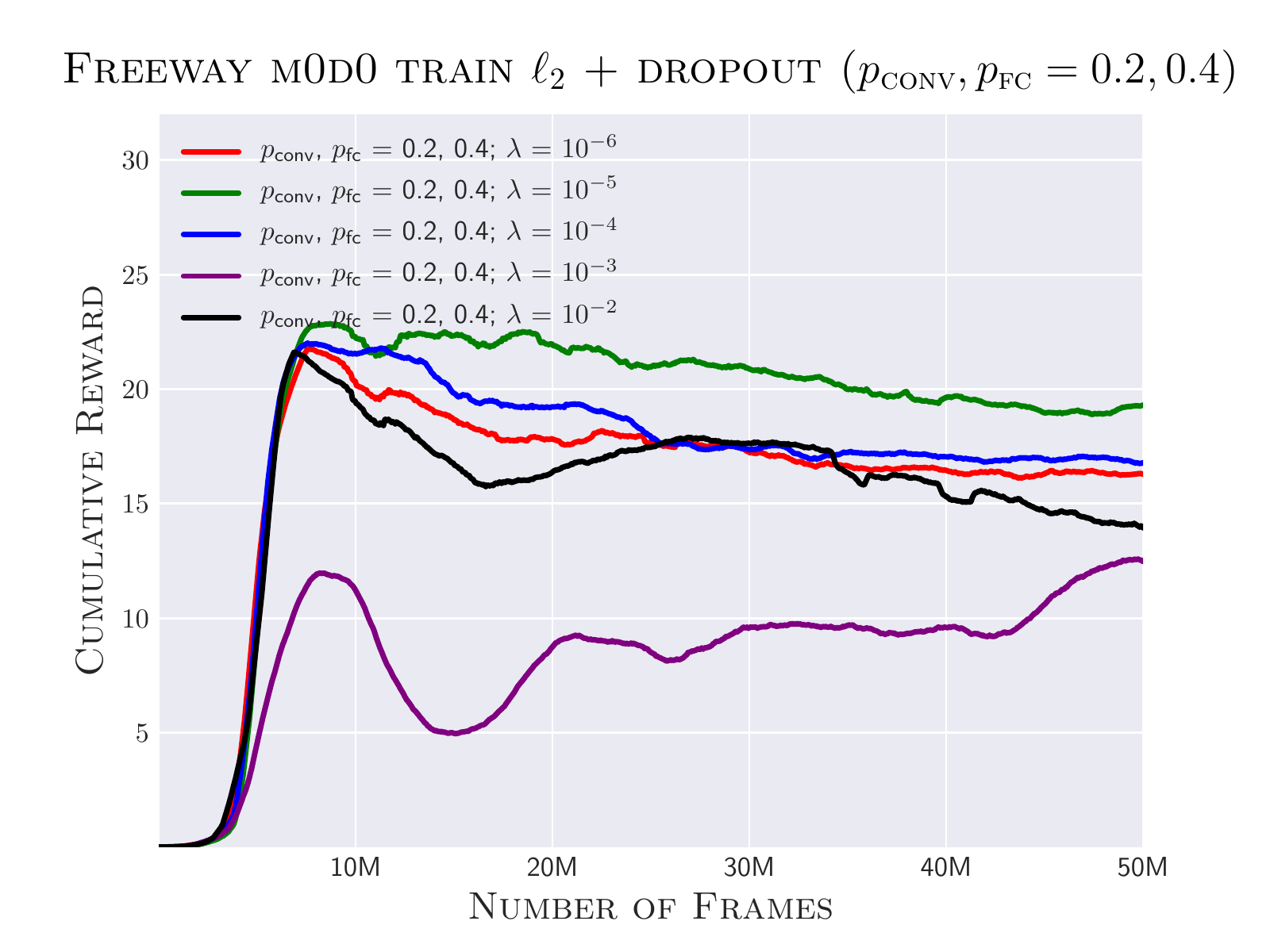}}
    \subcaptionbox*{}{\includegraphics[width = .32\textwidth]{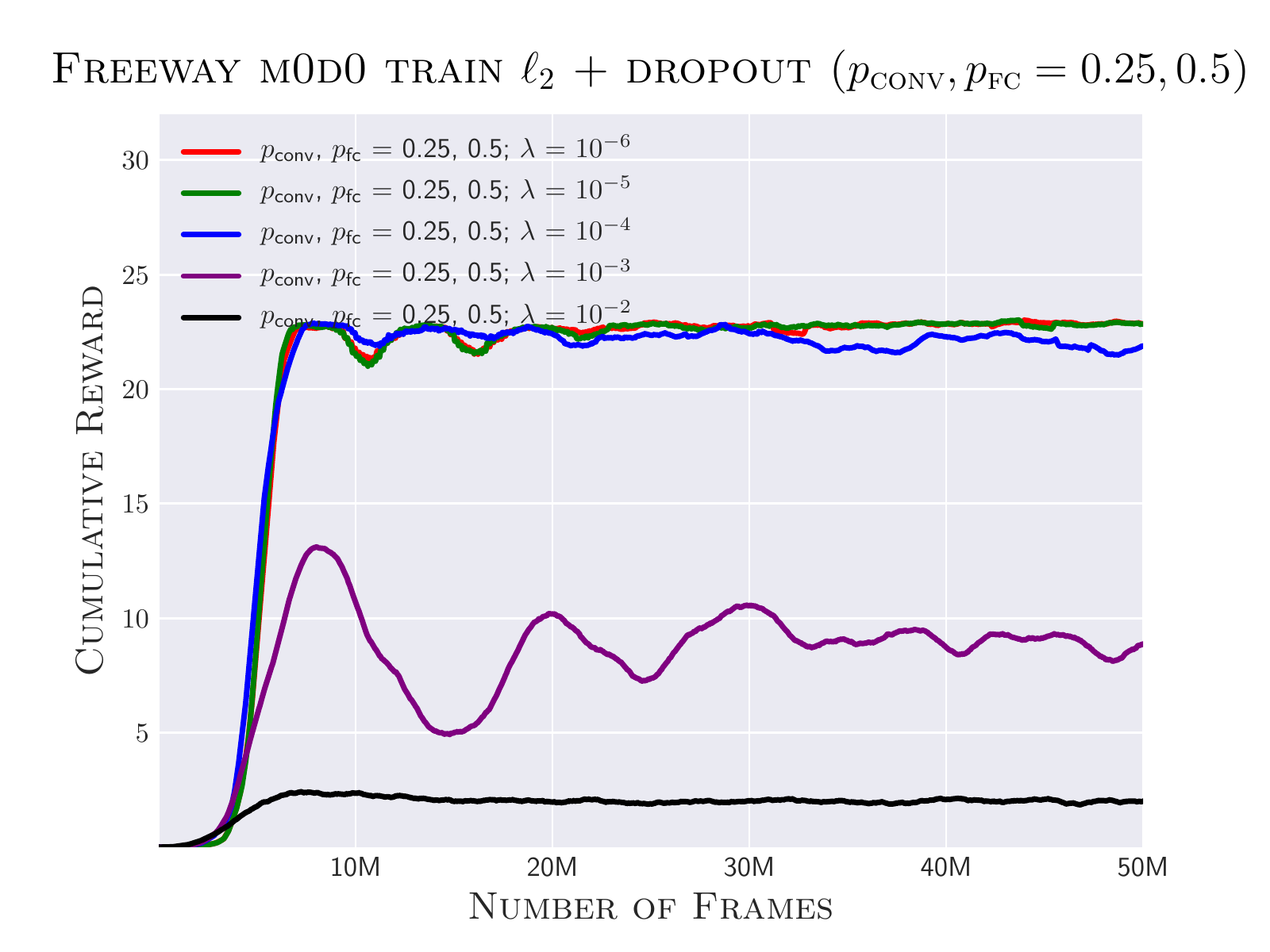}}
    \\
    \caption{Performance during training on the default flavour of \textsc{Freeway}. For each plot $p_\text{conv}, p_\text{fc}$ is held constant while varying the $\ell_2$ regularization term $\lambda$. Each parameter configuration is averaged over five seeds.}
    \label{fig:ablation_dropout_l2_train}
\end{figure}

Commonly, we see dropout and $\ell_2$ regularization combined in many supervised learning applications. We want to further explore the possibility that these two methods can provide benefits in tandem. We exhaust the cross product of the two sets of values examined above. We first analyze the impact these methods have on the training procedure in \textsc{Freeway} m0d0. Learning curves are presented in Figure~\ref{fig:ablation_dropout_l2_train}.

Interestingly, the combination of these methods can provide increased stability to the training procedure compared to the results in Figure~\ref{fig:ablation_dropout}. For example, the configuration $p_\text{conv}, p_\text{fc} = 0.1, 0.2$ scores less than $15$ when solely utilizing dropout. When applying $\ell_2$ regularization in tandem we can see the performance hover around $20$ for moderate values of $\lambda$. We continue observe the flatline behaviour for large values of $p_\text{conv}, p_\text{fc}$, regardless of $\ell_2$ regularization.

\begin{figure}[h!]
    \centering
    \subcaptionbox*{}{\includegraphics[width = .32\textwidth]{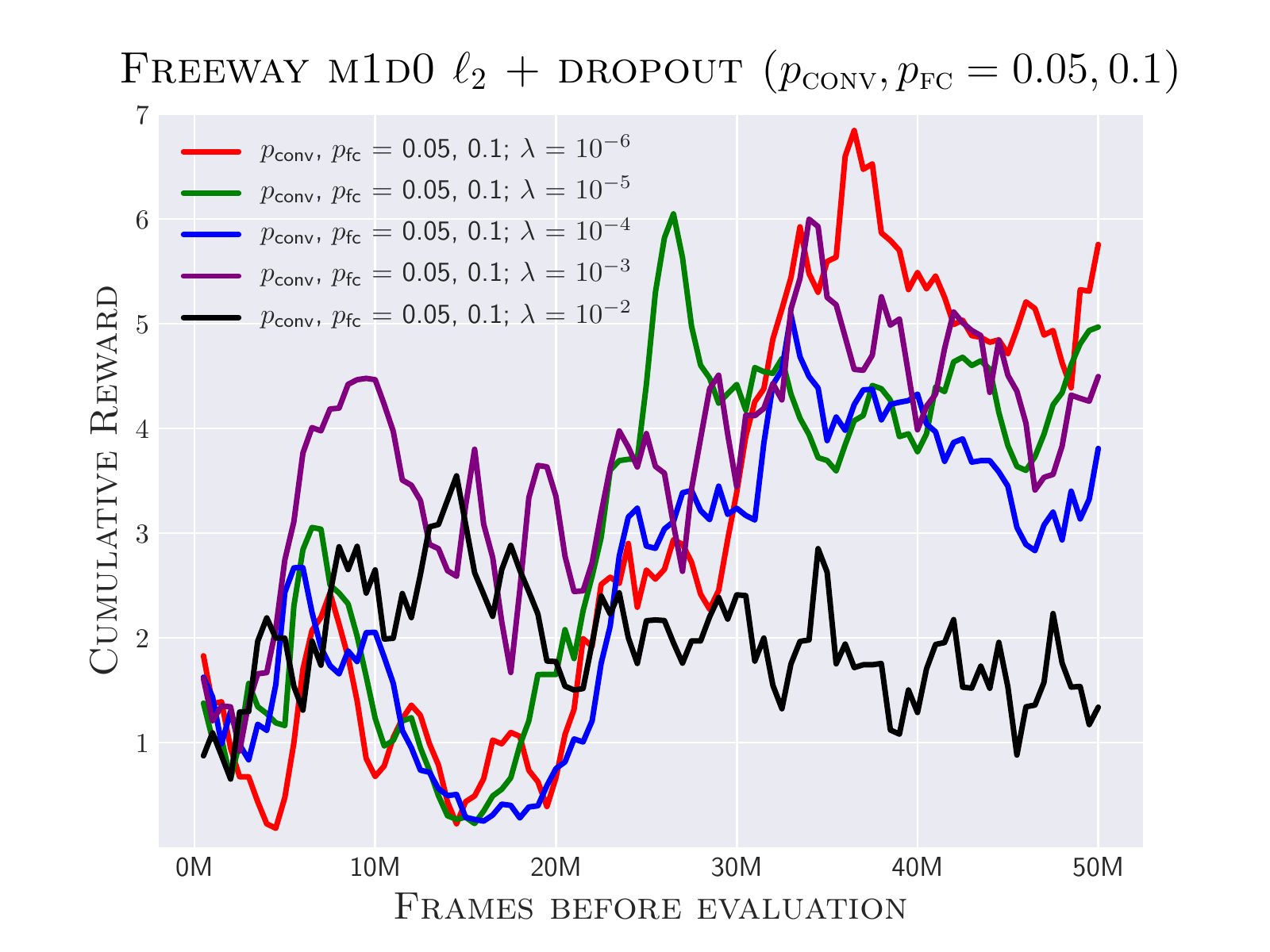}}
    \subcaptionbox*{}{\includegraphics[width = .32\textwidth]{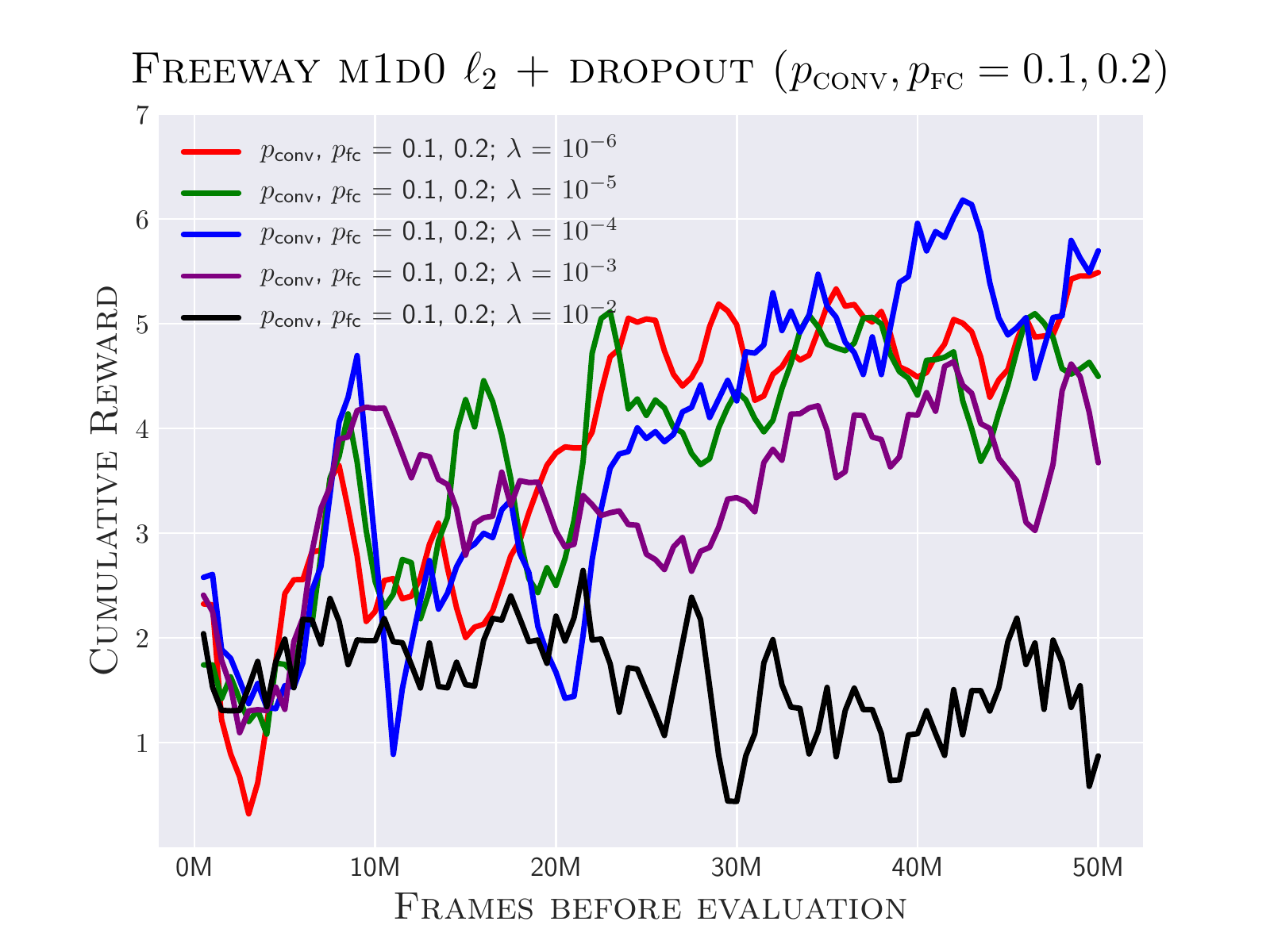}}
    \subcaptionbox*{}{\includegraphics[width = .32\textwidth]{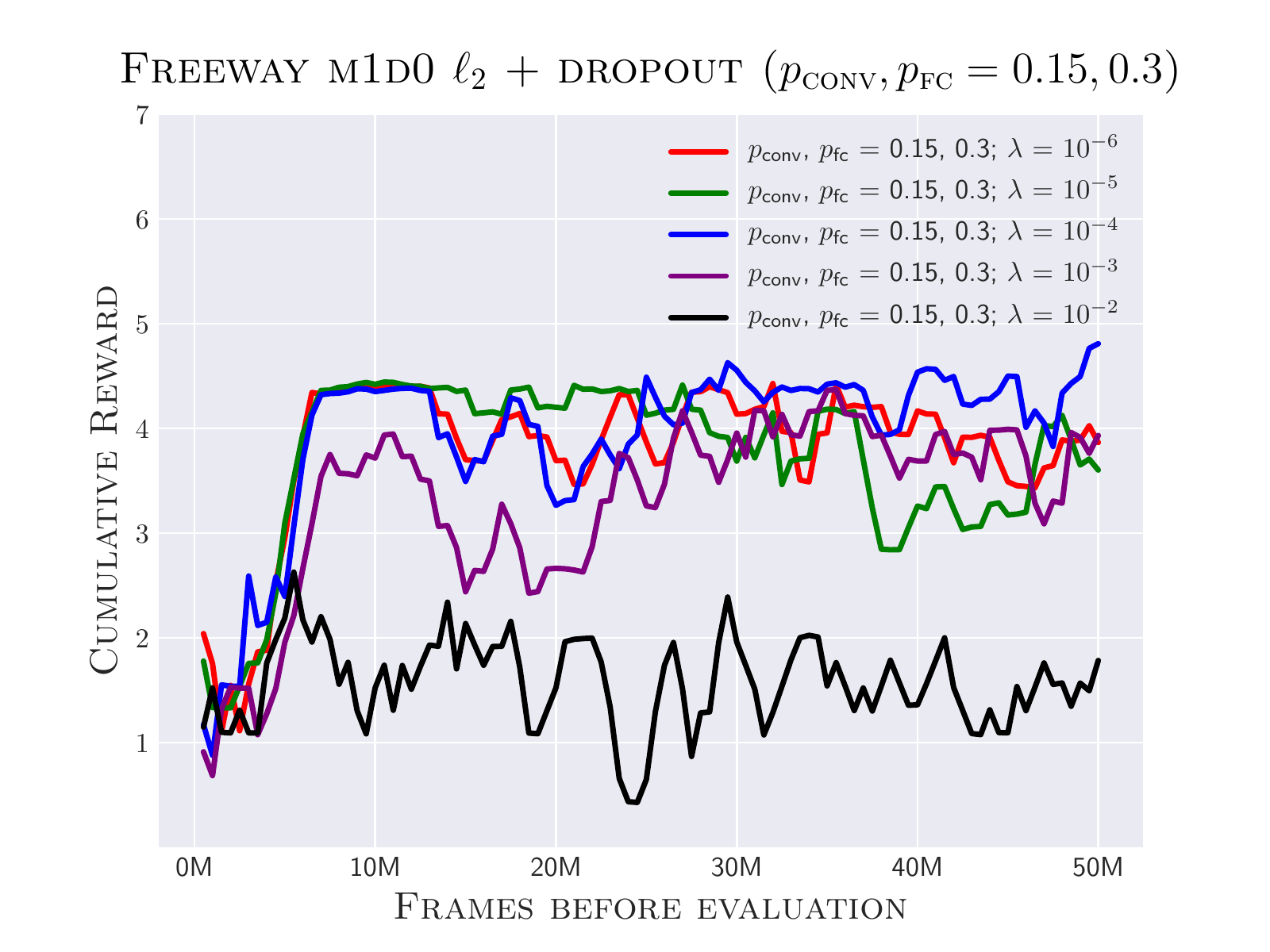}} \\
    \subcaptionbox*{}{\includegraphics[width = .32\textwidth]{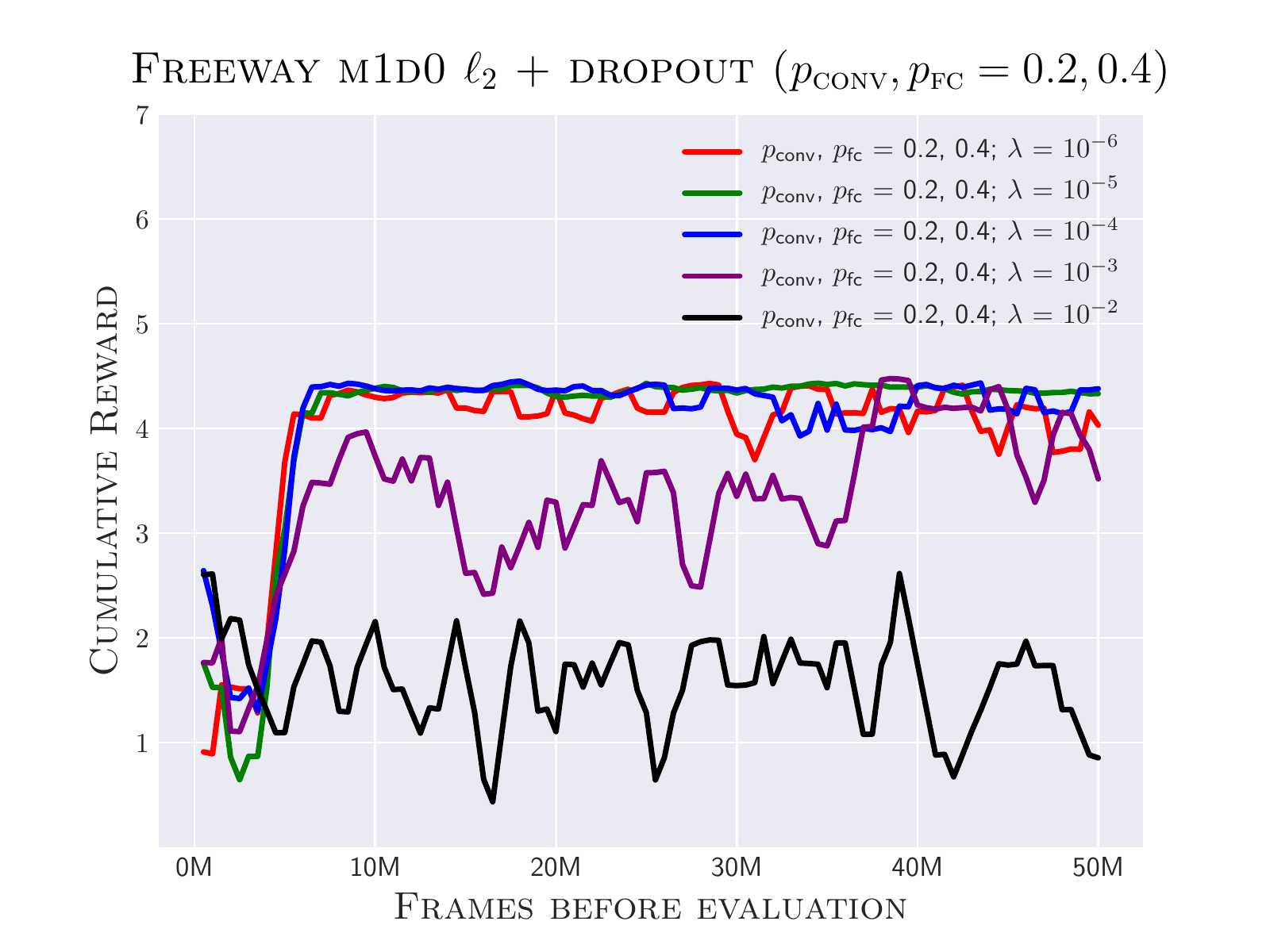}}
    \subcaptionbox*{}{\includegraphics[width = .32\textwidth]{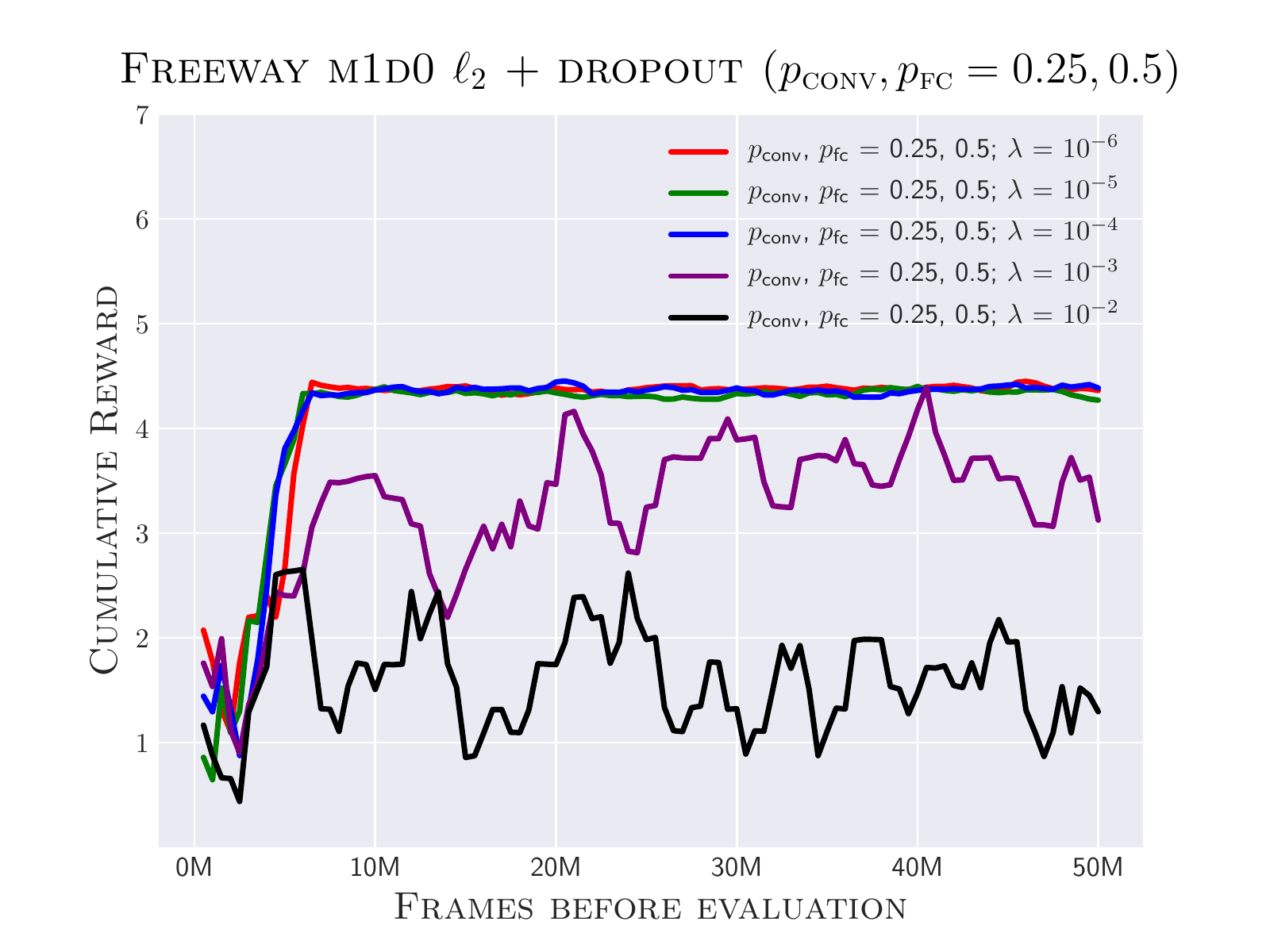}}
    \caption{Evaluation performance for \textsc{Freeway} m1d0 post-training on \textsc{Freeway} m0d0 with dropout and $\ell_2$. For each plot $p_\text{conv}, p_\text{fc}$ is held constant while varying the $\ell_2$ regularization term $\lambda$. Each configuration is averaged over five seeds.}
    \label{fig:ablation_dropout_l2_eval_m1d0}
\end{figure}

\begin{figure}[h!]
    \centering
    \subcaptionbox*{}{\includegraphics[width = .32\textwidth]{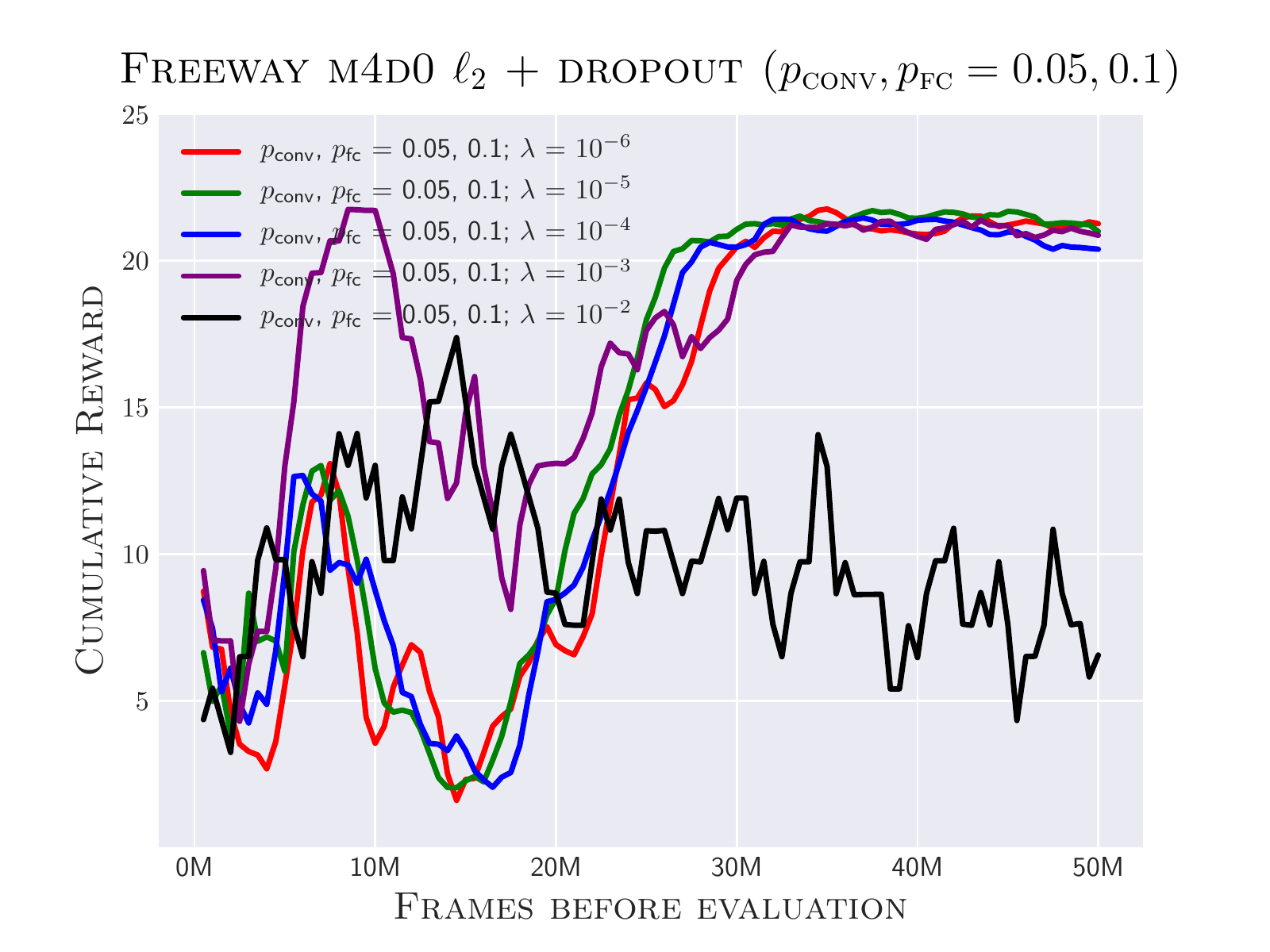}}
    \subcaptionbox*{}{\includegraphics[width = .32\textwidth]{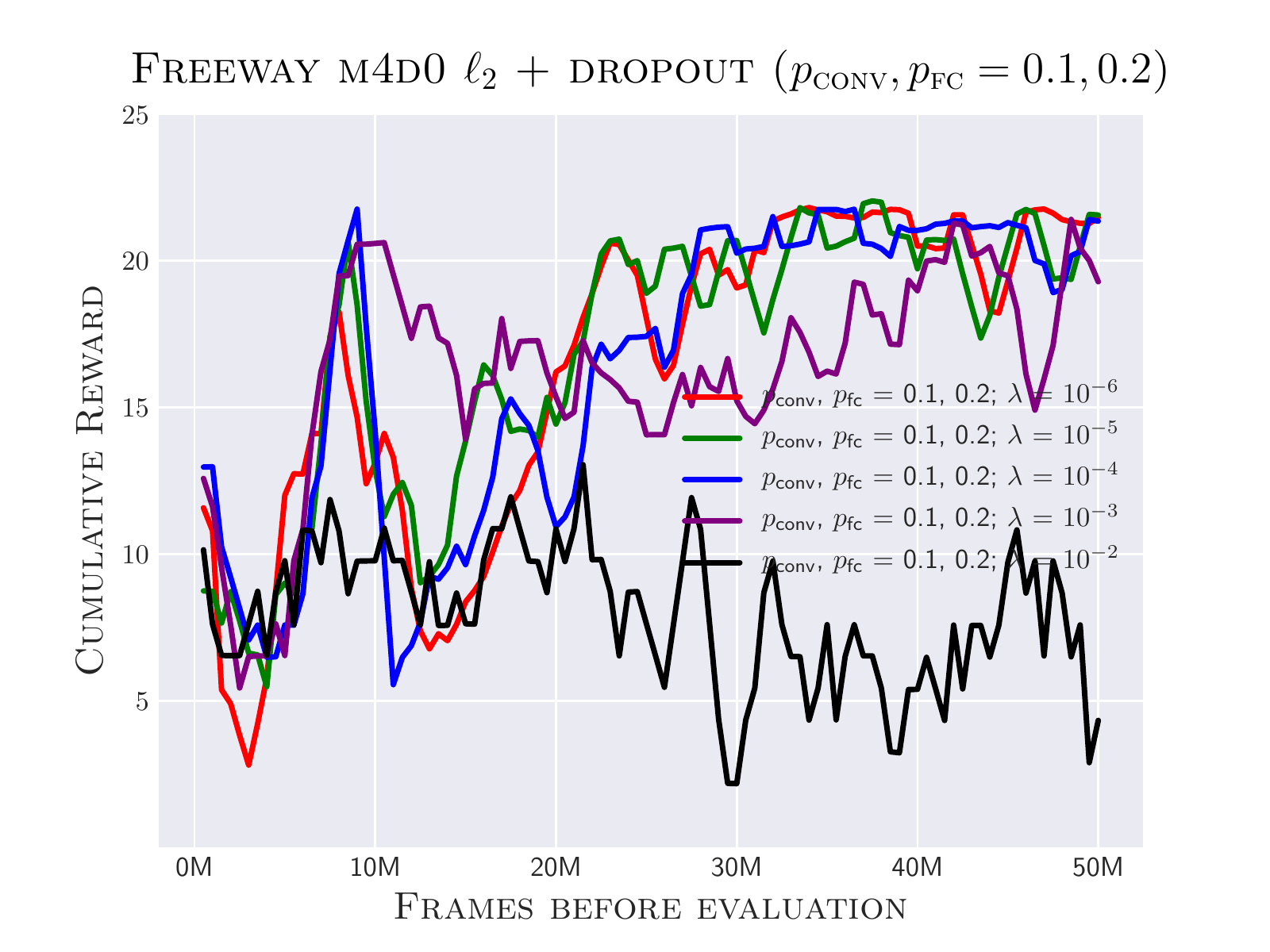}}
    \subcaptionbox*{}{\includegraphics[width = .32\textwidth]{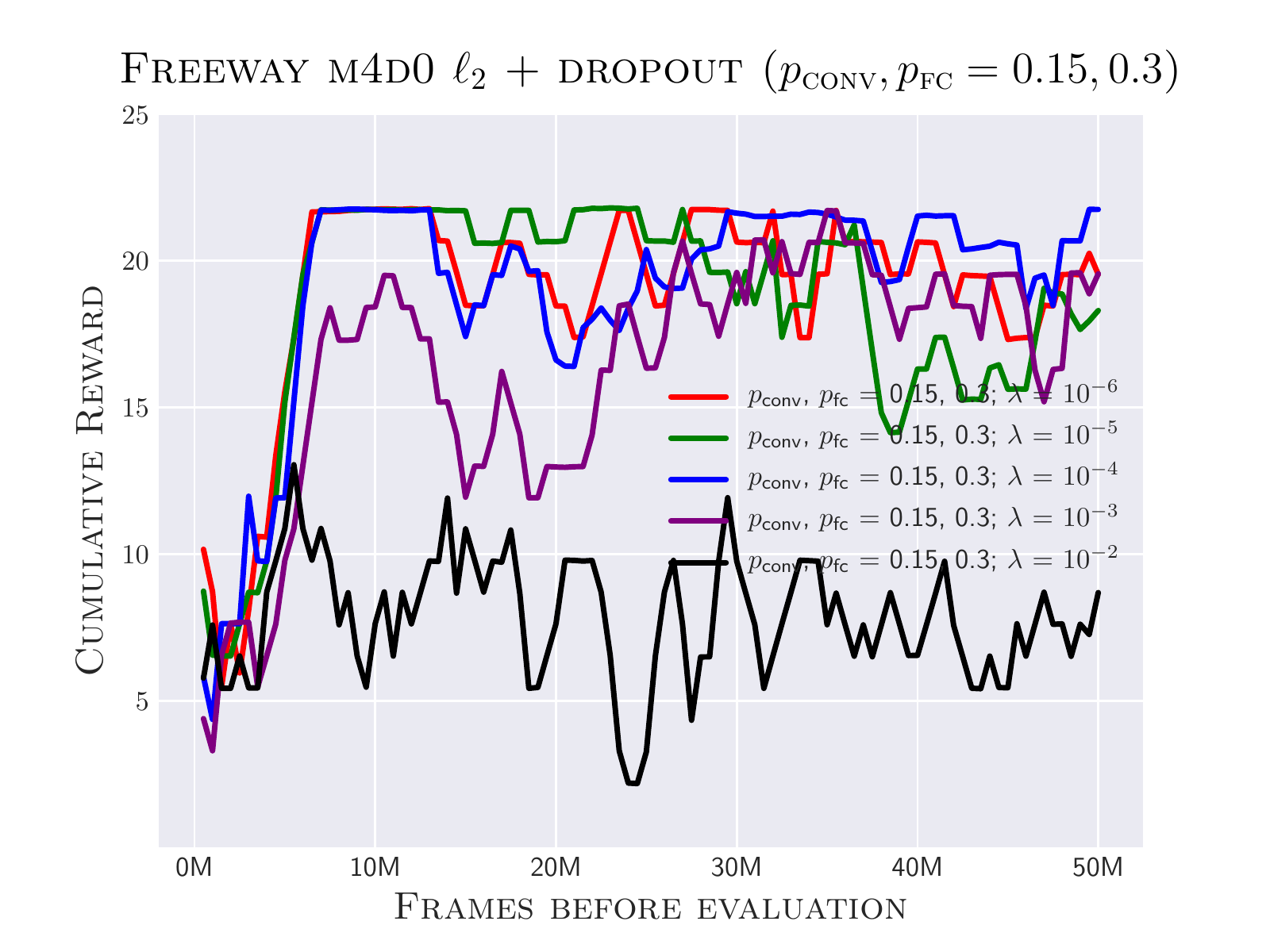}} \\
    \subcaptionbox*{}{\includegraphics[width = .32\textwidth]{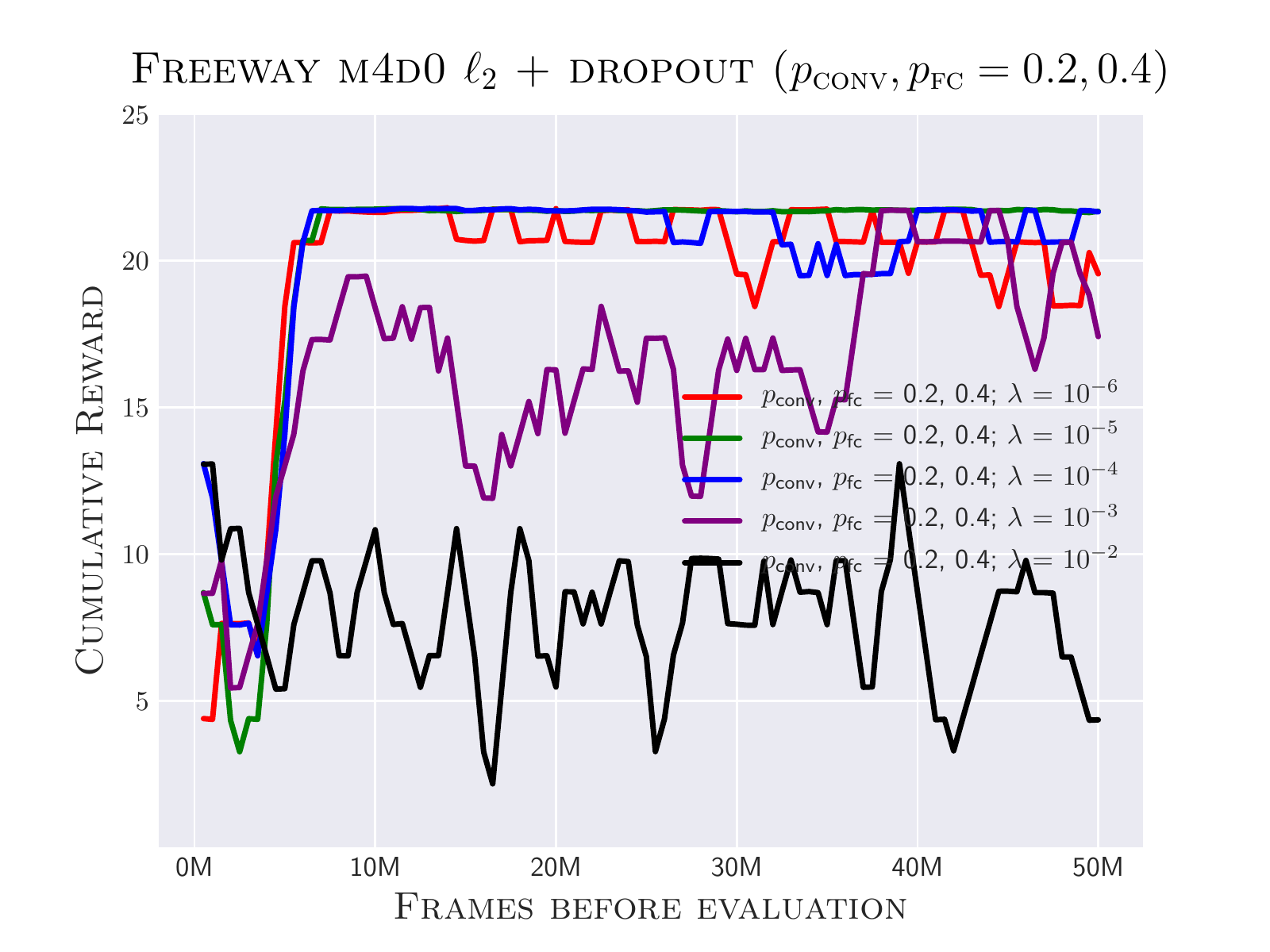}}
    \subcaptionbox*{}{\includegraphics[width = .32\textwidth]{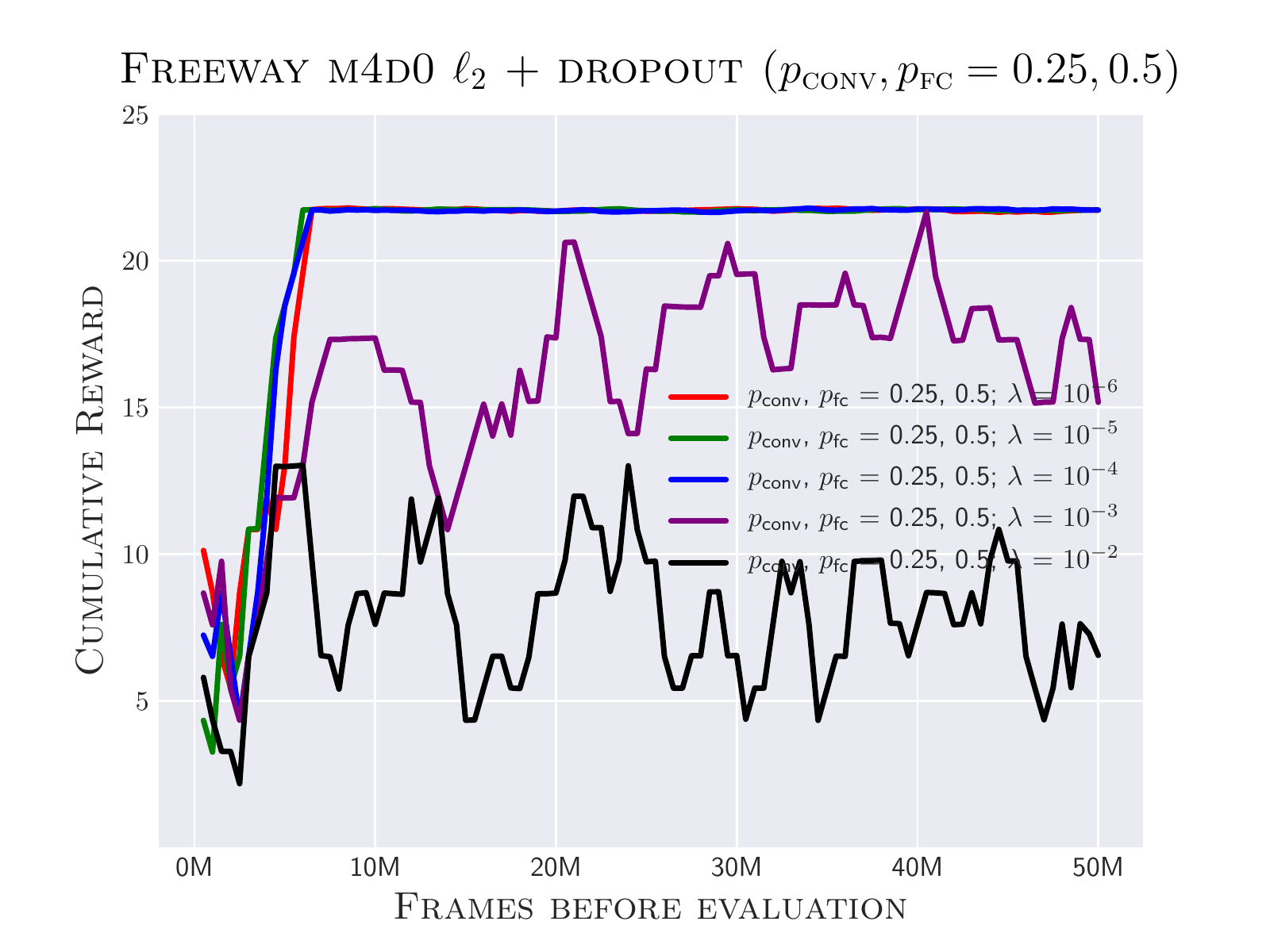}}
    \caption{Evaluation performance for \textsc{Freeway} m4d0 post-training on \textsc{Freeway} m0d0 with dropout and $\ell_2$. We used the same method described in Figure~\ref{fig:ablation_dropout_l2_eval_m1d0}.}
    \label{fig:ablation_dropout_l2_eval_m4d0}
\end{figure}

We now examine the evaluation performance for each parameter configuration in both \textsc{Freeway}~m1d0, and \textsc{Freeway}~m4d0.
These results are presented in Figure~\ref{fig:ablation_dropout_l2_eval_m1d0} for m1d0, and Figure~\ref{fig:ablation_dropout_l2_eval_m4d0} for m4d0.

We observe that $\ell_2$ regularization struggled to provide much benefit in \textsc{Freeway} m4d0. Reasonable values of dropout seem to aid generalization performance in both modes tested. It does seem that balancing the two methods of regularization can provide some benefits, such as an increased training stability and more consistent zero-shot generalization performance.

From the beginning we maintained a heuristic prescribing a balance between training performance and zero-shot generalization performance. In order to strike this balance we chose the parameters $p_\text{conv}, p_\text{fc} = 0.05, 0.1$ for the dropout rate, and $\lambda = 10^{-4}$ for the $\ell_2$ regularization parameter. These seemed to strike the best balance in early testing and the results in the ablation study seem to confirm our intuitions. 

\clearpage

\subsection*{Policy Evaluation Learning Curves}
\label{appendix:policy_transfer_cuves}
We provide learning curves for policy evaluation from a fixed representation in the default flavour of each game we analyzed. Each subplot results from evaluating a policy in the target flavour which was trained with and without regularization in the default flavour. We specifically took weight checkpoints during training every $500,000$ frames, up to 50M frames in total. Each checkpoint was then evaluated in the target flavour for $100$ episodes averaged over five runs. The regularized representation was trained using a dropout rate of $p_\text{conv}, p_\text{fc} = 0.05, 0.1$, and $\lambda = 10^{-4}$ for $\ell_2$ regularization.

\begin{figure}[h!]
\centering
\subcaptionbox*{}{\includegraphics[width = .28\textwidth]{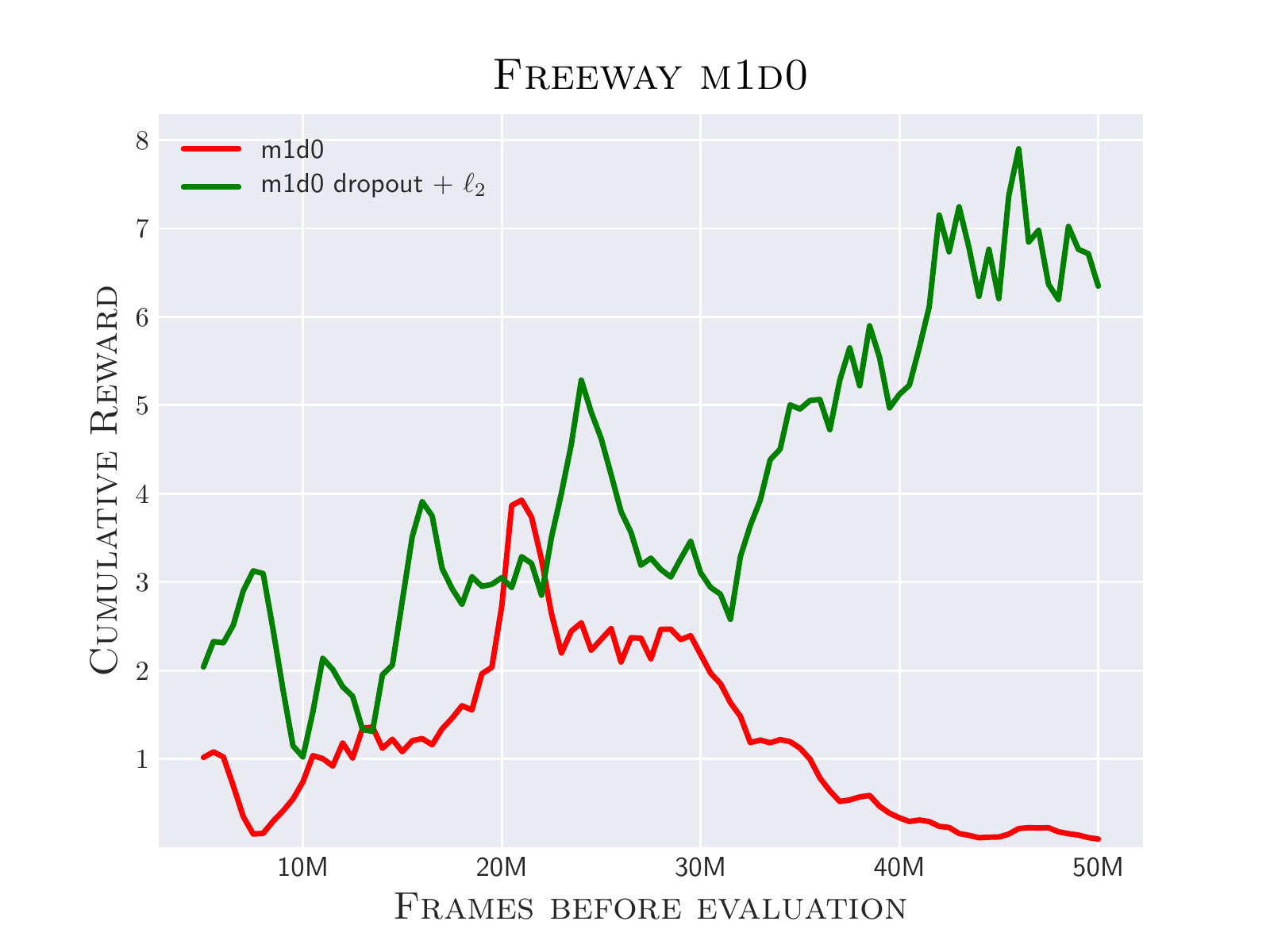}}
\subcaptionbox*{}{\includegraphics[width = .28\textwidth]{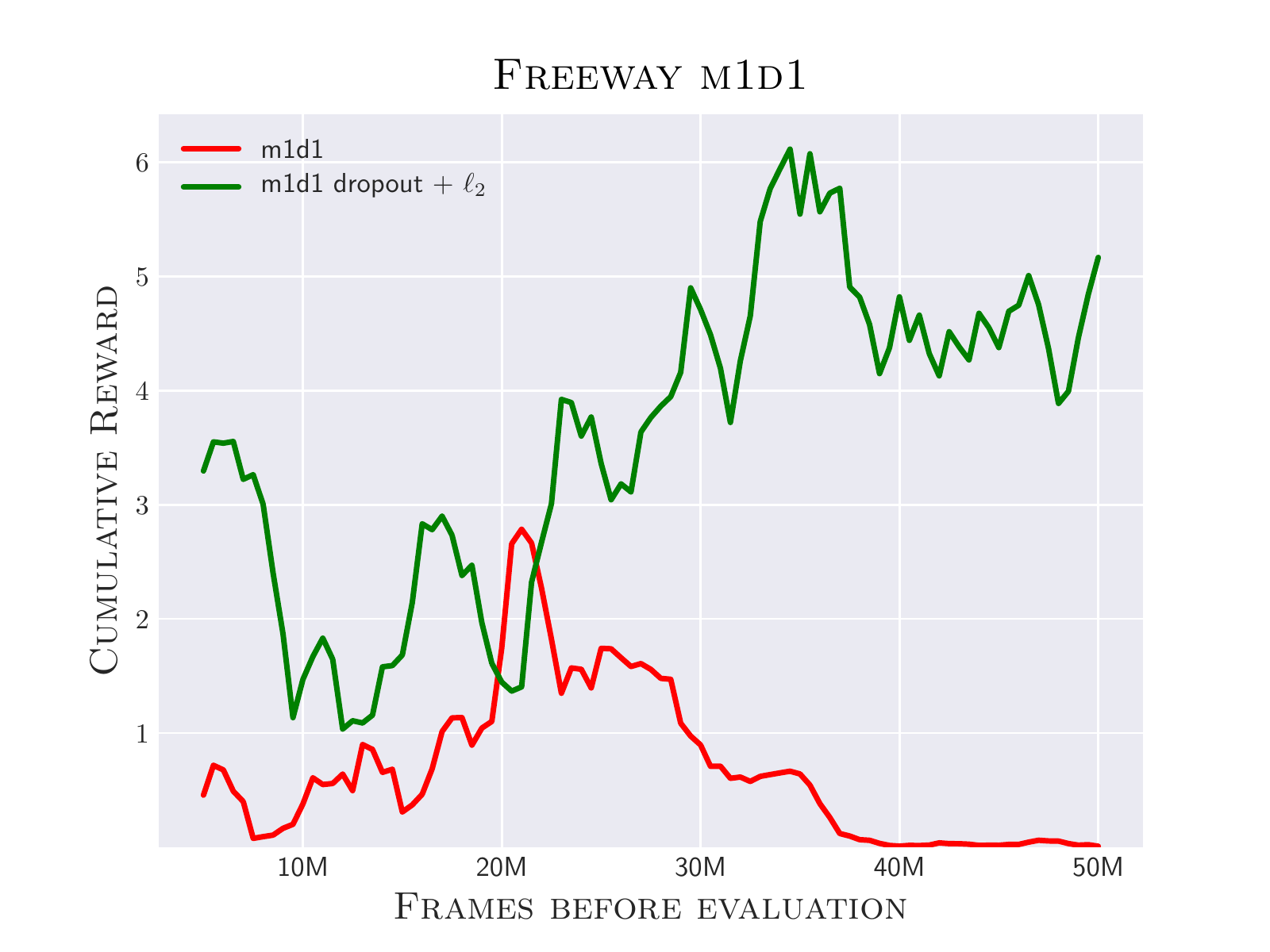}}
\subcaptionbox*{}{\includegraphics[width = .28\textwidth]{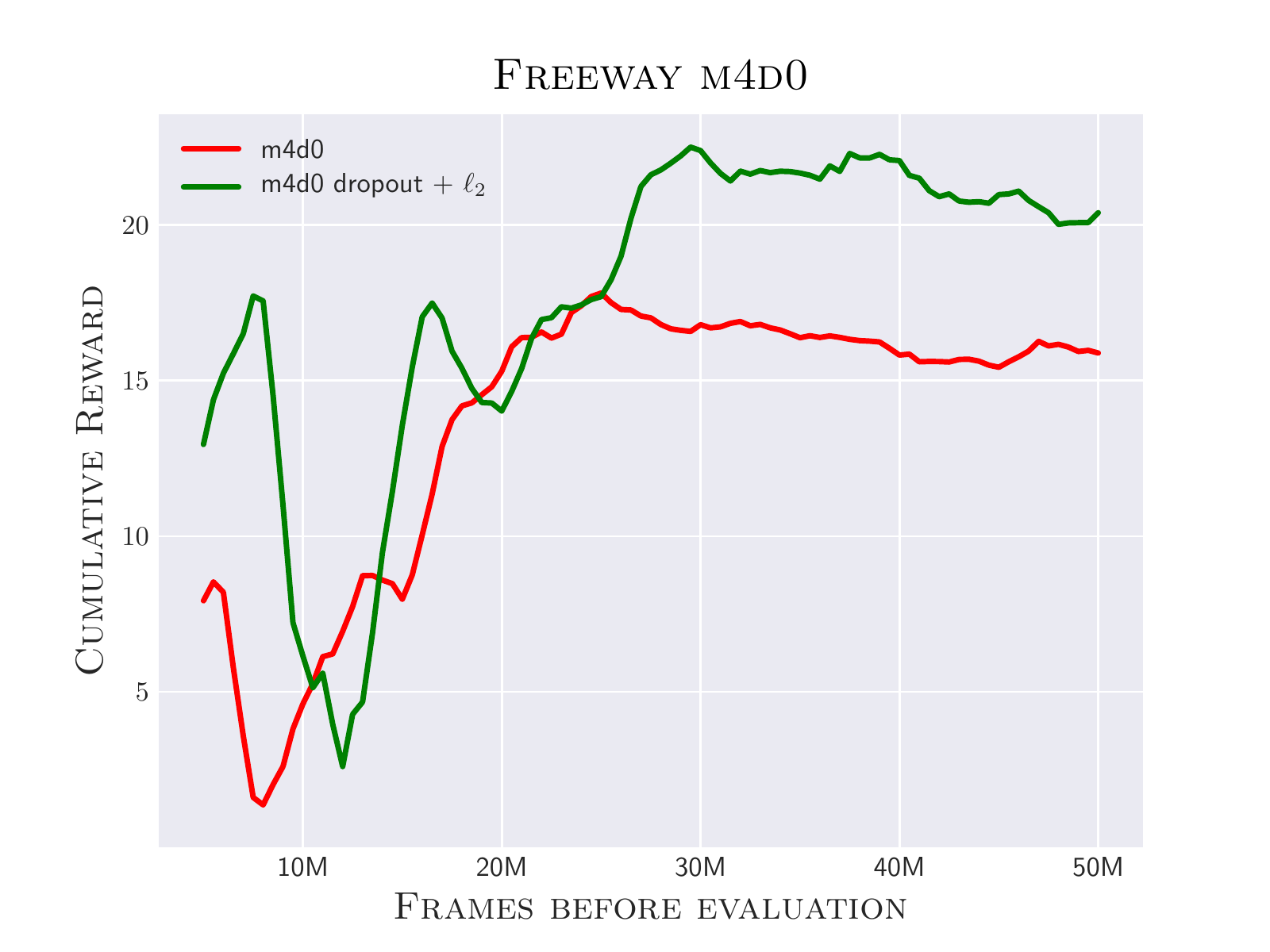}}
\\
\subcaptionbox*{}{\includegraphics[width = .28\textwidth]{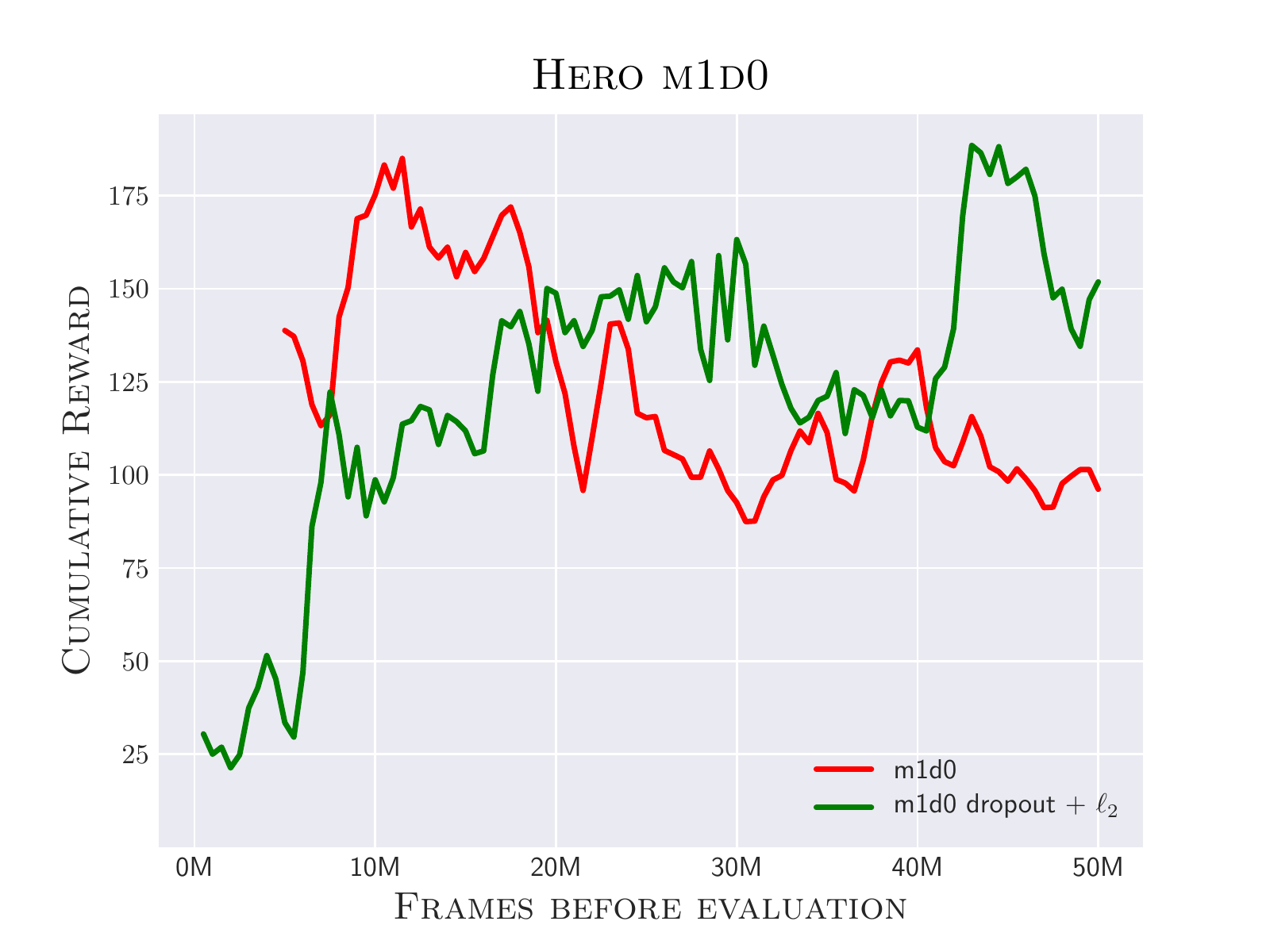}}
\subcaptionbox*{}{\includegraphics[width = .28\textwidth]{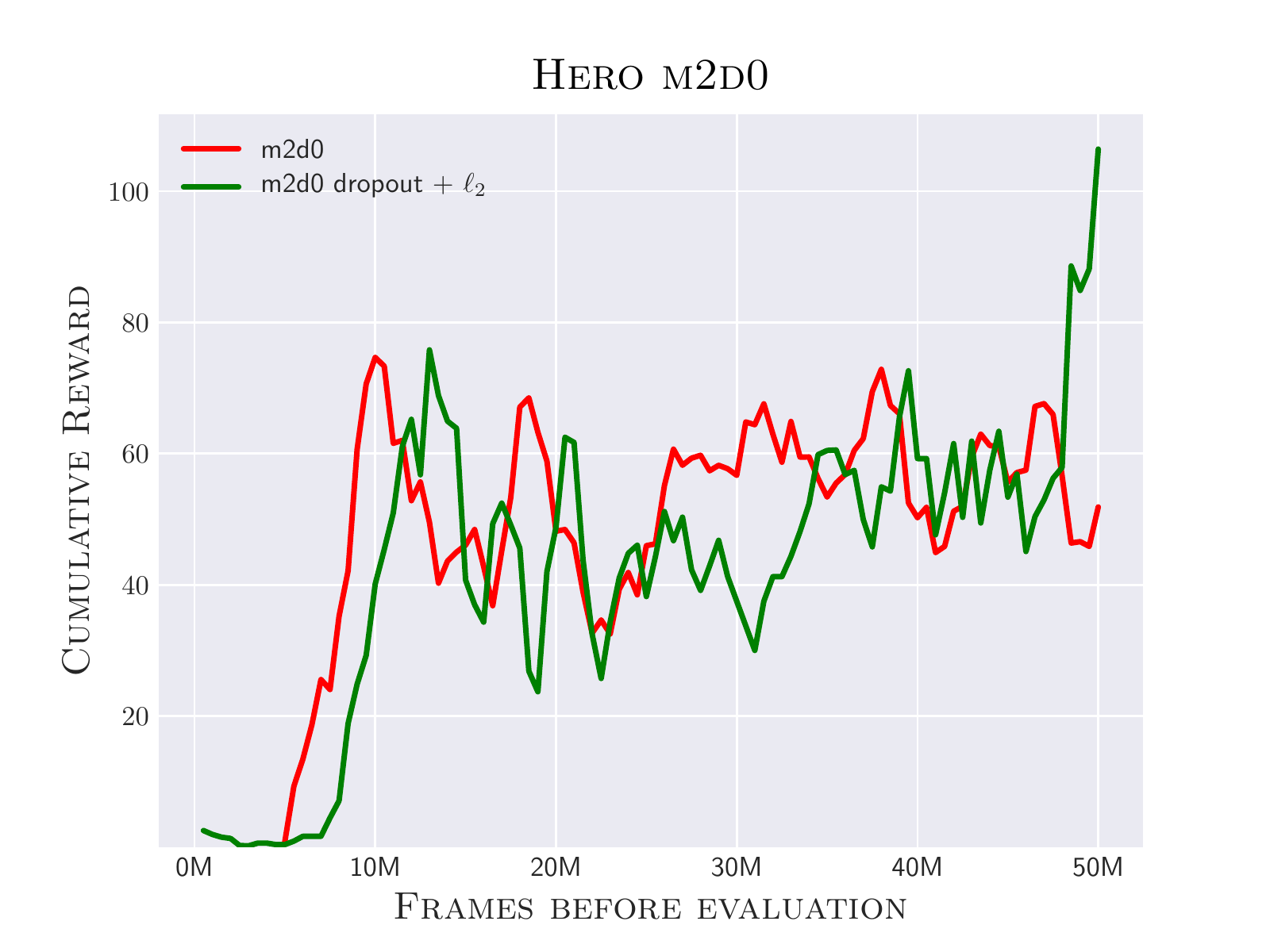}}
\\
\subcaptionbox*{}{\includegraphics[width = .28\textwidth]{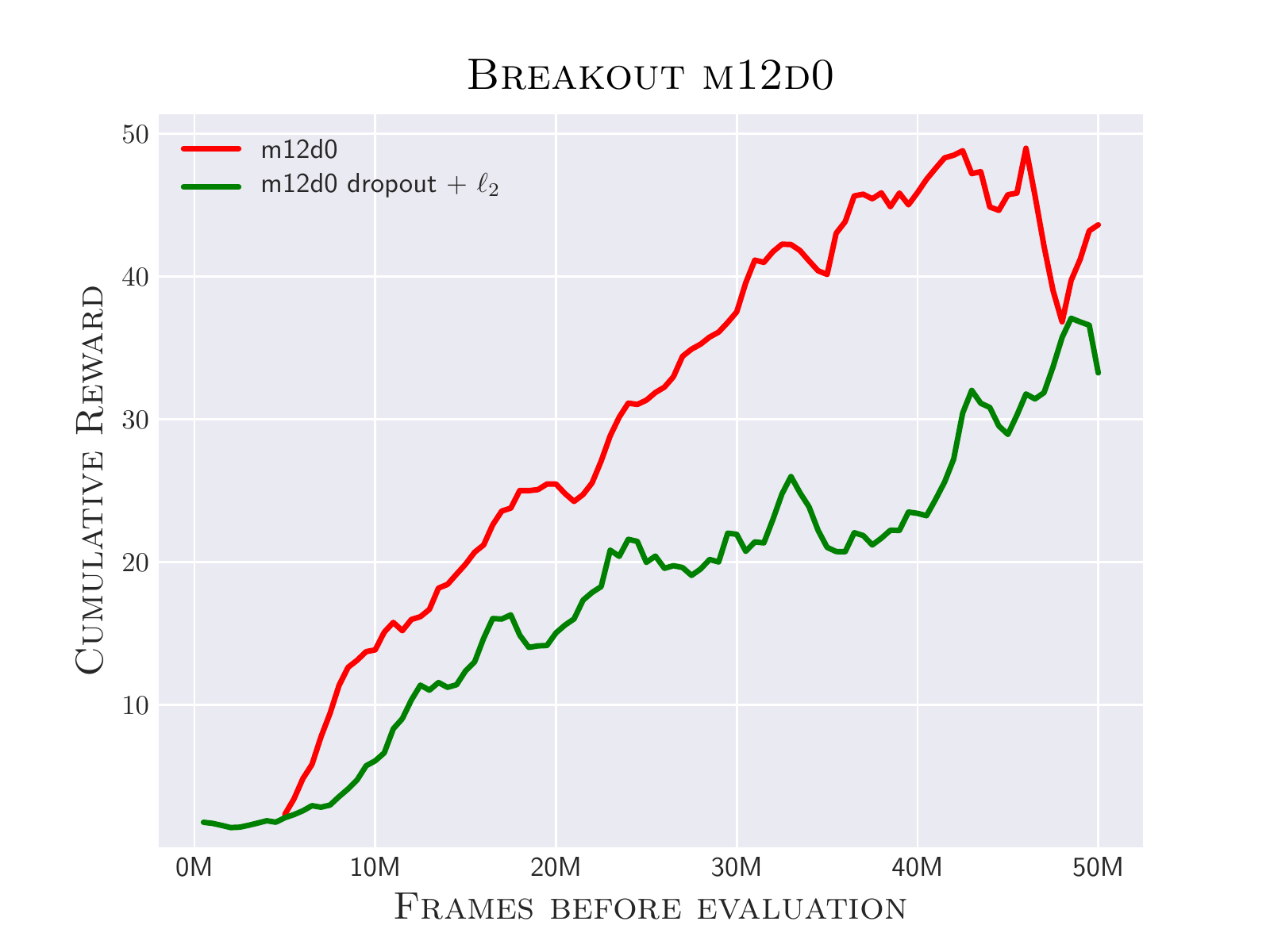}}
\\
\subcaptionbox*{}{\includegraphics[width = .28\textwidth]{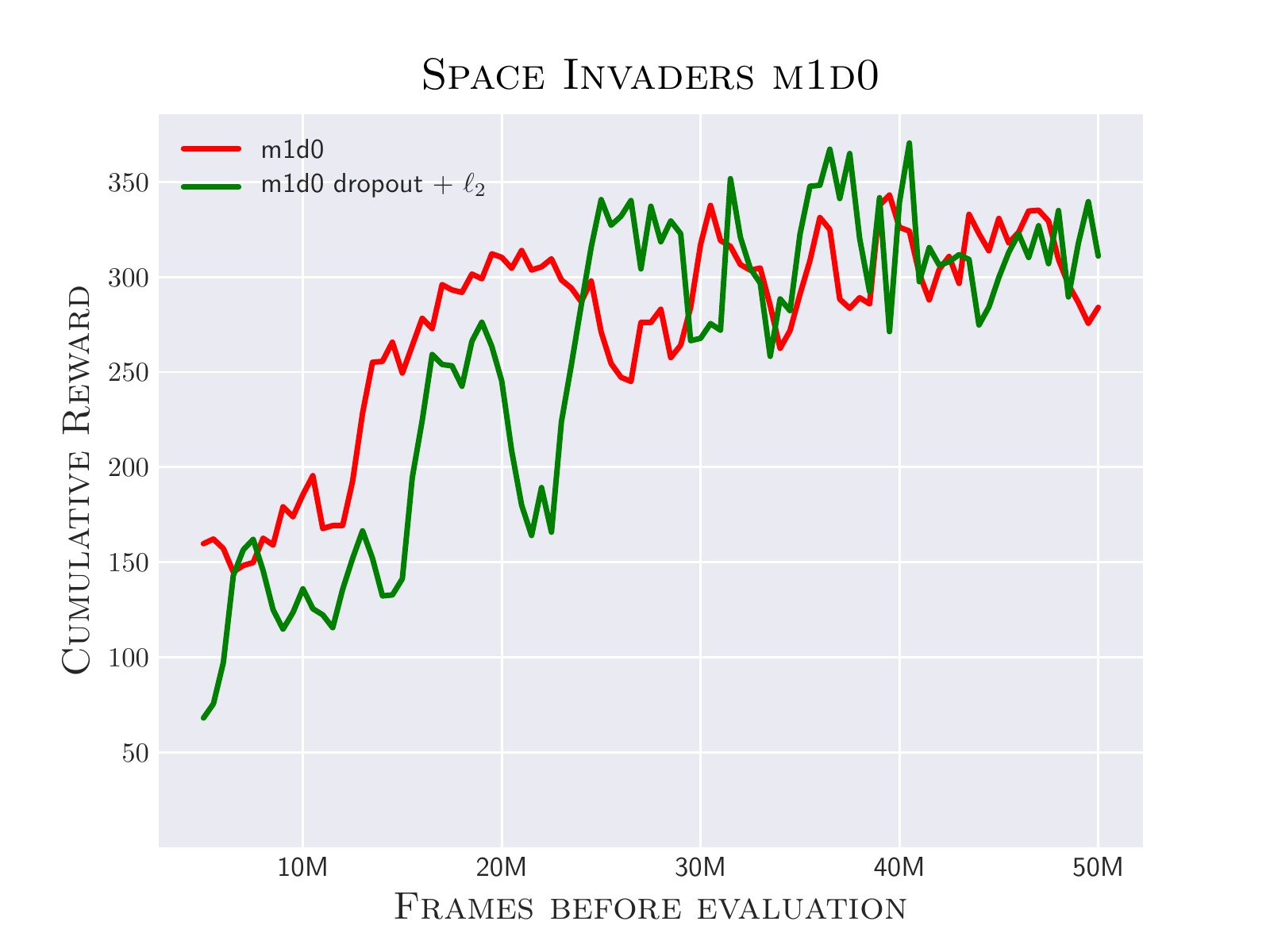}}
\subcaptionbox*{}{\includegraphics[width = .28\textwidth]{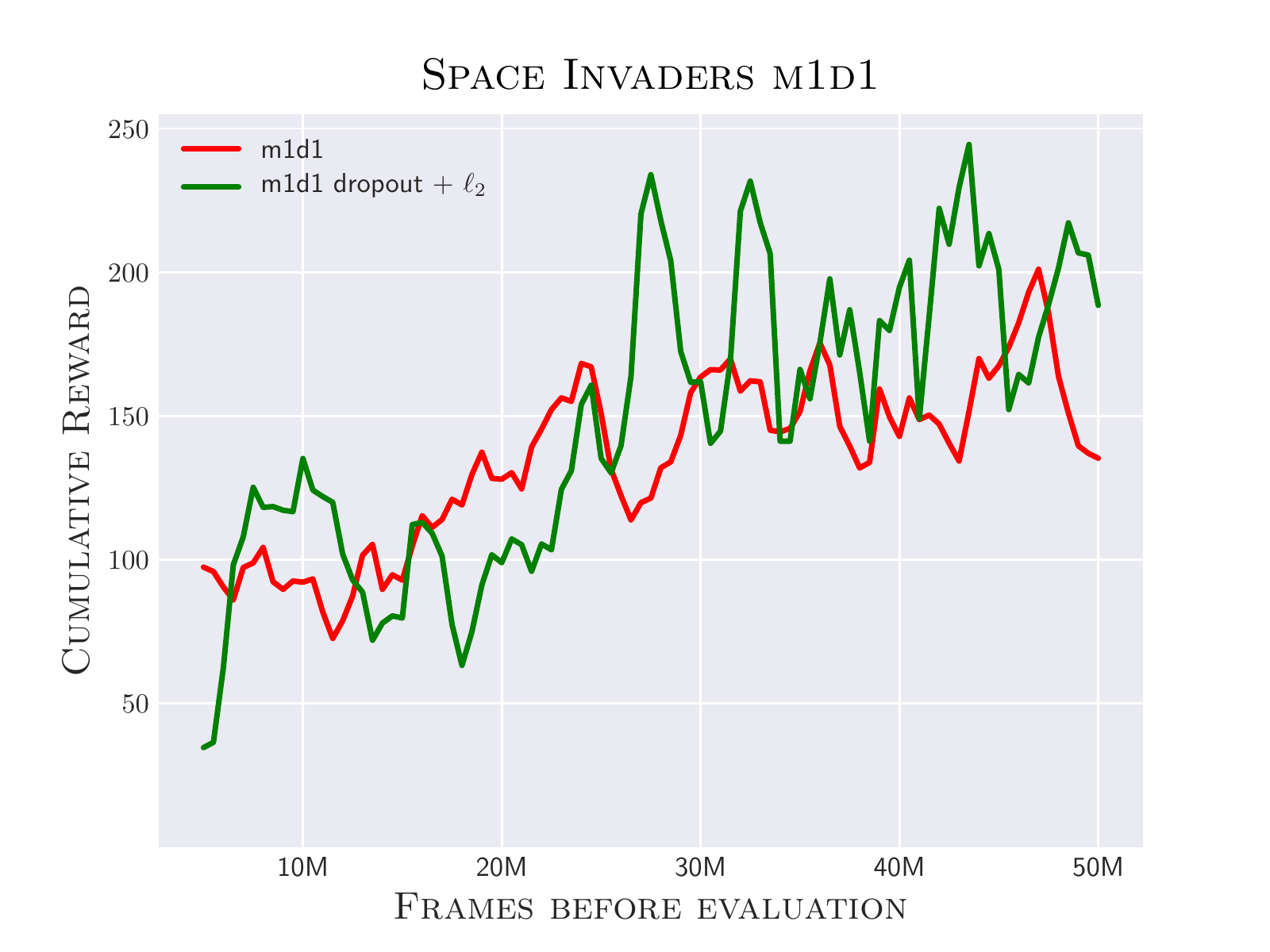}}
\subcaptionbox*{}{\includegraphics[width = .28\textwidth]{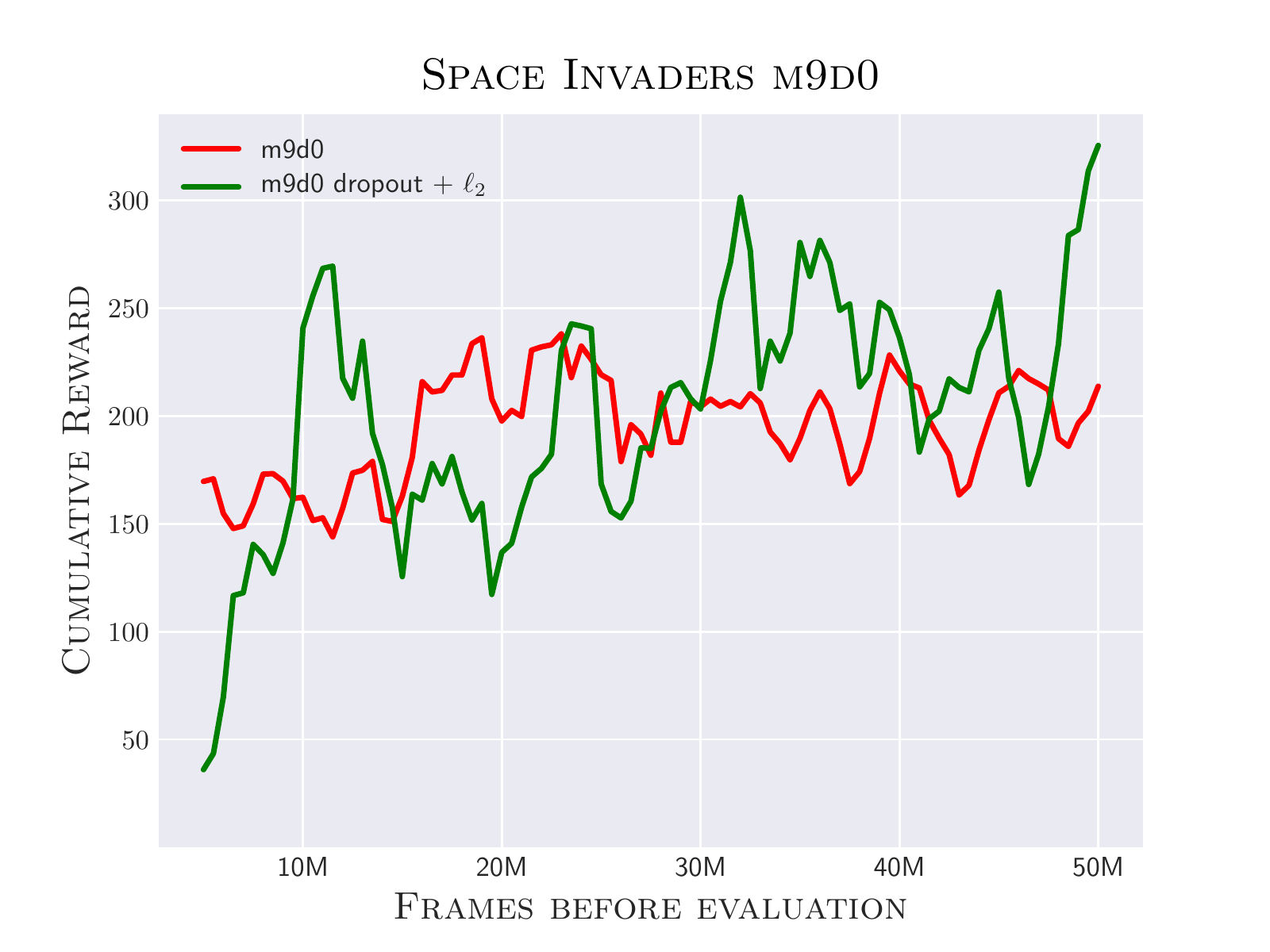}}
\caption{Performance curves for policy evaluation results. The x-axis is the number of frames before we evaluated the $\epsilon$-greedy policy from the default flavour on the target flavour. The y-axis is the cumulative reward the agent incurred. Green curves depict performance with regularization and red curves without.}
\end{figure}

\end{document}